\documentclass[10pt,twocolumn,letterpaper]{article}
\newtheorem{theorem}{Theorem}
\newtheorem{corollary}{Corollary}
\newcommand{\zq}[1]{\textcolor{black}{#1}}

\usepackage[pagenumbers]{iccv} 

\usepackage[pagebackref,breaklinks,colorlinks,allcolors=iccvblue]{hyperref}
\usepackage{amsmath}

\definecolor{iccvblue}{rgb}{0.21,0.49,0.74}

\def\paperID{3710} 
\def\confName{ICCV}
\def\confYear{2025}

\title{Feature Alignment with Equivariant Convolutions for Burst Image Super-Resolution}

\author{Xinyi Liu\\
Xi’an Jiaotong University\\
{\tt\small xinyil2525@gmail.com}
\and
Feiyu Tan\\
Xi’an Jiaotong University\\
{\tt\small tanfy929@stu.xjtu.edu.cn}
\and
Qi Xie\\
Xi’an Jiaotong University\\
{\tt\small xie.qi@xjtu.edu,cn}
\and
Qian Zhao\footnote{Corresponding author}\\
Xi’an Jiaotong University\\
{\tt\small timmy.zhaoqian@xjtu.edu.cn}
\and
Deyu Meng\\
Xi’an Jiaotong University\\
{\tt\small dymeng@xjtu.edu.cn}
}

\begin{document}

\maketitle
\begin{abstract}
Burst image processing (BIP), which captures and integrates multiple frames into a single high-quality image, is widely used in consumer cameras. \zq{As a typical BIP task, Burst Image Super-Resolution (BISR) has achieved notable progress through deep learning in recent years.} Existing BISR methods typically involve \zq{three key stages:} alignment, \zq{upsampling, and fusion}, often in varying orders and implementations. \zq{Among these stages}, alignment is particularly critical for ensuring accurate feature matching and further reconstruction. However, existing methods often rely on techniques such as \zq{deformable convolutions and optical flow to realize alignment, which either focus only on local transformations or lack theoretical grounding, thereby} limiting their performance. To alleviate these issues, we propose a novel framework \zq{for BISR}, featuring an equivariant convolution-based alignment, ensuring consistent transformations between the image and feature domains. This enables \zq{the alignment transformation to be learned via explicit supervision in the image domain and easily applied in the feature domain in a theoretically sound way,} effectively improving alignment accuracy. Additionally, we design an effective reconstruction module with advanced deep architectures for upsampling and fusion to obtain the final BISR result. Extensive experiments on BISR benchmarks show the superior performance of our approach in both quantitative metrics and visual quality.


\end{abstract}

    
\section{Introduction}
\label{sec:intro}

\zq{Image super-resolution is an important task in image processing. Conventionally, it is mainly dealt with in the context of Single Image Super
Resolution (SISR) \cite{wang2020sr1, dong2015sr2} and significant progress has been made in the last decades. In recent years, by the advances in image acquisition technologies, a new kind of super-resolution technique,} Burst Image Super-Resolution (BISR) \cite{bhat2021ntire, bhat2022ntire} has emerged as an increasingly valuable alternative. Unlike SISR, BISR reconstructs a high-resolution (HR) image by leveraging a sequence of low-resolution (LR) images captured in rapid succession, making it inherently more robust to noise and artifacts. Despite its advantages, BISR faces significant challenges, including \zq{accurate alignment for} handling motion variations and effective multi-frame fusion.


\begin{figure}[tp]
    \centering
    \includegraphics[width=\linewidth]{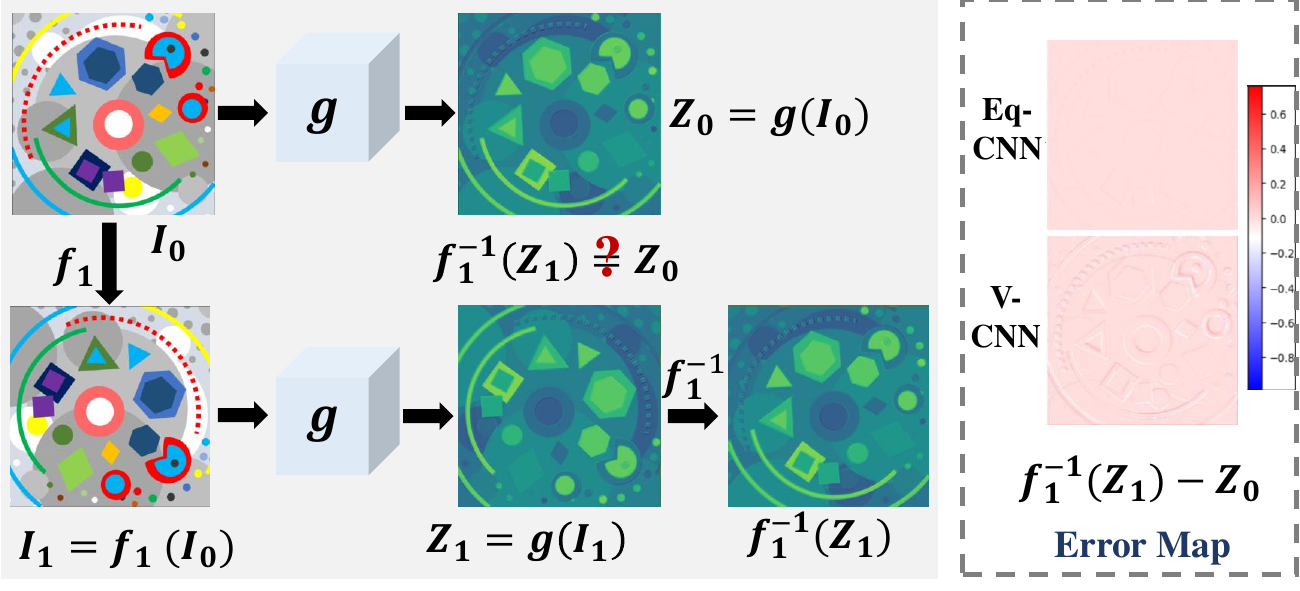}
    \vspace{-0.6cm}
    \caption{Illustration of transformation consistency in vanilla (V-CNN) and equivariant (Eq-CNN) convolutional networks. \zq{$f_1$ denotes a transformation (rotation in this example) and $g$ is a CNN that extracts features from images. Suppose $I_1$ is the image obtained by applying $f_1$ to $I_0$, i.e., $I_1=f_1(I_0)$, and $Z_0$ and $Z_1$ are features extracted from $I_0$ and $I_1$, respectively. We expect that $Z_1$ can be close to $f_1(Z_0)$, the transformation of $Z_0$, such that one can align $Z_1$ to $Z_0$ in the feature domain by applying the inverse transformation $f_1^{-1}$, which can be learned by explicit supervision in the image domain. The right box compares the error between $f_1^{-1}(Z_1)$ and $Z_0$, and it can be observed that Eq-CNN can more effectively achieve this goal than V-CNN.}}
    \vspace{-0.7cm}
    \label{fig:rotation}
\end{figure}

The general pipeline for BISR typically involves three key stages: alignment, upsampling, and fusion, with their order and implementation varying across methods. \zq{Among them,} alignment 
generally plays a crucial role in addressing spatial misalignments between successive frames. It ensures accurate feature matching and enables high-quality image reconstruction. \zq{Currently, existing methods mainly adopt two kinds of approaches to realize alignment, i.e., Deformable Convolution Networks (DCN) \cite{dai2017deformable} and optical flow \cite{opticalflow}. For instance, in \cite{bipnet, dudhane2023burstormer}, DCN was used to extract and align features, demonstrating strong capabilities in modeling local spatial transformations, and in \cite{kang2025burstm}, optical flow was adopted, enabling supervised alignment in the image domain. However, although they have demonstrated their effectiveness in BISR, these two approaches still have limitations. Specifically, DCN may not well model the global transformation \cite{kang2025burstm} and is also difficult to be explicitly supervised using the image domain information, while the optical flow estimated in the image domain may not be strictly valid in the feature domain.}

 
To alleviate these limitations \zq{of alignment in existing methods,} we propose to leverage equivariant convolutional networks (Eq-CNNs) \cite{weiler2018learning,xie2022fourier,fonvunfolding} {with learnable transformation matrices} as a solution. \zq{Compared with vanilla convolutional networks (V-CNNs) \cite{krizhevsky2012imagenet, edsr}, Eq-CNNs can extract features that are theoretically equivariant to input images under certain spatial transformations, e.g., rotation and translation. Then, suppose each source frame within burst images can be approximately modeled by such a transformation (or more specifically, rotation plus translation) of the reference frame due to the acquisition mechanism \cite{bhat2021ntire}. In that case, the equivariance property of Eq-CNNS enables us to learn the transformation (or its inverse) with the image domain supervision and then apply the inverse transformation in the feature domain to achieve an easy while theoretically sound alignment, 
as illustrated in Fig.~\ref{fig:rotation}.}


\zq{With the aligned features as aforementioned, we further designed a reconstruction module for upsampling and fusion to generate the final sRGB image using advanced techniques. Specifically, considering its ability in capturing intricate inter-frame correlations, we adopt the Multi-Dconv Head Transposed Attention (MDTA) block \cite{zamir2022restormer} for feature interaction among frames; and due to its flexibility in multi-scale upsampling, we use the implicit neural representation (INR) technique \cite{lte} to upsample the features for final fusion, following \cite{kang2025burstm}.}


To summarize, our contributions are as follows:

\begin{itemize}
\item We propose a new alignment framework \zq{for BISR} based on Eq-CNN, \zq{which enables us to learn the alignment transformation with image domain supervision and apply it in the feature domain in a theoretically sound way. The corresponding theoretical analysis also advances the theory of Eq-CNN to a certain extent.} 
\item \zq{We incorporate the proposed alignment framework with advanced techniques for upsampling and fusion, including MDTA from Restomer and INR, and build a new deep model for the BISR task.}
\item \zq{We apply the proposed model to BISR benchmarks,} demonstrate its superiority 
\zq{against current state-of-the-art methods.}
\end{itemize}












\section{Related Work}
\label{sec:related}
\begin{figure*}[tp]
    \centering
\includegraphics[width=\linewidth]{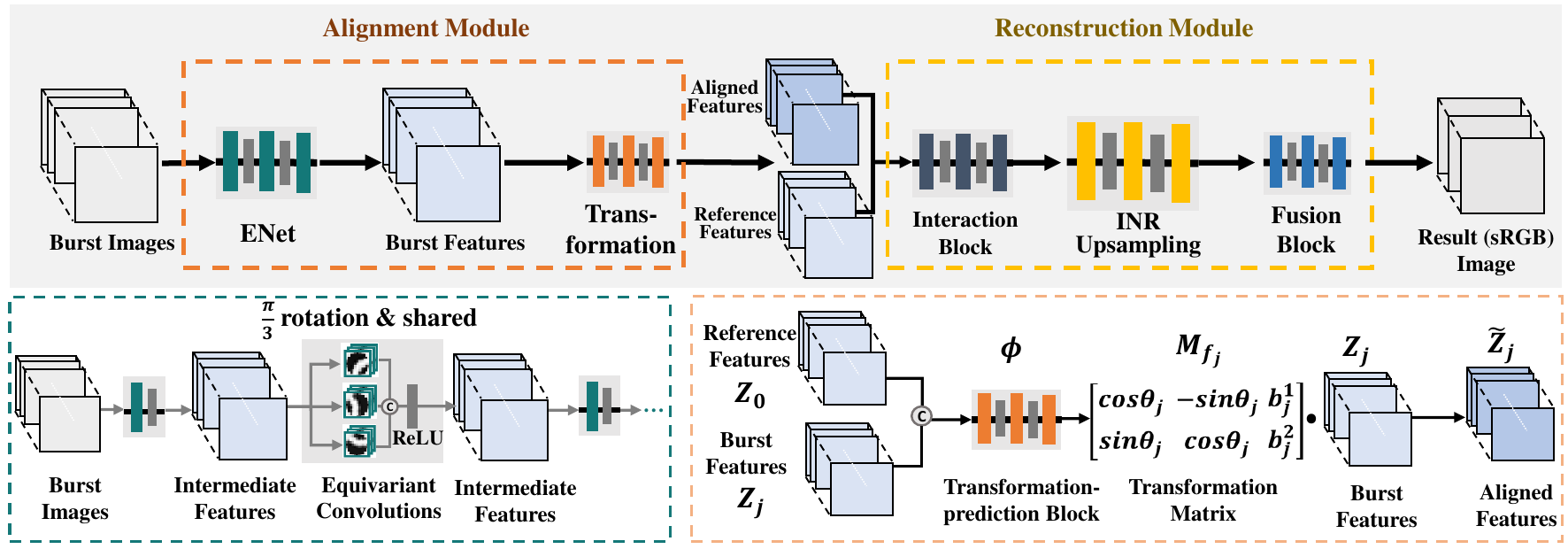}
    \vspace{-5mm}
    \caption{Overview of our proposed method. The top row shows the whole workflow. The bottom left shows the detailed equivariant convolution layers of ENet. The bottom right shows the process of feature alignment by predicted transformation, as in Section \ref{sec:align}. }
    \vspace{-0.5cm}
    \label{fig:network}
\end{figure*}
\subsection{Burst image super-resolution}
BISR and its related task, \zq{Muti Frame Super-Resolution (MFSR)}, have been extensively studied using both traditional approaches and deep learning techniques. The pioneering work by Tsai et al. \cite{tsai1984multiframe} tackled the problem in the frequency domain, while subsequent research \cite{irani1991improving, ur1992improved, 650118} put more focus on the spatial domain for resolution enhancement. 

With the rapid advances of deep learning, \zq{significant progress has been made in this area.} Initial studies \cite{ustinova2017deep, molini2019deepsum} employed \zq{relatively} simple network architectures to address the MFSR task. \zq{Then, Bhat et al. \cite{dbsr} proposed a BISR pipeline} incorporating alignment, feature fusion, and upsampling modules, \zq{which has inspired numerous studies \cite{mfir, luo2022bsrt, bipnet, gmtnet, dudhane2023burstormer, bsrd, di2024qmambabsr}. The first real-world burst SR benchmark was also introduced in \cite{dbsr}.}

\zq{Within the BISR pipeline, alignment plays an important role and was specifically emphasized by Kang et al. \cite{kang2025burstm}. While many methods adopted DCN \cite{dai2017deformable} for alignment \cite{bipnet, dudhane2023burstormer}, it is not effective enough to cover global alignments, as pointed in \cite{kang2025burstm}. In contrast, Kang et al. \cite{kang2025burstm} introduced optical flow \cite{opticalflow} to achieve alignment, improving performance. However, since the optical flow is estimated in the image domain, there is no strict guarantee that the alignment transformation is theoretically valid in the feature domain. This motivates us to develop a more theoretically sound alignment technique via Eq-CNN.}

\zq{In addition to alignment, the upsampling and fusion stages have also been advanced by recently developed neural architectures and techniques. For example,} Transformer-based architectures \cite{transformer} were adopted in the Burstomer \cite{dudhane2023burstormer} and GMTNet \cite{gmtnet} methods to enhance feature representation by long-range dependency modeling; MAMBA \cite{gu2023mamba} was introduced in QMambaBSR \cite{di2024qmambabsr} to better integrate sub-pixel information through efficient state-space modeling; diffusion models \cite{ddpm} was employed in BSRD \cite{bsrd} to refine reconstruction results; and INR \cite{liif,lte} technique was used in BurstM \cite{kang2025burstm} to realize multi-scale upsampling.

\subsection{Equivariant convolutions}
One of the key reasons for the success of CNNs in computer vision tasks is their inherent translation-equivariance, which ensures spatial consistency by making a translation in the input equivalent to translations in all intermediate feature maps and the output. Based on this concept, rotation-equivariant convolutions have been further developed. For instance, GCNN \cite{gcnn} and HexaConv \cite{hoogeboom2018hexaconv} explicitly incorporate $\frac{\pi}{2}$ and $\frac{\pi}{3}$ rotational equivariance, respectively. Moreover, Xie et al. \cite{fconv} proposed a Fourier series expansion-based filter parameterization, extending rotation equivariance to nearly continuous degrees, which has demonstrated its advanced capabilities and practical value in applications such as \cite{fonvunfolding}. \zq{The equivariance property of Eq-CNNs provides a possibility for supervised learning of a transformation using the image domain information, which is still theoretically valid in the feature domain. This is crucial in the designing of our alignment framework, as discussed with more detials in Section~\ref{sec:pre}.}

\section{Proposed Method\label{sec:method}}

\zq{We first provide} preliminaries of equivariant convolutions \zq{to motivate our alignment framework for BISR. Then, we discuss the details of our proposed method, including the network architectures of both the alignment and reconstruction modules. We also provide} a theoretical justification for the validity of the proposed \zq{alignment framework.}


\subsection{Motivation}\label{sec:pre}
\zq{We first briefly introduce the concept of equivariance in deep learning. Suppose $g$ is a deep feature extractor mapping from input to the feature space and $F$ is a transformation group, we say $g$ is equivariant with respect to $F$ if for any $f\in F$, it holds that $f(g(I)) = g(f(I))$, or equivalently, $g(I) = f^{-1}\big(g(f(I))\big)$ if $f$ is invertible. Note that here, we abuse the notation $f$ a little to denote the same transformation applied in different domains. It is well known that V-CNNs are only equivariant with respect to translation, while previous studies on equivariance \cite{weiler2018learning, fconv} constructed CNNs that are also equivariant with respect to rotation and reflection, which are often referred to as equivariant CNNs (Eq-CNNs).}

\zq{The equivariance property of Eq-CNNs enables us to develop an effective and theoretically sound approach for alignment. Specifically, we notice that burst images are captured by repeated exposures in a short time, and thus, the misalignments between frames are mainly caused by slight camera shifts and are generally not very spatially significant. Therefore, we may approximately model each source frame as a rotation and translation of the reference frame \cite{bhat2021ntire}. Then, if we use an Eq-CNN to extract features of the original frames, the transformation consistency between the image domain and feature domain can be naturally guaranteed. Consequently, we can learn such transformations (or their inverses) via the image domain supervision while applying them in the feature domain for alignment. The details are provided in Section \ref{sec:align}.}


\subsection{Our method}

\subsubsection{Problem setting and processing pipeline}
Given $B$ low-resolution (LR) RAW burst frames $\{I_j^{L}\}_{j=0}^{B-1}$ with each $I_j^{L} \in \mathbb{R}^{h \times w \times 1}$, we first process it a 4-channel format following the RGGB Bayer pattern \cite{kang2025burstm, dudhane2023burstormer}. Then, one frame is selected as the reference frame, which serves as the reference for high-resolution (HR) reconstruction, and the rest frames are used to assist the reference one in super-resolving. \zq{The reconstructed HR image $I^S\in\mathbb{R}^{sh \times sw \times 3}$ is in sRGB format, where $s$ is the scale factor. As shown in Fig.~\ref{fig:network}, our processing pipeline includes two main steps, i.e., alignment and reconstruction. The alignment step aims to extract and align features from the LR burst images using the Eq-CNN, and the reconstruction step tries to upsample and fuse the features to get the final reconstruction.}


\subsubsection{Alignment Module \label{sec:align}}
Let \( I_0^L \) denote the RAW LR reference frame and \( \{I_j^L\}_{j=1}^{B-1} \) represent the remaining source frames in the burst image. Following the discussions in Section \ref{sec:pre}, we approximately model the relationship between each $I_j^L~(j\neq0)$ and $I_0^L$ as
\begin{equation}\label{eq:e_i}
    I_j^L = f_j(I_0^L), 
\end{equation}
where $f_j$ is a rotation-translation transformation. After feature extraction using an Eq-CNN $g$, which is referred to as \emph{ENet} in Fig.~\ref{fig:network}, we can obtain features $Z_0=g(I_0^L)$ and $Z_j=g(I_j^L)$, respectively. Assuming the equivariance property of $g$ strictly holds, we have that
\begin{equation}\label{eq:e_x}
    Z_j =g(I_j^L)= g(f_j(I_0^L)) = f_j(g(I_0^L)) = f_j(Z_0). 
\end{equation}
This indicates that, if we can accurately estimate $f_j$ or $f_j^{-1}$ in the image domain,  the equivariance property of Eq-CNN allows us to directly apply $f_j^{-1}$ to $Z_j$:
\begin{equation}
    \tilde{Z}_j = f_j^{-1}(Z_j), \label{eq:inv}
\end{equation}
such that $\tilde{Z}_j$ is well algined to $Z_0$.

Therefore, we learn $f_j^{-1}$ via the image domain supervision with the following loss:
\begin{equation}\label{eq:loss_align}
    \mathcal{L}_{\text{align}} = \frac{1}{B-1} \sum\nolimits_{j=1}^{B-1} \| f_j^{-1}(I_j^L) - I_0^L \|_2^2.
\end{equation}
\zq{In practice, however, due to the discretization of the rotation angles in Eq-CNNs, we cannot strictly guarantee that $f_j^{-1}(Z_j)=Z_0$. Nevertheless, we can provide a theoretical justification to ensure that $f_j^{-1}(Z_j)$ can be close to $Z_0$ by minimizing Eq.~\eqref{eq:loss_align}. To ensure the smooth flow of the text, we put the theoretical results in Theorem \ref{thm:1} and Corollary \ref{cor1} of Section \ref{sec:theory} and continue to discuss our method here.}


\zq{The next question is then how to parameterize the transformation $f_j^{-1}$. Since we assume $f_j$ is a rotation-translation transformation, its inverse $f_j^{-1}$ is also a rotation-translation transformation, which can be parameterized using a matrix $M_{f_j}\in\mathbb{R}^{2\times3}$ with the following form:
\begin{equation}
    M_{f_j} = \begin{bmatrix}
    \cos\theta_j & -\sin\theta_j & b_j^1 \\
    \sin\theta_j & \cos\theta_j & b_j^2
    \end{bmatrix},
\end{equation}
where $\theta_j$ is the rotation angle, and $\boldsymbol{b}_j=(b_j^1,b_j^2)^T$ is the translation vector. Then, a pixel at location $(x_1,x_2)^T$ will be mapped to a new location $(x_1',x_2')^T$ after applying $f_j^{-1}$:
\begin{equation}
\begin{bmatrix}
x_1' \\
x_2'
\end{bmatrix}
= M_{f_j} \cdot \begin{bmatrix}
x_1 \\
x_2 \\
1
\end{bmatrix}
= \begin{bmatrix}
x_1\cos\theta_j -x_2\sin\theta_j + b_j^1 \\
x_1\sin\theta_j +x_2\cos\theta_j + b_j^2
\end{bmatrix}.
\end{equation}
Then we further parameterize $\{\theta_j,\boldsymbol{b}_j\}$ using a network block $\phi$, referred to as transformation prediction block in Fig.~\ref{fig:network}, with $Z_0,Z_j$ as its input:
\begin{equation}
    \{\theta_j,\boldsymbol{b}_j\}=\phi\left(\mathrm{concat}[Z_0,Z_j]\right),
\end{equation}
such that we can directly predict the alignment transformation $f_j^{-1}$ during inference.
}

\subsubsection{Reconstruction Module}
\zq{After alignment, the aligned features $\{\tilde{Z}_j\}_{j=0}^{B-1}$ of all frames (we let $\tilde{Z}_0\triangleq Z_0 $ for convenience) are then further processed for reconstructing the HR sRGB image.}

\noindent\textbf{Feature interaction}. Before upsampling, \zq{interacting the features between the reference and source frames} is important for enriching \zq{information of each frame.} While existing methods either concatenate the features of reference and source frames directly \cite{bipnet, dudhane2023burstormer, kang2025burstm} or use attention maps for pixel-wise interaction \cite{fbanet}, \zq{there are still limitations in modeling capacity and computational efficiency. Therefore, we convert to another solution by adopting the MDTA block from Restormer \cite{zamir2022restormer}. The key designing mechanism behind MDTA is to apply} self-attention across channels rather than spatial dimensions, computing cross-covariance to generate attention maps that implicitly encode global context. Such a design not only reduces computational complexity but also enables more effective feature integration by modeling long-range dependencies between features of the reference and source frames.
The interaction process for the aligned features of all frames is as follows ($j=0,\dots,B-1$):
\begin{align}
\textbf{Q}_j &= W_d^QW_p^Q(\text{concat}[\tilde{Z}_j, \tilde{Z}_0]), \\
\textbf{K}_j &= W_d^KW_p^K(\text{concat}[\tilde{Z}_j, \tilde{Z}_0]), \\
\textbf{V}_j &= W_d^VW_p^V(\text{concat}[\tilde{Z}_j, \tilde{Z}_0]), \\
\hat{Z}_j & = \mathbf{V}_j \cdot \text{Softmax}(\mathbf{K}_j \cdot \mathbf{Q}_j/\alpha_j) + \tilde{Z}_j,
\end{align}
where $\{\hat{Z}_j\}_{j=0}^{B-1}$ are the features after interaction, $W_p^{(\cdot)}$ and $W_d^{(\cdot)}$ refer to $1\times1$ pixel-wise and $3\times3$ depth-wise convolutions, respectively, and $\alpha_j$ is a learnable scaling parameter. 
The term $\text{Softmax}(\mathbf{K}_j \cdot \mathbf{Q}_j/\alpha_j)$ captures feature correlations and dynamically weights value vectors based on similarity, enabling context-aware fusion. Compared with element-wise multiplication or concatenation, this attention mechanism is more effective in capturing global correlations and adjusting feature contributions. \zq{After the above interaction process, the features are expected to be sufficiently enriched and ready for upsampling and final fusion.}

\noindent\textbf{Upsampling and final fusion.}
For upsampling, we adopt the LTE framework \cite{lte}, leveraging the INR technique and frequency domain processing, to recover the high-frequency details. \zq{LTE has two advantages particularly important for the BISR task. Firstly, as shown in \cite{kang2025burstm}, due to the nature of INR, the LTE-based architecture is good at multi-scale BISR, such that we can train a single model to cover multiple application scenarios. Secondly, as a core component of LTE, the grid sampling mechanism} enables the upsampling of sub-pixel information from original frames. This is especially crucial for burst images \zq{since the fine-grained details introduced by slight camera shifts can be expected to be effectively recovered by such a mechanism.} \zq{After upsampling,} a fusion block with the channel attention mechanism is introduced to fuse the upscaled features, and a skip connection is employed to also preserve the information from the original reference frame.


To be specific, the upsampling and final fusion process can be formulated as
\begin{equation}
    I^S = \text{PS}\left(\tilde{I}_0^L\uparrow + \text{Avg}_W\left(\{\Phi_{\text{up}_j}(\hat{Z}_j)\}_{j=0}^{B-1}\right)\right),
\end{equation}
where \( I^S \in \mathbb{R}^{sh \times sw \times 3} \) is the final output in sRGB format, \(\text{PS}(\cdot)\) denotes the pixel shuffle operation, \(\text{Avg}_W(\cdot)\) refers to the weighted average operation with parameters learned via convolutions from \(\hat{Z}_j\)s, \(\Phi_{\text{up}_j}(\cdot)\) denotes the LTE-based upsampling, and \(\tilde{I}_0^L\uparrow \in \mathbb{R}^{(sh/2) \times (sw/2)\times 12}\) is the upsampled reference frame obtained by
\begin{equation}
    \tilde{I}_0^L\uparrow = \text{Conv}_{1\times 1}(\text{Up}[I_0^L]),
\end{equation}
where $\text{Up}[\cdot]$ refers to the bilinear upsampling operation, and $\text{Conv}_{1\times 1}(\cdot)$ is a $1\times 1$ convolution layer.



\subsubsection{Training loss}
The whole network is trained in an end-to-end way using the following loss:
\begin{equation}
    \mathcal{L} = \mathcal{L} _{\text{align}} + \mathcal{L} _{\text{fidelity}},
\end{equation}
where $\mathcal{L} _{\text{align}}$ is defined in Eq. \eqref{eq:loss_align}, and $\mathcal{L} _{\text{fidelity}}$ is defined as
\begin{equation}
    \mathcal{L} _{\text{fidelity}}=\left\|I^S-I^{\mathrm{GT}}\right\|_1,
\end{equation}
with $I^{\mathrm{GT}}$ being the ground truth HR sRGB image.

\subsection{Theoretical justification}\label{sec:theory}
In this section, we provide theoretical results to show the validity of the proposed alignment framework. Due to space limitations, we only provide the sketch of the results, and the full presentations and proofs are provided in the supplementary material. Firstly, considering the discretization of the rotation angles in common Eq-CNN implementations, we have the following theorem.
\begin{theorem}\label{thm:1}
For an image \( I_0 \) of size \( H \times W \times C \), a rotation-translation Eq-CNN \( g(\cdot) \) with discretized angles, and a rotation-translation transformation \( f_j(\cdot) \), under certain assumptions, the following result holds:
\begin{equation}\label{main_conclusion}
  \| f_j^{-1}\big(g(f_j(I_0))\big) - g(I_0) \|_{\infty}
  \leq C_1 h^2 + C_2 pht^{-1},
\end{equation}
where $t,p,h, C_1, C_2$ are constants.
\end{theorem}
Different from existing theories in Eq-CNN showing that input transformations can be predictably reflected in the feature domain, by measuring the error between $g(f_j(I_0))$ and $f_j(g(I_0))$, Theorem \ref{thm:1} further analyzes the residual errors caused by inverse transformation applied to these two objects in discrete settings of Eq-CNN, and suggests that such an error can also be upper-bounded. Such an analysis provides a theoretical understanding of how input-level inverse transformations affect feature relationships, which advances the theory of Eq-CNN to a certain extent. With Theorem \ref{thm:1}, we can easily obtain the following corollary.
\begin{corollary}\label{cor1}
Under the same assumptions and definitions of Theorem \ref{thm:1},
for an image $I_j$ with the same size of $I_0$, let ${Z_0}=g(I_0)$ and ${Z_j}=g(I_j)$ be the feature maps, where $Z_0,Z_j\in \mathbb{R}^{H\times W\times tC}$, and then the following result holds:
\begin{equation}\label{eq:thm1}
\begin{split}
\|f_j^{-1}\!(Z_j)\!-\!Z_0\|_{\infty}\!\leq\!C_3\!\left\|f_j^{-1}\!(I_{j})\!-\!I_0\!\right\|_2\!+\!C_1 h^2\!+\! C_2 {p h}{t^{-1}},
\end{split}
\end{equation}
where $t,p,h, C_1, C_2,C_3$ are constants.
\end{corollary}
Corollary \ref{thm:1} suggests that we can minimize the distance between $f_j^{-1}(Z_j)$ and $Z_0$, the main goal in Section \ref{sec:align}, through minimizing the distance between $f_j^{-1}(I_j^L)$ and $I_0^L$, which is we are trying to do by the loss defined in Eq. \eqref{eq:loss_align}.

\section{Experiments}
In this section, we conduct experiments to validate the effectiveness of our proposed method. We first evaluate the proposed method on standard benchmarks for BSIR in comparison with existing methods. Then, we conduct ablation studies to demonstrate the reasonablity of our method, specifically concerning the alignment mechanism.

\subsection{Experiments on BISR benchmarks}
\subsubsection{Settings}

\textbf{Datasets.}
We follow previous studies \cite{dudhane2023burstormer, kang2025burstm} and conduct experiments on two datasets:
\textbf{(1) SyntheticBurst Dataset} \cite{bhat2021ntire}, which consists of 46,839 burst sequences for training and 300 for validation.
Each burst sequence contains 14 RAW LR frames generated from an HR sRGB image using the standard pipeline \cite{bipnet, kang2025burstm}.  Specifically, 
unprocessing techniques \cite{brooks2019unprocessing} are firstly applied to simulate RAW sensor data, and random rotations and translations are implemented to simulate real camera motion.
Following \cite{kang2025burstm}, we generate multi-scale LR images through random down-sampling (×2, ×3, ×4). Finally, Bayer mosaicking and random noise are added to more closely reproduce real-world imaging conditions.
\textbf{(2) BurstSR Dataset} \cite{dbsr}, which comprises 200 full-size RAW burst sequences, with 5,405 patches of size 80×80 extracted for training and 882 patches for validation. The LR images are captured using a smartphone, while the HR ground truth images are obtained from a DSLR under the same scenes. Each LR burst sequence consists of 14 frames, and the scale factor between LR and HR images in this dataset is fixed (×4). 

\noindent \textbf{Competing methods and evaluation metrics.}
We evaluate our method against 8 representative ones, including traditional Bicubic interpolation and current state-of-the-art methods for the BISR task: DBSR \cite{dbsr}, MFIR \cite{mfir}, BIPNet \cite{bipnet}, GMTNet \cite{gmtnet}, Burstormer \cite{dudhane2023burstormer}, BSRT \cite{luo2022bsrt}, and BurstM \cite{kang2025burstm}. We employ two widely used metrics, PSNR and SSIM, to quantitatively assess the reconstruction quality of each method. Additionally, we report model complexity metrics, including the number of parameters and GFLOPs, to show the computational efficiency of each method as a reference.

\begin{table*}[ht]
    \centering
    \caption{Quantitative results on SyntheticBurst and BurstSR datasets. The best and second-best results are highlighted in bold and underlined, respectively.}
    \vspace{-2mm}
    \begin{tabular}{lcccccccccc}
        \toprule
        & \multicolumn{6}{c}{SyntheticBurst} & \multicolumn{2}{c}{BurstSR} & \multicolumn{2}{c}{BurstSR} \\
        \cmidrule(lr){2-7} \cmidrule(lr){8-9} 
        Method & \multicolumn{2}{c}{x2} & \multicolumn{2}{c}{x3} & \multicolumn{2}{c}{x4} & \multicolumn{2}{c}{x4} &  Params.(M) & FLOPs(G) \\
        \cmidrule(lr){2-3} \cmidrule(lr){4-5} \cmidrule(lr){6-7} \cmidrule(lr){8-9} 
        & PSNR & SSIM & PSNR & SSIM & PSNR & SSIM & PSNR & SSIM & \\
        \midrule
        Bicubic & 38.30 & 0.948 & 33.94 & 0.886 & 33.02 & 0.862 & 42.55 & 0.962 & - &-\\
        DBSR \cite{dbsr} & 40.51 & 0.965 & 40.11 & 0.959 & 40.76 & 0.959 & 48.05 & 0.984 & 13.01 & 111.71\\
        MFIR \cite{mfir} & 41.25 & 0.971 & 41.81 & 0.972 & 41.56 & 0.964 & 48.33 & 0.985 & 12.13 & 121.01\\
        BIPNet \cite{bipnet}& 37.58 & 0.928 & 40.83 & 0.955 & 41.93 & 0.960 & 48.49 & 0.985 & 6.7 & 326.47 \\
        Burstormer \cite{dudhane2023burstormer} & 37.06 & 0.925 & 40.26 & 0.953 & 42.83 & 0.973 & 48.06 & 0.986 & 2.5& 38.33 \\
        GMTNet \cite{gmtnet}& - & - & - & - & 42.36 & 0.961 & 48.95 & 0.986 & -&300 \\
        BSRT-Small \cite{luo2022bsrt} & 40.64 & 0.966 & 42.30 & 0.975 & 42.72 & 0.971 & 48.57 & 0.986 & 4.92 & 178.82\\
        BSRT-Large \cite{luo2022bsrt}& 40.33 & 0.965& 42.87 & 0.979 & \textbf{43.62} & \textbf{0.975} & 48.57 & 0.986 & 20.71 &362.63\\
        BurstM \cite{kang2025burstm}& \underline{46.01} & \textbf{0.985} & \underline{44.79} & \underline{0.982} & 42.87 & 0.973 & \underline{49.12} & \textbf{0.987} & 14.0 & 436.21\\
        \hline
        Ours & \textbf{46.10} & \textbf{0.985} & \textbf{44.95} & \textbf{0.983} & \underline{43.18} & \underline{0.974} & \textbf{49.22} & \textbf{0.987} & 8.7 & 170.21  \\
        \bottomrule
        \vspace{-0.6cm}
    \end{tabular}
    \label{tab:comparison_syn}
\end{table*}

\noindent \textbf{Implementation details.} 
All experiments are implemented using PyTorch on an NVIDIA 4090 GPU. On the SyntheticBurst dataset, the initial learning rate for our model is set to \(1 \times 10^{-4}\) and gradually adjusted to \(1 \times 10^{-6}\) over 300 epochs. The batch size is 1, and the patch size is \(48 \times 48\).
On the BurstSR dataset, we fine-tune the model pre-trained on SyntheticBurst following \cite{dudhane2023burstormer,kang2025burstm}, using an initial learning rate of \(1 \times 10^{-5}\) and CosineAnnealingLR to adjust it to \(1 \times 10^{-6}\) over 30 epochs. The batch size is 1, and the patch size is \(80 \times 80\).
For other compared deep learning-based methods, we test using the author-released models, except GMTNet for which we directly quote the results reported in the original paper since the model is not released.

\subsubsection{Results}
\begin{figure*}[tp]
\centering
\includegraphics[width=0.95\linewidth]{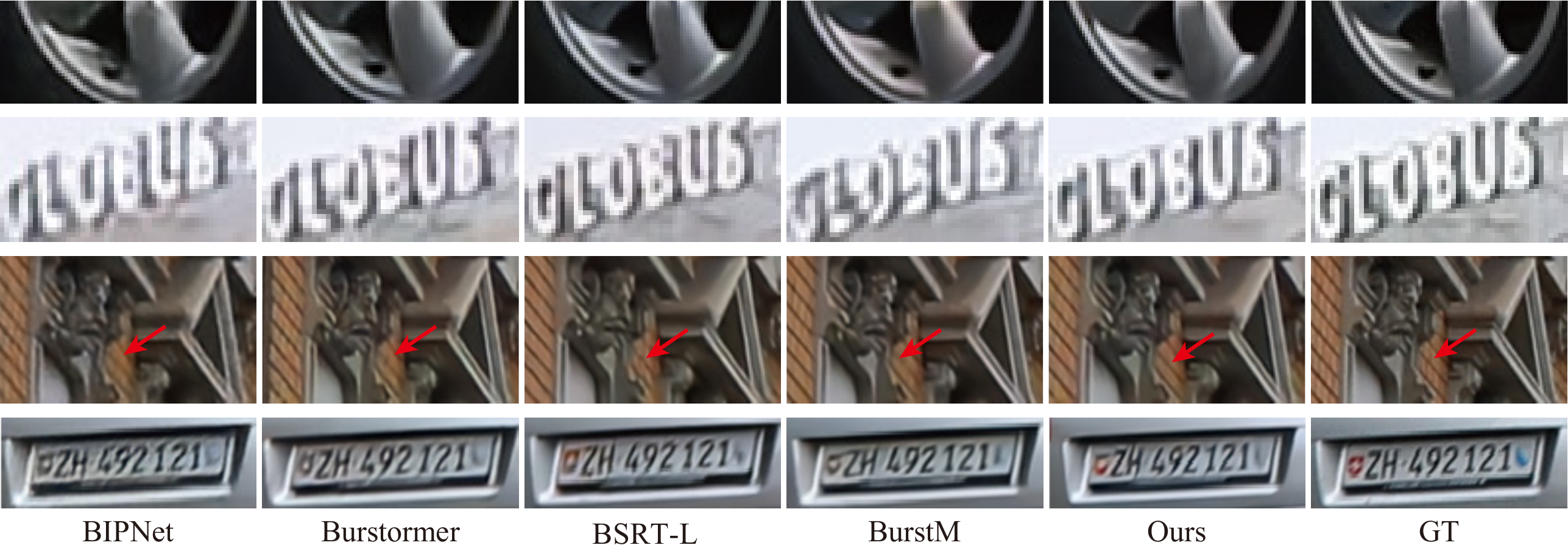}
\vspace{-0.5cm}
\caption{Visual comparison of ×4 BISR on the SyntheticBurst dataset.}
\vspace{-0.4cm}
\label{fig:syn_1}
\end{figure*}

\begin{figure*}[tp]
\centering
\includegraphics[width=0.95\linewidth]{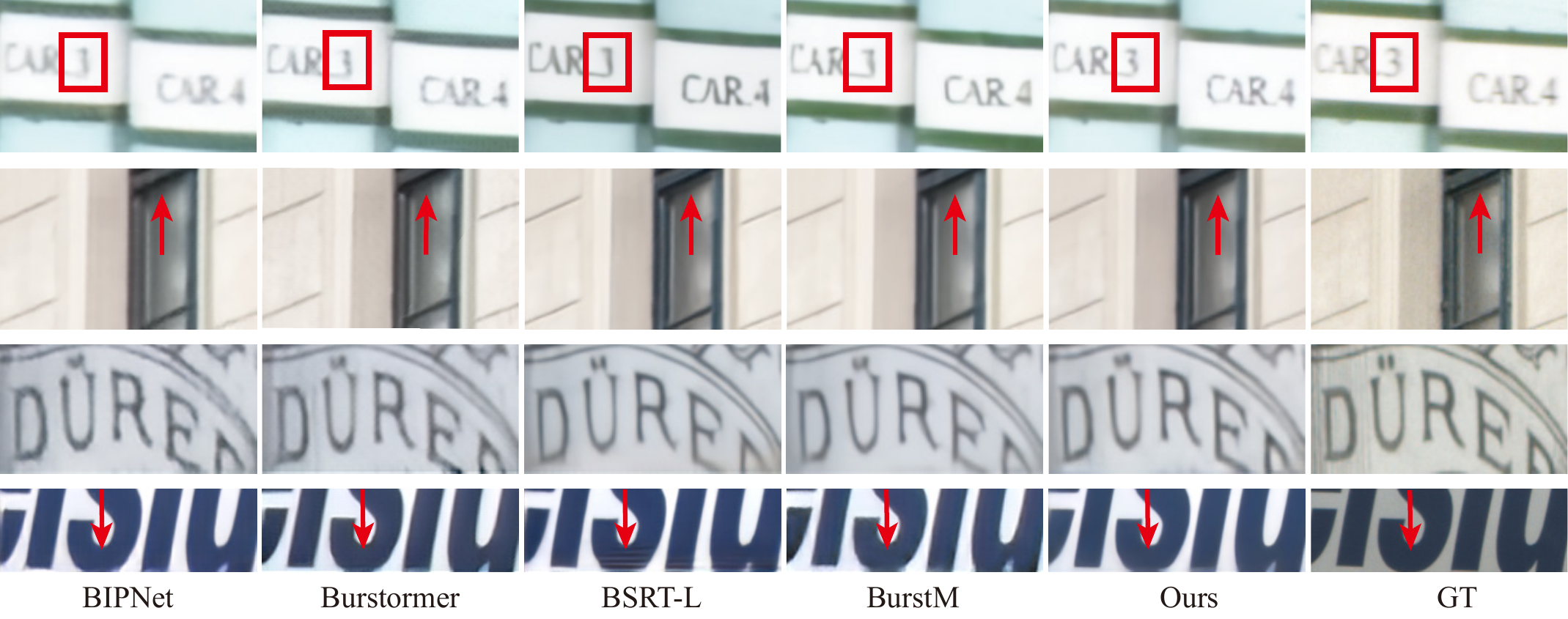}
\vspace{-0.4cm}
\caption{Visual comparison of ×4 BISR on the BurstSR dataset.}
\vspace{-0.3cm}
\label{fig:real_1}
\end{figure*}

\textbf{Results on SyntheticBurst Dataset} \cite{bhat2021ntire}. We present the quantitative and qualitative evaluation results in Table~\ref{tab:comparison_syn} and Fig.~\ref{fig:syn_1}, respectively, with full-size and additional visual results available in the supplementary material because of space limitation.

As shown in Table~\ref{tab:comparison_syn}, our method outperforms existing BISR approaches across nearly all evaluation metrics. Specifically, for the widely-used ×4 SR setting, our method achieves the results with a PSNR of 43.18 and an SSIM of 0.974, surpassing competing methods with comparable model complexities. This quantitatively demonstrates its effectiveness in reconstructing the original HR sRGB image. Notably, our approach outperforms the current state-of-the-art multi-scale BISR method BurstM \cite{kang2025burstm} while requiring fewer parameters and less FLOPs, showing both its efficiency and effectiveness. Furthermore, our method consistently performs well across different SR factor settings, suggesting its promising generalization ability.  
Visual results in Fig.~\ref{fig:syn_1} show that our method achieves competitive performance in several aspects. For example, our approach better preserves fine-grained textual details while maintaining structural fidelity.
In addition, the method shows its ability to suppress noise without introducing unexpected artifacts or severe color distortions.
These qualitative advantages of our method are consistent with its quantitative performance.
It should be mentioned that though our model achieves slightly lower numeric results at the ×4 scale due to its fewer parameters (8.7M) compared to BSRT-Large (20.71M), it delivers comparable visual quality. 
Additional visual results of other scales are provided in the supplementary material for a more comprehensive comparison.

\noindent\textbf{Results on BurstSR Dataset}. The quantitative results on the BurstSR dataset are summarized in Table~\ref{tab:comparison_syn}. 
It can be seen that our method achieves the best performance in terms of both PSNR and SSIM among all competing ones. Note that, in this real dataset, although the degradation process of the LR burst images is unknown, and the relationship between the source and reference frames might be more complex than assumed, our method still performs promisingly. This indicates that, though relatively simple, the rotation-translation assumption for the align transformation made in our model is rational and effective in real scenarios. 

The visual results in Fig.~\ref{fig:real_1} further validate the effectiveness of our approach. Overall, our method keeps more fine-grained details and produces fewer unexpected artifacts compared with existing methods. For example, as shown in the first row, our result better keeps the morphology of characters and digital numbers, and in the last row, our method can better suppress artifacts while producing relatively sharper edges. More visual results on this dataset are provided in the supplementary material.

\begin{table}[t]
\centering
\vspace{-2mm}
\tabcolsep=3pt
\caption{Ablation Study with x4 SyntheticBurst Dataset\label{tab:ab}}
\vspace{-3mm}
\begin{tabular}{lccc} 
\toprule
Settings & PNSR & SSIM & Params.(M) \\ 
\midrule
(a) Align with RT & 42.97 & 0.972 & 9.0 \\ 
(b) Align with DConv & 42.81 & 0.970  & 11.5 \\ 
(c) w/o Eq-CNN & 42.76 &0.971& 10.1   \\ 
(d) w/o T-mat. & 42.80 & 0.972 &  8.7 \\ 
\hline
Ours & 43.18 & 0.974 & 8.7 \\ 
\bottomrule
\multicolumn{4}{l}{\footnotesize \textit{*RT: Restormer \cite{zamir2022restormer}}}  \\
\multicolumn{4}{l}{\footnotesize \textit{*DConv: Deformable convolution network \cite{dai2017deformable}}}  \\
\multicolumn{4}{l}{\footnotesize \textit{*w/o Eq-CNN: Replacing Eq-CNN with conventional CNN}}  \\
\multicolumn{4}{l}{\footnotesize \textit{*w/o T-mat: Removing the transformation matrix}}  \\
\vspace{-1cm}
\end{tabular}
\end{table}

 \subsection{Ablation Study\label{sec:ab}}


\begin{figure*}[tp]
    \centering    \includegraphics[width=0.95\linewidth]{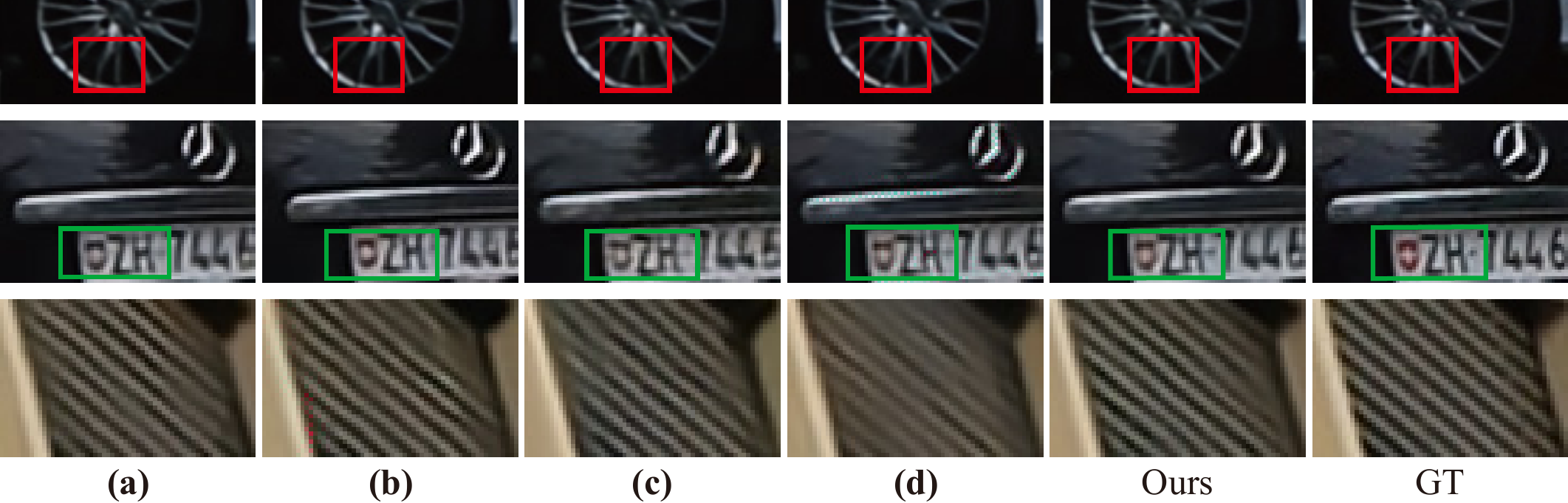}
    \vspace{-0.3cm}
    \caption{Visual results of the ablation study for x4 BISR on SyntheticBurst. The settings of (a)-(d) are referred to Table \ref{tab:ab} and Section \ref{sec:ab}.}
    \vspace{-0.4cm}
    \label{fig:ab_1}
\end{figure*}

\begin{figure}[tp]
    \centering
\includegraphics[width=\linewidth]{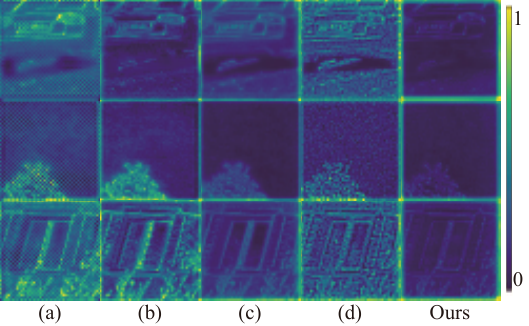}
    \vspace{-0.7cm}
    \caption{Error maps of aligned features for ablation studies on ×4 burst super-resolution using the SyntheticBurst dataset. Detailed settings of (a)-(d) can be referred to Table \ref{tab:ab} and Section \ref{sec:ab}.}
    \vspace{-0.8cm}
    \label{fig:ab_feat}
\end{figure}

In this subsection, we conduct experiments on the SyntheticBurst dataset at ×4 scale to validate the rationality and effectiveness of the proposed alignment framework in our model. 
The overall quantitative and visual results are summarized in Table \ref{tab:ab} and shown in Fig.~\ref{fig:ab_1}~-~Fig.~\ref{fig:ab_feat}, respectively. 

\noindent\textbf{Effectiveness of the overall alignment module (a) \& (b).} We first replace the whole alignment module in our method with implicit alignment strategies using Restormer (a) and deformable convolutions (b). 

As shown in Table~\ref{tab:ab}, both methods exhibit significant performance degradation, which can be more intuitively observed in the visual results illustrated in Fig.~\ref{fig:ab_1}~(a) and (b), that the fine-grained textures are not well kept. This can be attributed to the misalignment of features, as can be observed in Fig.~\ref{fig:ab_feat}~(a) and (b). These results clearly substantiate the effectiveness of our alignment module.

\begin{figure}[tp]
    \centering
    \vspace{-2mm}
\includegraphics[width=\linewidth]{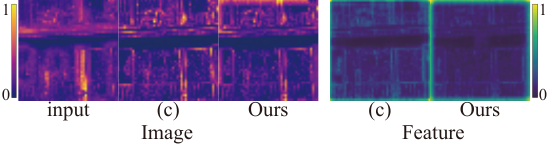}
\vspace{-0.6cm}
    \caption{Visual comparison of alignment error maps between CNN (c) Eq-CNN (Ours) in the image (left) and feature (right) domains, evaluated on the ×4 BISR task using the SyntheticBurst dataset. The error map is defined by measuring the pixel-wise difference between the reference and source frames.}
    \vspace{-0.8cm}
    \label{fig:edsr}
\end{figure}

\noindent\textbf{Effectiveness of equivariant feature extraction (c).} We then conduct an ablation study by replacing the ENet, which is an Eq-CNN, with a V-CNN without the rotation equivariance for feature extraction.
As shown in Table~\ref{tab:ab}~(c) and Fig.~\ref{fig:ab_1}~(c), this variant exhibits a noticeable performance drop compared to the proposed model in quantitative metrics and also produces blurry textures. The reason can be attributed to the lack of consistency of the alignment transformations between the image and feature domains, leading to mismatching among aligned features. This can be further substantiated by Fig.~\ref{fig:edsr}, within which we visualize the error maps of alignment results in both domains. It can be observed from this figure that, though the difference of image alignment is not very significant between the two variants, the Eq-CNN produces a much better alignment result than V-CNN in the feature domain. These results substantiate the reasonability and necessity of using Eq-CNN in our alignment framework.

Another interesting observation is that, though it does not perform well in the quantitative metrics, the visual results of this variant are comparable or even look better than that of other ablation variants as shown in Fig.~\ref{fig:ab_1}, and the alignment error in features is also significantly smaller than that of variants (a) and (b) as depicted in Fig.~\ref{fig:ab_feat}. This can be due to the explicit alignment mechanism using the learnable transformation and the translation-equivariance of V-CNNs, which indirectly suggests the effectiveness of our approach.


\noindent\textbf{Effectiveness of the learnable transformation matrix (d).} We then remove the transformation matrix, denoted as ``w/o T-mat.'' in Table~\ref{tab:ab}, and such a variant can be seen as implementing implicit alignment with the Eq-CNN. It can be observed from Fig.~\ref{fig:ab_feat} (d) that this leads to obvious feature misalignment and correspondingly inferior performance both in quantitative metrics and visual quality, highlighting the crucial role of explicit alignment.


\section{Conclusion and Discussion}
\label{sec:discussion}

In this work, we have proposed a new method for BISR. The key consideration of our method is the design of a new effective alignment framework for the BISR task with Eq-CNN. Within the proposed alignment framework, by the equivariance property of Eq-CNN, the align transformation can be learned with explicit image domain supervision and directly applied in the feature domain in a theoretically sound way. In addition, we have introduced effective upsampling and fusion blocks using advanced neural architectures, including MDTA from Restormer and INR. Extensive experiments on two representative BISR benchmarks have been conducted, showing the effectiveness of the proposed method, both quantitatively and visually, against current state-of-the-art methods.

Despite its promising performance for BISR, our method still has limitations that need further investigation. For example, currently, the transformation considered in our model is restricted to rotation and translation due to the ability of existing Eq-CNNs, which may not be precise enough to characterize the relationship between the reference and source frame in complex real-world scenarios. Tackling this issue requires developing new techniques and theories for equivariance networks, which could not only enhance the availability of our method in real applications but also advance the study of equivariance in deep learning.



{
    \small
    \bibliographystyle{ieeenat_fullname}
\bibliography{references}
}
\clearpage
\end{document}


\maketitle

\section{Theorem and Proofs}
In this section, we present a comprehensive version of Theorem 1 and Corollary 1, which was briefly introduced in the main text, along with the related lemmas and proofs, aiming to provide a solid theoretical foundation for our proposed method.

\subsection{Remark 1 and the Proof}
\textbf{Notations.}
For an input $r\in C^\infty(\mathbb{R}^2)$ and and transformations $\tilde{A}\in O(2)$ and $\tilde{b}\in \mathbb{R}^2$,  $\tilde{A}$ acts on $r$ by
\begin{equation}\label{fR}
	f^R_{\tilde{A}\tilde{b}}[r](x) = r(\tilde{A}^{-1}(x-\tilde{b})), \forall x\in\mathbb{R}^2.
\end{equation}
For a feature map $e\in C^\infty(E(2))$, $E(2)=\mathbb{R}^2 \ltimes O(2)$, and a transformation $\tilde{A}\in O(2)$, $\tilde{A}$ act on $e$ by
\begin{equation}\label{fE}
	f^E_{\tilde{A}\tilde{b}}[e](x,A,\tilde{b}) = e(\tilde{A}^{-1}(x-\tilde{b}),\tilde{A}^{-1}A), \forall (x,A)\in E(2).
\end{equation}
Let $\Psi$ denote the convolution on the input layer,
which maps an input $r\in C^\infty(\R^2)$ to a feature map defined on $E(2)$:
\begin{equation}\label{input_Conv}
	\Psi[r](x,A) =  \int_{\R^2}\varphi_{in}\left(A^{-1} \delta \right)r(x-\delta)d\sigma(\delta), ~\forall (x, A)\in E(2),
\end{equation}
where $\sigma$ is a measure on $\R^2$ and $\varphi$ is the proposed parameterized filter.
$\Phi$  denotes the convolution on the intermediate layer,
which maps a feature map $e\in C^\infty(E(2))$ to another feature map defined on $E(2)$:
\begin{equation}\label{regular_Conv}
	\begin{split}
		\Phi[e](x,\!B) \!\!=\!\!\!
		\int_{\!O(2)}\!\int_{\R^2}\!\!\varphi_{\!A}\!\left(B^{-\!1}\delta\right)\!e(x\!-\!\delta, \!B\!A)d\sigma(\delta)dv(A), ~\forall (x, B)\in E(2),
	\end{split}
\end{equation}
where $v$ is a measure on $O(2)$, $A,B\in O(2)$ denote orthogonal transformations in the considered group,
and $\varphi_{\tilde{A}}$ indicates the filter with respect to the channel of the feature map indexed by $\tilde{A}$, i.e., $\left.e(x,A)\right|_{A = \tilde{A}}$.
$\Upsilon$  denotes the convolution on the final layer,
which maps a feature map $e\in C^\infty(E(2))$ to a function defined on $\R^2$:
\begin{equation}\label{output_Conv}
	\Upsilon[e](x) \!=\!  \int_{O(2)}\int_{\R^2}\!\!\!\varphi_{out}\!\left(B^{-1}\delta\right)e(x\!-\!\delta, B)d\sigma(\delta)dv(B), ~\forall x\in \R^2.
\end{equation}

Then we will prove Remark 1.

\begin{Rem}
For $r\in C^\infty(\R^2)$, $e\in C^\infty(E(2))$ and $\tilde{A}\in O(2)$, the following results are satisfied:
\begin{equation}\label{Thm1}
	\begin{split}
		\Psi\left[\fR\left[r\right]\right] &= \fE\left[\Psi\left[r\right]\right],\\
		\Phi\left[\fE\left[e\right]\right] &= \fE\left[\Phi\left[e\right]\right],\\
		\Upsilon\left[\fE\left[e\right]\right] &= \fR\left[\Upsilon\left[e\right]\right],\\
	\end{split}
\end{equation}
where $\fR$, $\fE$, $\Psi$, $\Phi$ and $\Upsilon$
are defined by (\ref{fR}), (\ref{fE}), (\ref{input_Conv}), (\ref{regular_Conv}) and (\ref{output_Conv}), respectively.
\end{Rem}

\begin{proof}
(1) For any $x \in \R^2 $, $A\in O(2)$, and $\tilde{b}\in \mathbb{R}^2$ we can obtain
\begin{equation}\label{Psi}
	\begin{split}
		&\Psi\left[\fR\left[r\right]\right](x, A)\\
		= &\int_{\R^2}\varphi_{in}\left(A^{-1}\delta\right)\fR\left[r\right](x-\delta)d\sigma(\delta)\\
		= &\int_{\R^2}\varphi_{in}\left(A^{-1}\delta\right)r(\tilde{A}^{-1}(x-\delta-\tilde{b}))d\sigma(\delta).\\
	\end{split}
\end{equation}
Let $\hat{\delta} = \tilde{A}^{-1}\delta $, since $|det(\tilde{A})|=1$, and we have
\begin{equation}\label{Psi2}
	\begin{split}
		&\int_{\R^2}\varphi_{in}\left(A^{-1}\delta\right)r(\tilde{A}^{-1}(x-\delta-\tilde{b}))d\sigma(\delta) \\
		=  &\int_{\R^2}\varphi_{in}\left(A^{-1}\tilde{A}\hat{\delta}\right)r(\tilde{A}^{-1}(x-\tilde{b})-\hat{\delta}))d\sigma(\hat{\delta})\\
		=&\int_{\R^2}\varphi_{in}\left((\tilde{A}^{-1}A)^{-1}\hat{\delta}\right)r(\tilde{A}^{-1}(x-\tilde{b})-\hat{\delta}))d\sigma(\hat{\delta})\\
		=& \Psi[r](\tilde{A}^{-1}(x-\tilde{b}), \tilde{A}^{-1}A) \\
		=& \fE\left[\Psi[r]\right](x, A, \tilde{b}).
	\end{split}
\end{equation}
This proves that $\Psi\left[\fR\left[r\right]\right] = \fE\left[\Psi\left[r\right]\right]$.

(2) Similar to the proof in (1), for any $ x \in \R^2 $, $B\in O(2)$, we can obtain
\begin{equation}\label{Phi}
	\begin{split}
		&\Phi\left[\fE\left[e\right]\right](x, B)\\
		= &\int_{\R^2}\int_{O(2)}\varphi_A\left(B^{-1}\delta\right)\fE\left[e\right](x-\delta, BA, \tilde{b})d\sigma(\delta)v(A)\\
		= &\int_{\R^2}\int_{O(2)}\varphi_A\left(B^{-1}\delta\right)e(\tilde{A}^{-1}(x-\delta-\tilde{b}), \tilde{A}^{-1}BA)d\sigma(\delta)v(A)\\
		= &\int_{\R^2}\int_{O(2)}\varphi_A\left(B^{-1}\tilde{A}\hat{\delta}\right)e(\tilde{A}^{-1}(x-\tilde{b})-\hat{\delta}, \tilde{A}^{-1}BA)d\sigma(\hat{\delta})v(A)\\
		= &\int_{\R^2}\int_{O(2)}\varphi_A\left((\tilde{A}^{-1}B)^{-1}\hat{\delta}\right)e(\tilde{A}^{-1}(x-\tilde{b})-\hat{\delta}, \tilde{A}^{-1}BA)d\sigma(\hat{\delta})v(A)\\
		=& \Phi\left[ e \right](\tilde{A}^{-1}(x-\tilde{b}), \tilde{A}^{-1}B)\\
		=& \fE\left[\Phi\left[ e \right]\right](x, B, \tilde{b}).
	\end{split}
\end{equation}

(3) For any $x \in \R^2 $, we can deduce that
\begin{equation}\label{Upsilon}
	\begin{split}
		&\Upsilon  \left[\fE\left[e\right]\right](x)\\
		= &\int_{\R^2}\int_{O(2)}\varphi_{out}\left(B^{-1}\delta\right)\fE\left[e\right](x-\delta, B, \tilde{b})d\sigma(\delta)v(B)\\
		= &\int_{\R^2}\int_{O(2)}\varphi_{out}\left(B^{-1}\delta\right)e(\tilde{A}^{-1}(x-\delta-\tilde{b}), \tilde{A}^{-1}B)d\sigma(\delta)v(B)\\
		= &\int_{\R^2}\int_{O(2)}\varphi_{out}\left(\left(\tilde{A}^{-1}B\right)^{-1}\hat{\delta}\right)e(\tilde{A}^{-1}(x-\tilde{b})-\hat{\delta}, \tilde{A}^{-1}B)d\sigma(\hat{\delta})v(B).\\
		=&\Upsilon[e](\tilde{A}^{-1}(x-\tilde{b}))\\
		=&\fR\left[\Upsilon[e]\right](x).\\
	\end{split}
\end{equation}

This proves that $\Upsilon\left[\fE\left[e\right]\right] = \fR\left[\Upsilon\left[e\right]\right]$.
\end{proof}

\subsection{Remark 2 and the Proof}
\textbf{Notations.}
We assume that an image $I \in R^{n\times n}$ represents a two-dimensional grid function obtained by discretizing a smooth function, i.e., for $i, j = 1,2, \cdots, n$,
\begin{equation}\label{X}
	I_{ij} = r(\delta_{ij}),
\end{equation}
where $\delta_{ij} = \left(\left(i-\frac{n+1}{2}\right)h, \left(j-\frac{n+1}{2}\right)h\right)^T$. We represent $Z$ as a three-dimensional grid function sampled from a smooth function $e: \R^2\times S\to \R$, i.e., for $i, j = 1,2, \cdots, n$,
\begin{equation}\label{F_}
	Z_{ij}^{A,\tilde{b}} = e(\delta_{ij}, A, \tilde{b}),
\end{equation}
where $\delta_{ij} = \left(\left(i-\frac{n+1}{2}\right)h, \left(j-\frac{n+1}{2}\right)h\right)^T$ and $A\in S$, $S$ is a subgroup of $O(2)$, and $\tilde{b}\in \mathbb{R}^{2}$ is translation. For $i, j = 1,2, \cdots, p$, and $A,B\in S$, we have
\begin{equation}\label{Phi_}
	\begin{split}
		&\tilde{\Psi}_{ij}^A = \varphi_{in}\left(A^{-1}\delta_{ij}\right),\\
		&\tilde{\Phi}_{ij}^{B,A} = \varphi_A\left(B^{-1}\delta_{ij}\right),\\
		&\tilde{\Upsilon}_{ij}^{A} = \varphi_{out}\left(A^{-1}\delta_{ij}\right),
	\end{split}
\end{equation}
where $\delta_{ij} = \left(\left(i-\nicefrac{(p+1)}{2}\right)h, \left(j-\nicefrac{(p+1)}{2}\right)h\right)^T$, and $\varphi$ and $\varphi_A$ are parameterized filters.
Let
\begin{equation}\label{xy}
	\begin{split}
		\delta_{ij} &= \left(\left(i-\frac{p+1}{2}\right)h, \left(j-\frac{p+1}{2}\right)h\right)^T,\\
		x_{ij} &= \left(\left(i-\frac{n+p+2}{2}\right)h, \left(j-\frac{n+p+2}{2}\right)h\right)^T.
	\end{split}
\end{equation}
For $\forall A\in S$ and $i,j = 1,2,\cdots,n,$ the convolution of $\tilde{\Psi}$ and $I$ is
\begin{equation}\label{input_}
	\begin{split}
		\left(\tilde{\Psi}\star I\right)^{A}_{ij} &=  \sum_{(\ia,\ja)\in\Lambda}\varphi_{in}\left(A^{-1}\delta_{\ia\ja}\right)r\left(x_{ij}-\delta_{\ia\ja}\right),\\
	\end{split}
\end{equation}
where $\Lambda$ is a set of indexes, denoted as $ \Lambda = \{(i,j)|i,j=1,2,\cdots,p\}$.
For any $B\in S$ and $i,j = 1,2,\cdots,n,$ the convolution of $\tilde{\Phi}$ and $Z$ is
\begin{equation}\label{regular_}
	\begin{split}
		\left(\tilde{\Phi}\star Z\right)^{B}_{ij} &=  \!\!\!\sum_{(\ia,\ja)\in\Lambda, A\in S} \!\!\!\! \varphi_{A}\left(B^{-1}\delta_{\ia\ja}\right)e\!\left(x_{ij}-\delta_{\ia\ja}, BA\right), \\
	\end{split}
\end{equation}
where $\Lambda = \{(i,j)|i,j=1,2,\cdots,p\}$.
For $i,j = 1,2,\cdots,n,$ the convolution of $\tilde{\Upsilon}$ and $Z$ is
\begin{equation}\label{output_}
	\begin{split}
		\left(\tilde{\Upsilon}\star Z\right)_{ij} &=  \!\!\!\sum_{(\ia,\ja)\in\Lambda, B\in S}\!\!\!\!\varphi_{out}\left(B^{-1}\delta_{\ia\ja}\right)e\!\left(x_{ij}-\delta_{\ia\ja}, B\right)\\
	\end{split}
\end{equation}
where $\Lambda = \{(i,j)|i,j=1,2,\cdots,p\}$.

The transformations on $I$ and $Z$ are defined by
\begin{equation}\label{Pi_}
	\begin{split}
		& \left(\tilde{f}_{\tilde{A}\tilde{b}}^{R}(I)\right)_{ij} = f_{\tilde{A}\tilde{b}}^{R}[r](x_{ij}),  \left(\tilde{f}_{\tilde{A}\tilde{b}}^{\tilde{E}}(Z)\right)_{ij}^{A\tilde{b}} = f_{\tilde{A}\tilde{b}}^{E}[e](x_{ij},A,\tilde{b}),\\
		&\forall i,j = 1,2,\cdots,n, \forall A,\tilde{A} \in S.
	\end{split}
\end{equation}

Then we will prove the Theorem 2. We firstly introduce the following necessary lemma.
\begin{lemma}
	For smooth functions $r:\R^2\to\R$ and $\varphi: \R^2\to\R$,
	if for  $\delta\in \R^2$, the follow conditions are satisfied:
	\begin{equation}\label{conditionlemma}
		\begin{split}
			|r(\delta)| \leq F_1,&~|\varphi(\delta)|\leq F_2, \\
			\|\nabla r(\delta)\| \leq G_1,&~ \|\nabla \varphi(\delta)\|\leq G_2, \\
			\|\nabla^2 r(\delta)\| \leq H_1,&~ \|\nabla^2 \varphi(\delta) \|\leq H_2, \\
			\forall \|\delta\|\geq(\nicefrac{p+1}{2})h,&~ \varphi(\delta)=0
		\end{split}
	\end{equation}
	where $p, h>0$, $\nabla$ and ${\nabla}^2$ denote the operators of gradient and Hessian matrix, respectively,
	then, $\forall \A\in S, y\in\R$ the following results are satisfied:
\begin{equation}\label{Lemma33}
		\begin{split}
&\left|\int_{R^2}\varphi\lb(\A^{-1}\delta\rb)r\lb(x-\delta\rb)d\sigma(\delta)\right. - \left. \sum_{{i,j}\in\Lambda}\varphi\lb(\A^{-1}\delta_{ij}\rb)r\lb(x-\delta_{ij}\rb)h^2\right|\leq\frac{(p+1)^2C}{4}h^4,
		\end{split}
	\end{equation}
	where $\Lambda = \{(i,j)|i,j=1,2,\cdots,p\}$,  $\delta_{ij} = \left(\left(i-\nicefrac{(p+1)}{2}\right)h, \left(j-\nicefrac{(p+1)}{2}\right)h\right)^T$ and $C = F_1H_2+F_2H_1+2G_1G_2$.
\end{lemma}

The specific proof of lemma 1 can be referred to \cite{xie2022fourier}. Based on lemma 1, let us prove Remark 2.

\begin{Rem} Assume that an image $I\in \R^{n\times n}$ is discretized from the smooth function $r:\R^2\to\R$ by (\ref{X}), a feature map $Z\in \R^{n\times n \times t}$ is discretized from the smooth function $e:\R^2\times S\to\R$ by (\ref{F_}), $|S|=t$, and filters $\tilde{\Psi}$, $\tilde{\Phi}$ and $\tilde{\Upsilon}$ are generated from $\varphi_{in}$, $\varphi_{out}$ and $\varphi_{A}, \forall A\in S$, by (\ref{Phi_}), respectively. If for any $A\in S$, $x\in \R^2$, the following conditions are satisfied:
	\begin{equation}\label{condition_t2}
		\begin{split}
			&|r(x)| , |e(x,A)|\leq F_1,\\
			&\|\nabla r(x)\| , \|\nabla e(x,A) \|\leq G_1,\\
			&\|\nabla^2 r(x) \| , \|\nabla^2 e(x,A) \|\leq H_1,\\
			&|\varphi_{in}(x)|, |\varphi_{A}(x)|, |\varphi_{out}(x)|\leq F_2, \\
			&\|\nabla \varphi_{in}(x) \|, \|\nabla \varphi_{A}(x) \|, \|\nabla \varphi_{out}(x) \|\leq G_2, \\
			&\|\nabla^2 \varphi_{in}(x)\|, \|\nabla^2 \varphi_{A}(x) \| , \|\nabla^2 \varphi_{out}(x) \|\leq H_2, \\
			&\forall \|x\|\geq\nicefrac{(p+1)h}{2},~ \varphi_{in}(x), \varphi_{A}(x), \varphi_{out}(x)  =0,
		\end{split}
	\end{equation}
	where $p$ is the filter size, $h$ is the mesh size, and $\nabla$ and ${\nabla}^2$ denote the operators of gradient and Hessian matrix, respectively, then for any $\tilde{A}\in S$, the following results are satisfied:
	\begin{equation}\label{Thm2}
		\begin{split}
			&\left\|\tilde{\Psi}\star\fRd\left(I\right) - \fEd\left(\tilde{\Psi}\star I \right)\right\|_{\infty}
			\leq\frac{C}{2}(p+1)^2h^2  , \\
			&\left\|\tilde{\Phi}\star\fEd\left(Z\right) - \fEd\left(\tilde{\Phi}\star Z\right)\right\|_{\infty}
			\leq\frac{C}{2}(p+1)^2h^2t  ,\\
			&\left\|\tilde{\Upsilon}\star \fEd\left(Z\right) - \fRd\left(\tilde{\Upsilon}\star Z\right)\right\|_{\infty}
			\leq\frac{C}{2}(p+1)^2h^2t  ,
		\end{split}
	\end{equation}
	where $C = F_1H_2+F_2H_1+2G_1G_2$, $\fRd$, $\fEd$, $\tilde{\Psi}$, $\tilde{\Phi}$ and $\tilde{\Upsilon}$
	are defined by (\ref{Phi_}) and (\ref{Pi_}), respectively. The operators $\star$ involved in Eq. (\ref{Thm2}) are defined in (\ref{input_}), (\ref{regular_}) and (\ref{output_}), respectively, and $\|\cdot\|_{\infty}$ represents the infinity norm.
\end{Rem}

\begin{proof}
	For any $x\in \R, A,B\in S$, let
	\begin{equation}\label{Phi-hat}
		\hat{\Psi}[r](x,A) = \sum_{(\ia,\ja)\in\Lambda}\varphi_{in}\left(A^{-1}\delta_{\ia\ja}\right)r\left(x - \delta_{\ia\ja}\right),
	\end{equation}
	where $\Lambda = \{(\ia,\ja)|\ia,\ja=1,2,\cdots,p\}$. Then, for any $A\in S$, we can obtain
	\begin{equation}\label{4-1}
		\hat{\Psi}[r](x_{ij},A) = \lb(\tilde{\Psi}\star I\rb)_{ij}^{A}. 
	\end{equation}
	
	1) By \textbf{Remark 1}, we know that $ \Psi\left[\fR\left[r\right]\right] = \fE\left[\Psi\left[r\right]\right] $. Thus for any $A\in S$, we have
	\begin{equation}\label{4-2}
		\begin{split}
			&\left|\lb(\tilde{\Psi}\star\fRd\left(I\right) - \fEd\left(\tilde{\Psi}\star I \right)\rb)_{ij}^{A} \right|\\
			=&\left| \hat{\Psi}\lb[\fR[r]\rb](x_{ij},A) - \fE\lb[\hat{\Psi}[r]\rb](x_{ij},A)\right|\\
			\leq& \left| \hat{\Psi}\lb[\fR[r]\rb](x_{ij},A) - \frac{1}{h^2} \Psi\lb[\fR[r]\rb](x_{ij},A) \right|\\
			+& \left| \fE\lb[\hat{\Psi}[r]\rb](x_{ij},A) - \frac{1}{h^2}\fE\lb[\Psi[r]\rb](x_{ij},A) \right|.
		\end{split}
	\end{equation}
	Let $\hat{r} = \fR[r]$, and then it is easy to deduce that $\hat{r}$ satisfies the conditions in $\textbf{lemma 1}$. Then, by \textbf{lemma 1},
	\begin{equation}\label{4-3}
		\begin{split}
			& \left| \hat{\Psi}\lb[\fR[r]\rb](x_{ij},A) - \frac{1}{h^2} \Psi\lb[\fR[r]\rb](x_{ij},A) \right| \\
			=& \frac{1}{h^2}\left| \hat{\Psi}\lb[\fR[r]\rb](x_{ij},A)h^2 -  \Psi\lb[\fR[r]\rb](x_{ij},A) \right|\\
			=& \frac{1}{h^2}\left|\sum_{(i,j)\in\Lambda}\varphi_{in}\left(A^{-1}\delta_{ij}\right)\hat{r}\left(x_{ij} - \delta_{ij}\right)h^2-\int_{\R^2}\varphi_{in}\left(A^{-1}\delta\right)\hat{r}(x_{ij}-\delta)d\sigma(\delta) \right|\\
			\leq & \frac{(p+1)^2C}{4}h^2.
		\end{split}
	\end{equation}
	
	Besides, let $\hat{A} = \tilde{A}^{-1} A$ and $\hat{x}_{ij} = \tilde{A}^{-1}(x_{ij}-\tilde{b})$, and by \textbf{lemma 1}, we can also achieve,
	\begin{equation}\label{4-4}
		\begin{split}
			&\left| \fE\lb[\hat{\Psi}[r]\rb](x_{ij},A) - \frac{1}{h^2}\fE\lb[\Psi[r]\rb](x_{ij},A) \right|\\
			=& \frac{1}{h^2}\left| \fE\lb[\hat{\Psi}[r]\rb](x_{ij},A){h^2} -\fE\lb[\Psi[r]\rb](x_{ij},A) \right|\\
			=& \frac{1}{h^2}\left| \lb[\hat{\Psi}[r]\rb](\tilde{A}^{-1}(x_{ij}-\tilde{b}),\tilde{A}^{-1} A){h^2} -\lb[\Psi[r]\rb](\tilde{A}^{-1} (x_{ij}-\tilde{b}),\tilde{A}^{-1} A) \right|\\
			=& \frac{1}{h^2}\left|\sum_{(i,j)\in\Lambda}\varphi_{in}\left(A^{-1}\tilde{A}\delta_{ij}\right){r}\left(\tilde{A}^{-1}(x_{ij}-\tilde{b}) -\delta_{ij}\right)h^2-\int_{\R^2}\varphi_{in}\left(A^{-1}\tilde{A}\delta\right){r}(\tilde{A}^{-1}(x_{ij}-\tilde{b})-\delta)d\sigma(\delta) \right|\\
			=& \frac{1}{h^2}\left|\sum_{(i,j)\in\Lambda}\varphi_{in}\left(\hat{A}^{-1}\delta_{ij}\right){r}\left(\hat{x}_{ij} - \delta_{ij}\right)h^2-\int_{\R^2}\varphi_{in}\left(\hat{A}^{-1}\delta\right){r}(\hat{x}_{ij}-\delta)d\sigma(\delta) \right|\\
			\leq & \frac{(p+1)^2C}{4}h^2.
		\end{split}
	\end{equation}
	Thus, combining (\ref{4-2}), (\ref{4-3}) and (\ref{4-4}), we can achieve
	\begin{equation}\label{4-5}
		\begin{split}
			\left|  \hat{\Psi}\lb[\fR[r]\rb]\!(x_{ij},A,\tilde{b}) \!-\! \fE\lb[\hat{\Psi}[r]\rb]\!(x_{ij},A,\tilde{b})\right|\! \leq \! \frac{C}{2}(p+1)^2h^2.
		\end{split}
	\end{equation}
	In other word,
	\begin{equation}\label{4-5-2}
		\left|\lb(\tilde{\Psi}\star\fRd\left(I\right) - \fEd\left(\tilde{\Psi}\star I \right)\rb)_{ij}^{A} \right|
		\leq\frac{C}{2}(p+1)^2h^2 .
	\end{equation}
	This proves the first inequality in (\ref{Thm2}).
	
	2) For any $A,B\in S$, let $\hat{B} = \tilde{A}^{-1}B$, $r_A(x) = e(x, A)$, and  $\hat{\Psi}_A$ be a operator defined in the formulation of (\ref{Phi-hat}), while correlated to $\varphi_A$. Then, for any $i,j =1,2,\cdots,n$, $B\in S$,
	\begin{equation}\nonumber\label{4-6}
		\begin{split}
			&\left|\lb(\tilde{\Phi}\star\fEd\left(Z\right) - \fEd\left(\tilde{\Phi}\star Z\right)\rb)_{ij}^{B}\right|\\
			=&\left|\!\sum_{(\ia,\ja)\in\Lambda, A\in S}\!\!\!\!\!\!\varphi_{A}\!\left({B}^{-1}\delta_{\ia\ja}\right) e\left(\tilde{A}^{-1}\lb(x_{ij}\!-\delta_{\ia\ja}-\tilde{b}\rb), \tilde{A}^{-1}\!BA\right) -\!\!\!\!\!\!\!\! \sum_{(\ia,\ja)\in\Lambda, A\in S}\!\!\!\!\!\!\varphi_{A}\left(B^{-1}\tilde{A} \delta_{\ia\ja}\right) e\left(\tilde{A}^{-1}(x_{ij}-\tilde{b})\!-\!\delta_{\ia\ja}, \tilde{A}^{-1}\!BA\!\right) \right|\\
			\leq&\sum_{A\in S}\left|\sum_{(\ia,\ja)\in\Lambda}\!\!\!\varphi_{A}\left({B}^{-1}\delta_{\ia\ja}\right)r_{\hat{B}A}\left(\tilde{A}^{-1}(x_{ij}\!-\!\delta_{\ia\ja}-\tilde{b}\right) - \sum_{(\ia,\ja)\in\Lambda}\!\!\!\varphi_{A}\left(B^{-1}\tilde{A}  \delta_{\ia\ja}\right)r_{\hat{B}A}\left(\tilde{A}^{-1}(x_{ij}-\tilde{b})-\delta_{\ia\ja}\right) \right|\\
			=&\sum_{A\in S}\left|\sum_{(\ia,\ja)\in\Lambda}\!\!\!\varphi_{A}\left({B}^{-1}\delta_{\ia\ja}\right)\fR[r_{\hat{B}A}]\left(x_{ij}\!-\!\delta_{\ia\ja}\right) - \sum_{(\ia,\ja)\in\Lambda}\!\!\!\varphi_{A}\left({B}^{-1}\tilde{A}\delta_{\ia\ja}\right)r_{\hat{B}A}\left(\tilde{A}^{-1}(x_{ij}-\tilde{b})-\delta_{\ia\ja}\right) \right|\\
			=& \sum_{A\in S} \left|  \hat{\Psi}_A\lb[\fR[r_{\hat{B}A}]\rb](x_{ij},B,\tilde{b}) - \fE\lb[\hat{\Psi}_A[r_{\hat{B}A}]\rb](x_{ij},B,\tilde{b})  \right|.
		\end{split}
	\end{equation}
	Then by (\ref{4-5}), we can achieve that $\forall i,j =1,2,\cdots,n$, $B\in S$,
	\begin{equation}\label{Thm3-2}
		\left|\lb(\tilde{\Phi}\star\fEd\left(Z\right) - \fEd\left(\tilde{\Phi}\star Z\right)\rb)_{ij}^{B}\right|   \leq \frac{C}{2}(p+1)^2h^2t.
	\end{equation}
	This proves the second inequality in (\ref{Thm2}).
	
	3) For any $A,B\in S$, let $\hat{B} = \tilde{A}^{-1}B$, $r_A(x) = e(x, A)$, and $\hat{\Psi}_{out}$ be a operator defined in the formulation of (\ref{Phi-hat}),
	while correlated to $\varphi_{out}$.
	
	Then, we have that $\forall i,j =1,2,\cdots,n$,
	\begin{equation}\nonumber\label{4-7}
		\begin{split}
			&\left|\lb(\tilde{\Upsilon}\star\fEd\left(Z\right) - \fEd\left(\tilde{\Upsilon}\star Z\right)\rb)_{ij}\right|\\
			=&\left|\sum_{(\ia,\ja)\in\Lambda, B\in S}\!\!\!\!\varphi_{out}\left({B}^{-1}\delta_{\ia\ja}\right)e\left(\tilde{A}^{-1}\lb(x_{ij}\!-\!\delta_{\ia\ja}-\tilde{b}\rb), \tilde{A}^{-1}B\right)-\!\!\!\!\!\!\!\! \sum_{(\ia,\ja)\in\Lambda, B\in S}\!\!\!\!\!\varphi_{out}\left(B^{-1}\tilde{A}\delta_{\ia\ja}\right)\!e\left(\tilde{A}^{-1}({x}_{ij}-\tilde{b})\!-\delta_{\ia\ja}, \tilde{A}^{-1}B\right) \right|\\
			\leq&\sum_{B\in S}\left|\sum_{(\ia,\ja)\in\Lambda}\!\!\!\varphi_{out}\left({B}^{-1}\delta_{\ia\ja}\right)r_{\hat{B}}\left(\tilde{A}^{-1}\lb(x_{ij}\!-\!\delta_{\ia\ja}-\tilde{b}\rb)\right) -\!\!\! \sum_{(\ia,\ja)\in\Lambda}\!\!\!\varphi_{out}\left(B^{-1}\tilde{A}\delta_{\ia\ja}\right)r_{\hat{B}}\left(\tilde{A}^{-1}({x}_{ij}-\tilde{b})\!-\!\delta_{\ia\ja}\right) \right|\\
			=&\sum_{B\in S}\left|\sum_{(\ia,\ja)\in\Lambda}\!\!\!\varphi_{out}\left({B}^{-1}\delta_{\ia\ja}\right)\fR[r_{\hat{B}}]\left(x_{ij}\!-\!\delta_{\ia\ja}\right) - \sum_{(\ia,\ja)\in\Lambda}\!\!\!\varphi_{out}\left(B^{-1}\tilde{A}\delta_{\ia\ja}\right)r_{\hat{B}}\left(\tilde{A}^{-1}({x}_{ij}-\tilde{b})\!-\!\delta_{\ia\ja}\right) \right|\\
			=& \sum_{B\in S} \left|  \hat{\Psi}_{out}\lb[\fR[r_{\hat{B}}]\rb](x_{ij},B,\tilde{b}) - \fE \lb[\hat{\Psi}_{out}[r_{\hat{B}}]\rb](x_{ij},B,
            \tilde{b})  \right|.
		\end{split}
	\end{equation}
	\normalsize
	Then by (\ref{4-5}), we can achieve that $\forall i,j =1,2,\cdots,n$,
	\begin{equation}\label{Thm3-3}
		\left|\lb(\tilde{\Upsilon}\star\fEd\left(Z\right) - \fEd\left(\tilde{\Upsilon}\star Z\right)\rb)_{ij}\right|  \leq \frac{C}{2}(p+1)^2h^2t.
	\end{equation}
	This proves the third inequality in (\ref{Thm2}).
	
\end{proof}

\subsection{Theorem 1 and the Proof}
\textbf{Notations.}
In the following, we provide the corresponding formulations, just like \cite{xie2022fourier, fu2024rotation}.

For an input $r\in C^\infty(\mathbb{R}^2)$, translation $b\in \mathbb{R}^2$ and a degree $ \theta \in \lb[0, 2 \pi\rb]$, $A_{\theta}\in O(2)$ is the rotation matrix $\begin{bmatrix}
    \cos \theta, -\sin \theta \\
    \sin \theta, \cos \theta
\end{bmatrix}$. $A_{\theta}$ acts on $r$ by
\begin{equation}
	f^R_{\theta b}[r](x) = r(A_{\theta}^{-1}(x-b)), \forall x\in\mathbb{R}^2.
\end{equation}
For a feature map $e\in C^\infty(E(2))$, $E(2)=\mathbb{R}^2 \ltimes O(2)$, and a degree $\theta \in \lb[0, 2 \pi\rb]$.  $A_{\theta}$ acts on $e$ by
\begin{equation}
	f^E_{\theta b}[e](x,A,b) = e(A_{\theta}^{-1}(x-b),A_{\theta}^{-1}A), \forall (x,A,b)\in E(2).
\end{equation}

Considering an multi-channel image $I \in \mathbb{R}^{H\times W\times C}$ as input, which can be naturally represented by a two-dimensional grid function. Suppose the filter is of size $p\times p$, then, the mesh grids for filter and image can be respectively represented as follows:
\begin{equation}\label{xy}
	\begin{split}
		\delta_{kl} = \left(\left(k-\frac{p+1}{2}\right)h, \left(l-\frac{p+1}{2}\right)h\right)^T,~
		x_{kl} = \left(\left(k-\frac{H+1}{2}\right)h, \left(l-\frac{W+1}{2}\right)h\right)^T.
	\end{split}
\end{equation}
Then, each channel of $I$ can be obtained by discretizing a smooth function,
i.e., for $k= 1, 2, \cdots, W$, $l = 1,2, \cdots, H$, and $c = 1,2,\cdots, C$,
\begin{equation}\label{I}
	I_{kl}^c = r_c(x_{kl}),
\end{equation}
where  $r_c$ is latent function for  the $c^{th}$ channel.

We denote the equivariant number as $t$ and the correlated rotation group of the equivariant convolution as $S$, respectively. Then, $|S|=t$ and $S = \{A_\theta| \theta = \nicefrac{2\pi k}{t}, k = 1,2,\cdots, t \}$.
We represent the feature map of equivariant convolution as $Z\in\mathbb{R}^{H\times W\times t \times C}$. $Z$ is a four-dimensional grid function, whose $c^{th}$ channel is sampled from a smooth function $e_c: \R^2\times S\to \R$, i.e., for $k= 1, 2, \cdots, W$ and $l = 1,2, \cdots, H$,
\begin{equation}\label{F}
	Z_{kl}^{A,c} = e_c(x_{kl}, A),
\end{equation}
where  $A\in S$.

\textbf{Input layer.}
The filter of the input multi-channel convolution layer can be represented as
\begin{equation}\label{input_f}
	\tilde{\Psi}_{kl}^{A,c,d} = \varphi_{cd}\left(A^{-1}\delta_{kl}\right),
\end{equation}
where $\varphi_{cd}$ is the parameterized filter, $A\in S$, $c = 1,2,\cdots, n_{c}$, $d = 1,2,\cdots, n_{d}$, $n_{c}$ and $n_{d}$ are the input and output channel numbers, respectively.
Denoting multi-channel convolution of $\tilde{\Psi}$
and $I$ in the input layer as $Z = \hat\Psi(I)$, then it can be calculated by
\begin{equation}\label{input_d}
	\begin{split}
		\hat\Psi(I)^{A,d}=\sum_c\tilde{\Psi}^{A,c,d}*I^c,\\
	\end{split}
\end{equation}
where $*$ denotes the 2-D convolution operation.
It can be also rewritten in the following more detailed formulation:
\begin{equation}\label{input_d2}
	\begin{split}
		\hat{e}_d(x_{kl},A) =  \sum_{c,\delta\in\Lambda}\varphi_{cd}\left(A^{-1}\delta\right)r_c\left(x_{kl}-\delta\right),\\
	\end{split}
\end{equation}
where  $\Lambda$ is a set of indexes, denoted as $ \Lambda = \{\delta_{\hat k \hat l}|\hat k,\hat l=1,2,\cdots,p\}$, $A\in S$, $k = 1, 2, \cdots, W$ and $l = 1,2, \cdots, H$.

\textbf{Intermediate layer.}
The filter of the intermediate multi-channel convolution layer can be represented as
\begin{equation}\label{inter_f}
	\tilde{\Phi}_{kl}^{A,B,c,d} = \varphi_{Acd}\left(B^{-1}\delta_{kl}\right),
\end{equation}
where $\varphi_{Acd}$ is the parameterized filter, $A, B\in S$, $c = 1,2,\cdots, n_{c}$, $d = 1,2,\cdots, n_{d}$, $n_{c}$ and $n_{d}$ are the input and output channel numbers, respectively.
Denoting the multi-channel convolution of $\tilde{\Phi}$ and $Z$ in the intermediate layer as $\hat{Z} = \hat\Phi(Z)$, then it can be calculated by
\begin{equation}\label{regular_d}
	\begin{split}
		\hat\Phi(Z)^{B,d} = \sum_{c, A}\tilde{\Phi}^{A,B,c,d}*Z^{A,c}.\\
	\end{split}
\end{equation}
It can also be rewritten in the following more detailed formulation:
\begin{equation}\label{inter_d2}
	\begin{split}
		\hat{e}_d(x_{kl},B) =  \sum_{c, A, \delta\in\Lambda } \varphi_{Acd}\left(B^{-1}\delta\right)e_c\left(x_{kl}-\delta, BA\right).\\
	\end{split}
\end{equation}

\textbf{Output layer.}
The filter of the output multi-channel convolution layer can be represented as
\begin{equation}\label{output_f}
	\tilde\Upsilon_{kl}^{B,c,d} = \varphi_{cd}\left(B^{-1}\delta_{kl}\right),
\end{equation}
where $\varphi_{cd}$ is the parameterized filter, $B\in S$, $c = 1,2,\cdots, n_{c}$, $d = 1,2,\cdots, n_{d}$, $n_{c}$ and $n_{d}$ are the input and output channel numbers, respectively.
Denoting the multi-channel convolution of $\tilde{\Upsilon}$ and $Z$ in the output layer as $\hat Y = \hat\Upsilon(Z)$, then it can be calculated  by
\begin{equation}\label{output_d}
	\begin{split}
		\hat\Upsilon(Z)^{d} = \sum_{c,B} \tilde{\Upsilon}^{B,c,d} * Z^{B,c}.
	\end{split}
\end{equation}
It can be also rewritten in the following more detailed formulation:
\begin{equation}\label{output_d2}
	\begin{split}
		\hat{r}_d(x_{kl}) =   \sum_{c,B, \delta\in\Lambda} \varphi_{cd}\left(B^{-1}\delta\right)e_c\left(x_{kl}-\delta, B\right).\\
	\end{split}
\end{equation}

The transformations on each channel of the input image and the feature map are defined by
\begin{equation}\label{Pi_d}
	\begin{split}
		& \left(\fRdd(I)\right)^c_{kl} = f_{\theta b}^{R}[r_c](x_{kl}),~ \left(\fEdd(Z)\right)_{kl}^{A,c} = f_{\theta b}^{E}[e_c](x_{kl},A,b),\\
		&\forall k = 1,2,\cdots,H, l = 1,2,\cdots,W, c = 1,2,\cdots, C, \forall A \in S, \theta \in \lb[0, 2{\pi} \rb].
	\end{split}
\end{equation}
For expression conciseness we further denote
\begin{equation}
	\fdd[x] = \left\{\begin{array}{lll}
		\fRdd[x] & \mbox{if} &\forall x\in \mathbb{R}^{H\times W \times C}\\
		\fEdd[x] & \mbox{if} &\forall x\in \mathbb{R}^{H\times W \times t \times C}
	\end{array}\right..
\end{equation}
\textbf{Remark.} Following the \cite{fu2024rotation}, for a feature map $Z\in\mathbb{R}^{H\times W\times t \times C}$, we say the channel number of the correlated convolution layer is $tC$, due to the fact that $Z$ is usually reshaped into the shape of $H\times W \times tC$ for implementation convenience, and the flop of the correlated equivariant convolution layer is similar to a $tC$-channel convolution layer.

Then we will prove the Theorem 1. Before this, we first present the following necessary lemmas and the specific proof can be referred to \cite{fu2024rotation}.

{\begin{lemma}\label{lem_ei}
For an image ${I}$ with size $H\times W\times n_0$, and a $N$-layer rotation equivariant CNN network $\operatorname g(\cdot)$, whose channel number of the $i^{th}$ layer is $n_i$, rotation equivariant subgroup is $S\leqslant O(2)$, $|S|=t$, and activation function is set as ReLU. If the latent continuous function of the $c^{th}$ channel of ${I}$ denoted as $r_c: \mathbb{R}^2 \! \rightarrow \! \mathbb{R}$, and the latent continuous function of any convolution filters in the $i^{th}$ layer denoted as $\varphi^{i}: \mathbb{R}^2 \! \rightarrow \! \mathbb{R}$, where $i \in \{1, \cdots, N\}$, $c \in \{1, \cdots, n_{0}\}$, for any $x\in \R^2$, the following conditions are satisfied:
    \begin{equation}
        \begin{split}
            & |r_c(x)| \leq F_0, \|\nabla r_c(x)\| \leq G_0, \|\nabla ^2 r_c(x)\| \leq H_0, \\
            & |\varphi^{i}(x)| \leq F_i, \|\nabla \varphi^{i}(x)\| \leq G_i, \|\nabla ^2 \varphi^{i}(x)\| \leq H_i,\\
            &\forall \|x\|\geq\nicefrac{(p+1)h}{2},~ \varphi_{i}(x)=0,\\
        \end{split}
    \end{equation}
    where $p$ is the filter size, $h$ is the mesh size, and $\nabla$ and ${\nabla}^2$ denote the operators of gradient and Hessian matrix, respectively.
    Denote
    \begin{equation}\label{ei}
      e_d^i(x, B)= \lb\{\begin{array}{lcc}
                   \sum_{c, \delta\in\Lambda }   \varphi^1_{cd} (B^{-1} \delta) r_c(x - \delta) & \mbox{if} & i=1, \\
                   \sum_{c, A, \delta\in\Lambda }   \varphi^{i}_{Acd} (B^{-1} \delta) e^{i-1}_c(x - \delta, BA) & \mbox{if} & i\neq1, N
                                               \end{array}\rb.
    \end{equation}
    where  $ \Lambda = \lb\{\left(\left(k-\frac{p+1}{2}\right)h, \left(l-\frac{p+1}{2}\right)h\right)^T| k,l=1,2,\cdots,p\rb\}$, $\varphi^{1}_{cd}$ and $\varphi^{i}_{Acd}$ are filters in the first layer and other layers respectively. Then, for $\forall B\in S$ the following results are satisfied:
    \begin{equation}\label{l51}
    \lb|e_d^i(x, B)\rb|  \leq F_0 \mathcal{F}_i,  \\
    \end{equation}
    \begin{equation}\label{l52}
    \lb|\nabla e_d^i(x, B)\rb|  \leq \left(\sum_{m=1}^{i} \frac{G_m F_0}{F_m} + G_0 \right) \mathcal{F}_i, \\
    \end{equation}
    \begin{equation}\label{l53}
    \lb|\nabla^2 e_d^i(x, B)\rb|  \leq \left(\sum_{m=1}^{i} \frac{H_m F_0}{F_m} + 2 \sum_{l=1}^{i} \frac{G_l}{F_l} \sum_{m=1}^{l-1} \frac{G_m F_0}{F_m} + 2 \sum_{m=1}^{i} \frac{G_m G_0}{F_m} + H_0 \right) \mathcal{F}_i,
    \end{equation}
    where $\mathcal{F}_i = \prod \limits_{k=1}^i n_{k-1} p^2 F_k$, $\forall i=1,2,\cdots,N-1$.
\end{lemma}}

\begin{lemma} \label{theta_bound}
For an image ${I}$ with size $H\times W$, assume the latent continuous function of ${I}$ denoted as $r: \mathbb{R}^2 \! \rightarrow \! \mathbb{R}$ and the latent continuous function of the convolution filters denoted as $\varphi: \mathbb{R}^2 \! \rightarrow \! \mathbb{R}$.
If any $x\in \R^2$ the following conditions are satisfied $\forall \|x\|\geq\nicefrac{(p+1)h}{2},~\varphi(x)=0$, where $p$ is the filter size, $h$ is the mesh size. Additionally, we assume $ \|\nabla r(x) \| \leq G$ and $|\varphi(x)| \leq F$. For arbitrary rotation degree $ \theta\! \in \! \lb[0, 2\pi \rb]$, $A_{\theta} =  \begin{bmatrix}
    \cos\theta & -\sin\theta \\
    \sin\theta & \cos\theta \end{bmatrix}$ denotes the rotation matrix. If $f(\theta) = r\left(A_{\theta}^{-1} (x - \delta) \right)$, then the following result is satisfied:
\begin{equation}
    \left| \frac{\partial}{\partial \theta} f(\theta)\right| \leq \left( \operatorname{max}\{H,W\} + p+1 \right) h G.
\end{equation}
\end{lemma}

\begin{lemma}\label{derivative_F}
For an image ${I}$ with size $H\times W\times n_0$, and a $N$-layer rotation equivariant CNN network $\operatorname g(\cdot)$, whose channel number of the $i^{th}$ layer is $n_i$, rotation equivariant subgroup is $S\leqslant O(2)$, $|S|=t$, and activation function is set as ReLU. If the latent continuous function of the $c^{th}$ channel of ${I}$ denoted as $r_c: \mathbb{R}^2 \! \rightarrow \! \mathbb{R}$, and the latent continuous function of any convolution filters in the $i^{th}$ layer denoted as $\varphi^{i}: \mathbb{R}^2 \! \rightarrow \! \mathbb{R}$, where $i \in \{1, \cdots, N\}$, $c \in \{1, \cdots, n_{0}\}$, for any $x\in \R^2$, the following conditions are satisfied:
    \begin{equation}
        \begin{split}
            & |r_c(x)| \leq F_0, \|\nabla r_c(x)\| \leq G_0, \|\nabla ^2 r_c(x)\| \leq H_0, \\
            & |\varphi^{i}(x)| \leq F_i, \|\nabla \varphi^{i}(x)\| \leq G_i, \|\nabla ^2 \varphi^{i}(x)\| \leq H_i,\\
            &\forall \|x\|\geq\nicefrac{(p+1)h}{2},~ \varphi_{i}(x)=0,\\
        \end{split}
    \end{equation}
    where $p$ is the filter size, $h$ is the mesh size, $\nabla$ and ${\nabla}^2$ denote the operators of gradient and Hessian matrix, respectively. For an arbitrary $\theta \in \lb[0, 2 \pi \rb]$, $A_{\theta}$ denotes the rotation matrix. If $F(\theta)\!=\!\operatorname g\!\left[\fdd\!\right]\! ({I})\!\!=\!\!\hat{\Upsilon}\!\left[ \hat{\Phi}_{N-1}\!\cdots\!\hat{\Phi}_{i+1}\!\left[ \hat{\Phi}_{i}\! \cdots\!\hat{\Phi}_{2}\!\left[\hat{\Psi} \!\left[\!\fdd \right]\right]\!\cdots\!\right]\!\right]\!(I)$, then the following result is satisfied:
    
    \begin{equation}
    \begin{split}
        \left| F^{\prime} (\theta) \right| \leq \mathcal{F} \left( \operatorname{max}\{H,W\} + N \lb(p+1\rb) \right) h G_0, \\
    \end{split}
    \end{equation}
    where $\mathcal{F} = \prod \limits_{k=1}^N n_{k-1} p^2 F_k$.
\end{lemma}

\begin{lemma}\label{cor_derivative_F}
Under the same conditions with lemma \ref{derivative_F}, \\ If $F(\theta) = \fdd\left[\operatorname g\right]\!({I}) = \fdd\left[\hat{\Upsilon}\left[ \hat{\Phi}_{N-1} \cdots \hat{\Phi}_{i+1} \left[ \hat{\Phi}_{i} \cdots \hat{\Phi}_{2} \left[\hat{\Psi} \right]\cdots\right]\right]\right] (I)$, and then the following result is satisfied:
\begin{equation}
    \left| F^{\prime} (\theta) \right| \leq \mathcal{F} \operatorname{max}\{H,W\} h G_0, \\
\end{equation}
where $\mathcal{F} = \prod \limits_{k=1}^N n_{k-1} p^2 F_k$.
\end{lemma}

\begin{lemma}\label{Thm3}
For an image ${I}$ with size $H\times W\times n_0$, and a $N$-layer rotation-translation equivariant CNN network $\operatorname g(\cdot)$, whose channel number of the $i^{th}$ layer is $n_i$, rotation equivariant subgroup is $S\leqslant O(2)$, $|S|=t$, and activation function is set as ReLU. If the latent continuous function of the $c^{th}$ channel of ${I}$ denoted as $r_c: \mathbb{R}^2 \! \rightarrow \! \mathbb{R}$, and the latent continuous function of any convolution filters in the $i^{th}$ layer denoted as $\varphi^{i}: \mathbb{R}^2 \! \rightarrow \! \mathbb{R}$, where $i \in \{1, \cdots, N\}$, $c \in \{1, \cdots, n_{0}\}$, for any $x\in \R^2$, the following conditions are satisfied:
    \begin{equation}
        \begin{split}
            & |r_c(x)| \leq F_0, \|\nabla r_c(x)\| \leq G_0, \|\nabla ^2 r_c(x)\| \leq H_0, \\
            & |\varphi^{i}(x)| \leq F_i, \|\nabla \varphi^{i}(x)\| \leq G_i, \|\nabla ^2 \varphi^{i}(x)\| \leq H_i,\\
            &\forall \|x\|\geq\nicefrac{(p+1)h}{2},~ \varphi_{i}(x)=0,\\
        \end{split}
    \end{equation}
    where $p$ is the filter size, $h$ is the mesh size, $\nabla$ and ${\nabla}^2$ denote the operators of gradient and Hessian matrix, respectively. For an arbitrary $0 \leq \theta \leq 2 \pi$, $A_{\theta}$ denotes the rotation matrix, $b\in \mathbb{R}^2$ denotes the translation, and the following result is satisfied:
\begin{equation}\label{main_conclusion}
\begin{split}
  \lb|\fdd^{-1}\operatorname g\left[\fdd \right]\! ({I}) \!-\! \left[\operatorname g\right]\!({I})\rb|
\!\leq\! C_1 h^2\!+\! C_2 { p h}{t^{-1}},
\end{split}
\end{equation}
where
 \begin{equation}
    \begin{split}
        & C_1 = 2 N \mathcal{F} \cdot \sum_{i=1}^{N} \! \! \left( \! \frac{H_i F_0}{F_i} \! + \! 2 \frac{G_i}{F_i} \! \sum_{m=1}^{i-1} \! \frac{G_m F_0}{F_m} \! + \! 2 \frac{G_i G_0}{F_i} + H_0 \! \right), \\
        & C_2 = 2 \pi G_0 \mathcal{F} \lb(2 \operatorname{max}\{H,\!W\} p^{-1} + 2 N\rb), \mathcal{F} \!=\prod \nolimits_{i=1}^N n_{i-1} p^2 F_i.\\
    \end{split}
\end{equation}

\end{lemma}

\begin{proof}
Let $\hat{I} = \fdd I$, we can split the left part of Eq. (\ref{main_conclusion}) as
\begin{equation}\label{error_factoring}
    \begin{split}
    & \lb|\fdd^{-1}\operatorname g\left[\fdd \right]\! ({I}) \!-\! \left[\operatorname g \right]\!({I})\rb| \\
    = & \lb| \fdd^{-1}\operatorname g(\hat{I}) - \operatorname g \left[\fdd^{-1} \right](\hat{I}) \rb| \\
    \leq & \! \underbrace{ \left| \operatorname g\left[\fdd^{-1} \right]\! (\hat{I}) \! - \! \operatorname g\left[\tilde{f}_{\theta_{k}b}^{-1} \right]\! (\hat{I}) \right|}_{\langle 1 \rangle} \! + \! \underbrace{ \left| \operatorname g\left[\tilde{f}_{\theta_{k}b}^{-1} \right]\! (\hat{I}) \! - \! \tilde{f}_{\theta_{k}b}^{-1} \left[\operatorname g \right](\hat{I}) \right|}_{\langle 2 \rangle} \! + \! \underbrace{ \left| \tilde{f}_{\theta_{k}b}^{-1} \left[\operatorname g \right]\!(\hat{I}) \! - \! \tilde{f}_{\theta b}^{-1} \left[\operatorname g \right]\!(\hat{I})  \right|}_{\langle 3 \rangle},
    \end{split}
\end{equation}
where $\theta = \theta_{k} + \delta$, where $k = 1, 2, \cdots, t$, $0 \leq \delta \leq \nicefrac{2 \pi}{t}$. Next, we need to estimate the error bounds of the above three items, separately. It should be noted that the following proof is deduced without the ReLU activation function for concise. However, the conclusions are all still correct for networks with the ReLU activation function,  since ReLU does not disturb the equivariance or amplify the error bound.

Firstly, we prove the following inequality for the part $\langle 1 \rangle$ of Eq. (\ref{error_factoring}).
\begin{equation}\label{error_continue_to_seperate_01}
    \left| \operatorname g\left[\fdd^{-1} \right]\! ({\hat{I}}) - \operatorname g\left[\fdk^{-1} \right]\! (\hat{I}) \right| \leq \frac{2 \pi}{t} \mathcal{F} \left(\operatorname{max}\left\{H,W\right\} + N \lb(p+1\rb) \right) h G_0.
\end{equation}
Let us denote $F_1(\theta) = \operatorname g\left[\fdd \right]\! (\hat{I})$. Obviously the function $F_1(\theta)$ is continuous with respect to $\theta$, so we have the following conclusion by the Lagrange Mean Value Theorem \cite{sahoo1998mean}
\begin{equation}
\begin{split}
\left| \operatorname g\left[\fdd \right]\! (\hat{I}) - \operatorname g\left[\fdk \right]\! (\hat{I}) \right| & = \lb| F_1(\theta) - F_1(\theta_k) \rb| \\
& \leq \lb|F_1^{\prime}(\xi_1)\rb| \delta \\
& \leq \frac{2 \pi}{t} \lb|F_1^{\prime}(\xi_1)\rb|,
\end{split}
\end{equation}
where $0 < \xi_1 < \delta$ and by lemma \ref{derivative_F} we have $\left|F_1^{\prime} (\xi_1) \right| \leq \mathcal{F}\left(\operatorname{max}\left\{H, W\right\} + N \lb(p+1\rb) \right) h G_0$. Then we can prove Eq. (\ref{error_continue_to_seperate_01}).

Secondly, we prove the following inequality for the part $\langle 3 \rangle$ of Eq. (\ref{error_factoring}).
\begin{equation}\label{error_continue_to_seperate_02}
    \left| \fdk \left[\operatorname g \right](\hat{I}) - \fdd \left[\operatorname g \right](\hat{I})  \right| \leq \frac{2 \pi}{t} \mathcal{F} \operatorname{max}\left\{H,W\right\} h G_0.
\end{equation}
Let us denote $F_2(\theta) = \fdd \left[\operatorname g \right](\hat{I})$. Obviously the function $F_2(\theta)$ is continuous with respect to $\theta$, so we have the following conclusion by the Lagrange Mean Value Theorem \cite{sahoo1998mean}
    \begin{equation}
    \begin{split}
    \left| \fdk \left[\operatorname g \right](\hat{I})  - \fdd \left[\operatorname g \right](\hat{I})  \right| & = \lb|F_2(\theta) - F_2(\theta_k)\rb|\\
    & \leq \lb|F_2^{\prime}(\xi_2)\rb| \delta \\
    & \leq \frac{2 \pi}{t} \lb|F_2^{\prime}(\xi_2)\rb|,
    \end{split}
    \end{equation}
where $0 < \xi_2 < \delta$ and by lemma \ref{cor_derivative_F} we have $\left|F_2^{\prime} (\xi_2) \right| \leq \mathcal{F} \operatorname{max}\left\{H,W\right\} h G_0$. Then we can easily achieve Eq. (\ref{error_continue_to_seperate_02}).

Thirdly, we now prove the following inequality:
\begin{equation}\label{error_seperate}
    \begin{split}
    \underbrace{ \left| \operatorname g\left[\tilde{f}_{\theta_{k}b}^{-1} \right]\! (\hat{I}) \! - \! \tilde{f}_{\theta_{k}b}^{-1} \left[\operatorname g \right](\hat{I}) \right|}_{\langle 2 \rangle}
    \leq 2 \mathcal{F} \sum_{i=1}^{N} \left(\sum_{m=1}^{i} \frac{H_m F_0}{F_m} + 2 \sum_{l=1}^{i} \frac{G_l}{F_l} \sum_{m=1}^{l-1} \frac{G_m F_0}{F_m} + 2 \sum_{m=1}^{i} \frac{G_m G_0}{F_m} + H_0 \right) h^2 .
    \end{split}
\end{equation}
$\operatorname g(\cdot)$,  an N-layer rotation equivariant CNN network,  usually includes 1 input layer, $N-2$ intermediate layers, and 1 output layer.  We can formally define it as :
\begin{equation}
\operatorname g(\cdot) =  \hat{\Upsilon}\left[ \hat{\Phi}_{N-1} \cdots \hat{\Phi}_{i+1} \left[ \hat{\Phi}_{i} \cdots \hat{\Phi}_{2} \left[\hat{\Psi} \right]\cdots\right]\right] (\cdot).
\end{equation}
Then we have
\begin{equation}\label{hole_error_seperate}
    \begin{split}
    & \left| \operatorname g\left[\fdk^{-1} \right] (\hat{I}) -
    \fdk^{-1}\left[\operatorname g\right] (\hat{I}) \right| \\
    = & \lb| \hat{\Upsilon}\left[ \hat{\Phi}_{N-1} \cdots \hat{\Phi}_{i+1} \left[ \hat{\Phi}_{i} \cdots \hat{\Phi}_{2} \left[\hat{\Psi} \left[\fdk^{-1}\right]\right]\cdots\right]\right] (\hat{I}) - \fdk^{-1}\left[\hat{\Upsilon}\left[ \hat{\Phi}_{N-1} \cdots \hat{\Phi}_{i+1} \left[ \hat{\Phi}_{i} \cdots \hat{\Phi}_{2} \left[\hat{\Psi} \right]\cdots\right]\right]\right] (\hat{I}) \rb| \\
    \leq & \lb| \hat{\Upsilon}\left[ \hat{\Phi}_{N-1} \cdots \hat{\Phi}_{i+1} \left[ \hat{\Phi}_{i} \cdots \hat{\Phi}_{2} \left[\hat{\Psi} \left[\fdk^{-1}\right]\right]\cdots\right]\right] (\hat{I}) - \hat{\Upsilon}\left[ \hat{\Phi}_{N-1} \cdots \hat{\Phi}_{i+1} \left[ \hat{\Phi}_{i} \cdots \hat{\Phi}_{2} \left[\fdk^{-1}\left[\hat{\Psi} \right]\right]\cdots\right]\right] (\hat{I}) \rb| \\
    & + \lb| \hat{\Upsilon}\left[ \hat{\Phi}_{N-1} \cdots \hat{\Phi}_{i+1} \left[ \hat{\Phi}_{i} \cdots \hat{\Phi}_{2} \left[\fdk^{-1}\left[\hat{\Psi} \right]\right]\cdots\right]\right] (\hat{I}) - \hat{\Upsilon}\left[ \hat{\Phi}_{N-1} \cdots \hat{\Phi}_{i+1} \left[ \hat{\Phi}_{i} \cdots \fdk^{-1}\left[\hat{\Phi}_{2}\left[\hat{\Psi} \right]\right]\cdots\right]\right] (\hat{I}) \rb| \\
    & ~~~~~~~~~~~~~~~~~~~~~~~~~~~~~~~~~~~~~~~~~~~~~~~~~~~~~~~~~~~~~~~~~~~~~~~~~ \cdots \\
    & + \lb| \hat{\Upsilon}\left[ \hat{\Phi}_{N-1} \left[ \fdk^{-1} \cdots \hat{\Phi}_{i+1} \left[ \hat{\Phi}_{i} \cdots \hat{\Phi}_{2}\left[\hat{\Psi} \right]\right]\cdots\right]\right] (\hat{I}) - \hat{\Upsilon}\left[ \fdk^{-1} \left[ \hat{\Phi}_{N-1} \cdots \hat{\Phi}_{i+1} \left[ \hat{\Phi}_{i} \cdots \hat{\Phi}_{2}\left[\hat{\Psi} \right]\right]\cdots\right]\right] (\hat{I}) \rb| \\
    & + \lb| \hat{\Upsilon}\left[ \fdk^{-1} \left[ \hat{\Phi}_{N-1} \cdots \hat{\Phi}_{i+1} \left[ \hat{\Phi}_{i} \cdots \hat{\Phi}_{2}\left[\hat{\Psi} \right]\right]\cdots\right]\right] (\hat{I}) - \fdk^{-1} \left[ \hat{\Upsilon}\left[ \hat{\Phi}_{N-1} \cdots \hat{\Phi}_{i+1} \left[ \hat{\Phi}_{i} \cdots \hat{\Phi}_{2}\left[\hat{\Psi} \right]\right]\cdots\right]\right] (\hat{I}) \rb|.
    \end{split}
\end{equation}
We denote $\delta_1$, $\delta_i$ $(i=2, 3, \cdots, N-1)$, and $\delta_N$ as the filter indexes of the input layer, $i^{th}$ intermediate layer and output layer, respectively.  
The input channel number of the $i^{th}$ layer is set as $c_{i-1}=n_{i-1}$. 

1) For the input layer, with Eqs. (\ref{input_d2}), (\ref{inter_d2}) and (\ref{output_d2}), let $x$ denote the coordinate of position $(k, l)$, then we have

\begin{equation}\label{first_layer_inter}
    \begin{split}
    & \lb| \lb( \hat{\Upsilon}\left[ \hat{\Phi}_{N-1} \cdots \hat{\Phi}_{i+1} \left[ \hat{\Phi}_{i} \cdots \hat{\Phi}_{2} \left[\hat{\Psi} \left[\tilde{f}_{\theta_{k}b}^{-1} \right]\right]\cdots\right]\right] (\hat{I}) - \hat{\Upsilon}\left[ \hat{\Phi}_{N-1} \cdots \hat{\Phi}_{i+1} \left[ \hat{\Phi}_{i} \cdots \hat{\Phi}_{2} \left[ \tilde{f}_{\theta_{k}b}^{-1} \left[\hat{\Psi} \right] \right] \cdots \right]\right] (\hat{I}) \rb)_{ij}^{c_N} \rb| \\
    = & \left| \sum_{\substack{c_{N-1} \\ B_{N-1} \in S \\ \delta_{N} \in \Lambda}}  \cdots \sum_{\substack{c_1 \\ A \in S \\ \delta_{2} \in \Lambda}} \sum_{\substack{c_0 \\ \delta_{1} \in \Lambda}} \varphi^N_{c_{N-1} c_{N}}(B^{-1}_{N-1} \delta_{N}) \cdots \varphi^1_{c_0c_1}\left( A^{-1} \delta_{1}\right)  r_{c_0} \left(A_{\theta_{k}} (x - \delta_{N}  - \cdots - \delta_{2} - \delta_{1} + b)\right)\right. \\
    & - \left. \sum_{\substack{c_{N-1} \\ B_{N-1} \in S \\ \delta_{N} \in \Lambda}}  \cdots \sum_{\substack{c_1 \\ A \in S \\ \delta_{2} \in \Lambda}} \sum_{\substack{c_0 \\ \delta_{1} \in \Lambda}} \varphi^N_{c_{N-1} c_{N}}(B^{-1}_{N-1} \delta_{N}) \cdots \varphi^1_{c_0c_1}\left( A^{-1}A_{\theta_{k}}^{-1} \delta_{1}\right)  r_{c_0} \left(A_{\theta_{k}} (x - \delta_{N} - \cdots - \delta_{2}+b) - \delta_{1} \right) \right| \\
    = & \lb| \sum_{\substack{c_{N-1} \\ B_{N-1} \in S \\ \delta_{N} \in \Lambda}}  \cdots \sum_{\substack{c_1 \\ A \in S \\ \delta_{2} \in \Lambda}} \varphi^{N}_{c_{N-1} c_{N}}(B^{-1}_{N-1}\delta_{N}) \cdots \varphi^2_{A c_1 c_2}(B^{-1}_2 \delta_{2}) \rb. \\
    & \lb.\left(  \sum_{\substack{c_0 \\ \delta_{1} \! \in \! \Lambda}} \! \varphi^1_{c_0 c_1} \! \left(A^{-1} \delta_{1} \right) \! r \! \left(\!A_{\theta_{k}} (x\! - \! \delta_{N} \! -\! \cdots \! \delta_{1}+b )\!\right) \! - \! \sum_{\substack{c_0 \\ \delta_{1} \! \in \! \Lambda}} \! \varphi^1_{c_0 c_1} \! \left(A^{-1} A_{\theta_{k}}^{-1} \delta_{1}\right) r_{c_0} \! \left(A_{\theta_k} (x \! - \! \delta_{N} \! - \! \cdots  \! \delta_{2}+b) \! - \! \delta_{1} \! \right) \!  \right) \rb|\\
    \leq & \sum_{\substack{c_{N-1} \\ B_{N-1} \in S \\ \delta_{N} \in \Lambda}}  \cdots \sum_{\substack{c_1 \\ A \in S \\ \delta_{2} \in \Lambda}} \lb| \varphi^{N}_{c_{N-1} c_{N}}(B^{-1}_{N-1} \delta_{N}) \rb|  \cdots \lb| \varphi^2_{A c_1 c_2}(B^{-1}_2 \delta_{2}) \rb| \\
    & \lb| \sum_{\substack{c_0 \\ \delta_{1} \! \in \! \Lambda}} \! \varphi^1_{c_0 c_1} \! \left(A^{-1} \delta_{1} \right) \! r \! \left(\!A_{\theta_k} (x\! - \! \delta_{N} \! -\! \cdots \! - \! \delta_{1}+b\!)\!\right) \! - \! \sum_{\substack{c_0 \\ \delta_{1} \! \in \! \Lambda}} \! \varphi^1_{c_0 c_1} \! \left(A^{-1} A_{\theta_{k}}^{-1} \delta_{1}\right) r_{c_0} \! \left(A_{\theta_k} (x \! - \! \delta_{N} \! - \! \cdots \! - \! \delta_{2}+b) \! - \! \delta_{1} \! \right) \! \rb| \\
    \leq & \sum_{\substack{c_{N-1} \\ B_{N-1} \in S \\ \delta_{N} \in \Lambda}}  \cdots \sum_{\substack{c_1 \\ A \in S \\ \delta_{2} \in \Lambda}}\sum_{c_0} F_N  \cdots F_2 \\
    & \lb| \sum_{\delta_{1} \! \in \! \Lambda} \! \varphi^1_{c_0 c_1} \! \left(A^{-1} \delta_{1} \right) \! r \! \left(\!A_{\theta_k}(x\! - \! \delta_{N} \!  \cdots  - \! \delta_{1}+b \!)\!\right) \! - \! \sum_{\delta_{1} \! \in \! \Lambda} \! \varphi^1_{c_0 c_1} \! \left(A^{-1}A_{\theta_{k}}^{-1} \delta_{1}\right) r_{c_0} \! \left(A_{\theta_k} (x \! - \! \delta_{N} \!- \! \cdots \! - \! \delta_{2}+b) \! - \! \delta_{1} \! \right) \! \rb| \! .
    \end{split}
\end{equation}
Let us denote $\hat{x} = x - \delta_{N} - \delta_{N-1} - \cdots - \delta_{2} + b$. Utilizing Eq. (\ref{Thm2}) from Remark 2 for the input layer, we can deduce the following result:
\begin{equation}\label{l724}
    \begin{split}
    & \lb| \sum_{\delta_{1} \! \in \! \Lambda} \! \varphi^1_{c_0 c_1} \! \left(A^{-1} \delta_{1} \right) \! r \! \left(\!A_{\theta_k}(x\! - \! \delta_{N} \!  \cdots  - \! \delta_{1} +b)\!\right) \! - \! \sum_{\delta_{1} \! \in \! \Lambda} \! \varphi^1_{c_0 c_1} \! \left(A^{-1}A_{\theta_{k}}^{-1} \delta_{1}\right) r_{c_0} \! \left(A_{\theta_k} (x \! - \! \delta_{N} \!- \! \cdots \! - \! \delta_{2}+b) \! - \! \delta_{1} \! \right) \! \rb|  \\
    = &\lb| \sum_{\delta_1 \in \Lambda} \varphi^1_{c_0 c_1} \! \left(A^{-1} \delta_{1} \right) r \! \left(\!A_{\theta_k} (\hat{x} - \! \delta_{1} \!)\!\right) \! - \! \sum_{\delta_1 \in \Lambda} \varphi^1_{c_0 c_1} \! \left( A^{-1} A_{\theta_{k}}^{-1} \delta_{1}\right) \! r_{c_0} \! \left(A_{\theta_k} \hat{x} \! - \! \delta_{1} \right) \rb| \\
    \leq & n_0 \frac{C_1}{2} (p+1)^2 h^2 ,\\
    \end{split}
\end{equation}
where we do not specifically indicate the numbers of input and output channels, i.e. $\varphi^1(x)=\varphi^1_{c_0 c_1}(x)$, and we have \\$C_1 = H_1 F_0 + F_1 H_0 + 2 G_1 G_0$. 
Therefore, according to Eqs. (\ref{first_layer_inter}) and (\ref{l724}), we have:
\begin{equation}\label{1_layer}
    \begin{split}
    & \lb| \lb( \hat{\Upsilon}\left[ \hat{\Phi}_{N-1} \cdots \hat{\Phi}_{i+1} \left[ \hat{\Phi}_{i} \cdots \hat{\Phi}_{2} \left[\hat{\Psi} \left[\fdk^{-1} \right]\right]\cdots\right]\right] (\hat{I}) - \hat{\Upsilon}\left[ \hat{\Phi}_{N-1} \cdots \hat{\Phi}_{i+1} \left[ \hat{\Phi}_{i} \cdots \hat{\Phi}_{2} \left[ \fdk^{-1} \left[\hat{\Psi} \right] \right] \cdots \right]\right] (\hat{I}) \rb)_{ij}^{c_N} \rb| \\
    \leq & n_{N-1} p^2 F_N n_{N-2} p^2 F_{N-1} \cdots n_{1} p^2 F_{2} n_0 \frac{C_1}{2} (p+1)^2 h^2 \\
     \leq   &  \left(\prod \limits_{k=2}^{N} n_{k-1} p^2 F_k \right) n_0 \frac{(p+1)^2 h^2}{2} \left(H_1 F_0 + F_1 H_0 + 2 G_1 G_0\right) \\
     \leq&  2 \mathcal{F} \left(\frac{H_1}{F_1} F_0 + H_0 + 2\frac{G_1}{F_1} G_0\right) h^2.
    \end{split}
\end{equation}

2) For the any $i^{th}$ intermediate layer, $1<i<N$. We have:
\begin{equation}\label{l766}
    \begin{split}
    & \lb| \lb( \hat{\Upsilon}\left[ \hat{\Phi}_{N-1} \cdots \hat{\Phi}_{i} \left[ \fdk^{-1} \left[ \hat{\Phi}_{i-1} \cdots \hat{\Phi}_{2} \left[\hat{\Psi} \right]\cdots\right]\right]\right] (\hat{I}) - \hat{\Upsilon}\left[ \hat{\Phi}_{N-1} \cdots \hat{\Phi}_{i+1} \left[ \fdk^{-1} \left[ \hat{\Phi}_{i} \cdots \hat{\Phi}_{2} \left[\hat{\Psi} \right]\cdots\right]\right]\right] (\hat{I}) \rb)_{ij}^{c_N} \rb| \\
    = & \left| \sum_{\substack{c_{N-1} \\ B_{N-1} \in S \\ \delta_{N} \in \Lambda}} \! \cdots \sum_{\substack{c_{i-1} \\ B_{i-1} \in S \\ \delta_{i} \in \Lambda}} \! \varphi^{N}_{c_{N-1}c_{N}} \! (B^{-1}_{N-1} \delta_{N}) \! \cdots \varphi^{i}_{B_{i-1}c_{i-1}c_{i}} \! (B^{-1}_{i} \delta_{i}) \right. e^{i-1}_{c_{i-1}} \!\left( \! A_{\theta_k} \! (x \! - \! \delta_{N} \! - \! \cdots \! - \! \delta_{i+1} \! - \! \delta_{i}+b), A_{\theta_k} B_{i-1}\right) \\
    - & \! \sum_{\substack{c_{N-1} \\ B_{N-1} \in S \\ \delta_{N} \in \Lambda}} \cdots \sum_{\substack{c_{i-1} \\ B_{i-1} \in S \\ \delta_{i} \in \Lambda}} \! \varphi^{N}_{c_{N-1}c_{N}} \! (B^{-1}_{N-1} \delta_{N}) \cdots \varphi^{i}_{B_{i-1}c_{i-1}c_{i}} \! (B^{-1}_{i} A_{\theta_{k}}^{-1} \delta_{i}) \left. e^{i-1}_{c_{i-1}} \!\left(\!A_{\theta_k} \! (x - \delta_{N} - \cdots - \delta_{i+1}+b) -\delta_{i}, B_{i-1}\right) \right|\\
    \leq & \left| \sum_{\substack{c_{N-1} \\ B_{N-1} \in S \\ \delta_{N} \in \Lambda}} \cdots \sum_{\substack{c_{i} \\ B_{i} \in S \\ \delta_{i+1} \in \Lambda}} \! \varphi^{N}_{c_{N-1}c_{N}} \! (B^{-1}_{N-1} \delta_{N}) \varphi^{N-1}_{B_{N-2} c_{N-2}c_{N-1}} \!(B^{-1}_{N-1} \delta_{N-1}) \! \cdots \varphi^{i+1}_{B_i c_{i}c_{i+1}} \! (B^{-1}_{i+1} \delta_{i+1}) \right.\\
    & \left( \sum_{\substack{c_{i-1} \\ B_{i-1} \in S \\ \delta_{i} \in \Lambda}} \varphi^{i}_{B_{i-1}c_{i-1}c_{i}} \! (B^{-1}_{i} \delta_{i}) e^{i-1}_{c_{i-1}} \!( \! A_{\theta_k} \! (x \! - \! \delta_{N} \! - \! \cdots \! - \! \delta_{i+1} \! - \! \delta_{i} +b), A_{\theta_k}B_{i-1}) \right.\\
    - & \left.\left. \sum_{\substack{c_{i-1} \\ B_{i-1} \in S \\ \delta_{i} \in \Lambda}} \varphi^{i}_{B_{i-1}c_{i-1}c_{i}} \! (B^{-1}_{i} A_{\theta_{k}}^{-1} \delta_{i}) e^{i-1}_{c_{i-1}} \!\left(\!A_{\theta_k} \! (x - \delta_{N} - \cdots - \delta_{i+1}+b) - \delta_{i}, B_{i-1}\right) \right)\right|\\
    \leq & \sum_{\substack{c_{N-1} \\ B_{N-1} \in S \\ \delta_{N} \in \Lambda}} \sum_{\substack{c_{N-2} \\ B_{N-2} \in S \\ \delta_{N-1} \in \Lambda}} \cdots \sum_{\substack{c_{i} \\ B_{i} \in S \\ \delta_{i+1} \in \Lambda}} F_{N} F_{N-1} \cdots F_{i+1} \\
    &  \left| \left( \sum_{\substack{c_{i-1} \\ B_{i-1} \in S \\ \delta_{i} \in \Lambda}} \varphi^{i}_{B_{i-1}c_{i-1}c_{i}} \! (B^{-1}_{i} \delta_{i}) e^{i-1}_{c_{i-1}} \!( \! A_{\theta_k} \! (x \! - \! \delta_{N} \! - \! \cdots \! - \! \delta_{i+1} \! - \! \delta_{i}+b), A_{\theta_k}B_{i-1}) \right.\right.\\
    & - \left.\left. \sum_{\substack{c_{i-1} \\ B_{i-1} \in S \\ \delta_{i} \in \Lambda}} \varphi^{i}_{B_{i-1}c_{i-1}c_{i}} \! (B^{-1}_{i} A_{\theta_{k}}^{-1} \delta_{i}) e^{i-1}_{c_{i-1}} \!\left(\!A_{\theta_k} \! (x - \delta_{N} - \cdots - \delta_{i+1}+b) - \delta_{i}, B_{i-1}\right) \right) \right|.
    \end{split}
\end{equation}
Let us denote $\hat{x} = x - \delta_{N} - \cdots - \delta_{i+1}+b$. Then, by Eq. (\ref{Thm2}) in Remark 2 for the intermediate layer, we have:
\begin{equation}\label{l782}
    \begin{split}
    & \left|\!\sum_{\substack{c_{i-1} \\ B_{i-1} \in S \\ \delta_{i} \in \Lambda}} \!\!\! \varphi^{i}_{B_{i-1}c_{i-1}c_{i}} \! (B^{-1}_{i} \delta_{i}) e^{i-1}_{c_{i-1}} \!\! \left( \! A_{\theta_k} \! (\hat{x} \! - \! \delta_{i}), \! A_{\theta_{k}}B_{i-1}\! \right) \!\! - \!\!\!\!\! \sum_{\substack{c_{i-1} \\ B_{i-1} \in S \\ \delta_{i} \in \Lambda}} \!\!\! \varphi^{i}_{B_{i-1}c_{i-1}c_{i}} \! (B^{-1}_{i}\! A_{\theta_{k}^{-1}} \delta_{i}) e^{i-1}_{c_{i-1}} \!\left(\!A_{\theta_k} \hat{x} \! - \! \delta_{i}, \!B_{i-1}\!\right) \! \right|\\
    = & \sum_{c_{i-1}} \! \left| \! \sum_{\substack{B_{i-1} \in S \\ \delta_{i} \in \Lambda}} \!\!\! \varphi^{i}_{B_{i-1}c_{i-1}c_{i}} \! (B^{-1}_{i}\! \delta_{i}) e^{i-1}_{c_{i-1}} \!\! \left( \! A_{\theta_k} \!(\hat{x} \! - \! \delta_{i}), \! A_{\theta_{k}}\! B_{i-1}\!\right) \!\! -\!\!\!\!\! \sum_{\substack{B_{i-1} \in S \\ \delta_{i} \in \Lambda}}\!\!\!\! \varphi^{i}_{B_{i-1}c_{i-1}c_{i}} \! (B^{-1}_{i} \! A_{\theta_{k}}^{-1} \delta_{i}) e^{i-1}_{c_{i-1}} \!\!\left(\!\! A_{\theta_k} \hat{x} \!-\! \delta_{i}, \! B_{i-1}\!\right) \! \right| \\
    \leq & n_{i-1} \frac{C_{i}}{2} (p+1)^2 h^2,
    \end{split}
\end{equation}
where we do not specifically indicate the numbers of input and output channels, i.e. $\varphi^i(x)=\varphi^i_{Ac_{i-1} c_i}(x)$, and we have \\ $C_i = \operatorname{sup} \left( \lb\| \nabla^2 \varphi^i (x) \rb\| \lb| e^{i-1}_{c_{i-1}} (x, B) \rb| + \lb| \varphi^i (x)\rb| \lb\| \nabla^2 e^{i-1}_{c_{i-1}} (x, B) \rb\| + 2 \lb\| \nabla \varphi^i (x)\rb\| \lb\| \nabla e^{i-1}_{c_{i-1}} (x, B) \rb\| \right)$. Therefore, according to Eqs. (\ref{l766}) and (\ref{l782}), we have
\begin{equation}\label{i_layer}
    \begin{split}
    & \lb| \lb( \hat{\Upsilon}\left[ \hat{\Phi}_{N-1} \cdots \hat{\Phi}_{i} \left[ \fdk^{-1} \left[ \hat{\Phi}_{i-1} \cdots \hat{\Phi}_{2} \left[\hat{\Psi} \right]\cdots\right]\right]\right] (\hat{I}) - \hat{\Upsilon}\left[ \hat{\Phi}_{N-1} \cdots \hat{\Phi}_{i+1} \left[ \fdk^{-1} \left[ \hat{\Phi}_{i} \cdots \hat{\Phi}_{2} \left[\hat{\Psi} \right]\cdots\right]\right]\right] (\hat{I}) \rb)_{ij}^{c_N} \rb| \\
    \leq & n_{N-1} p^2 F_N \cdots n_{i} p^2 F_{i+1} n_{i-1} \frac{C_{i}}{2} (p+1)^2 h^2 \\
    = & \left(\prod_{k=i+1}^{N} n_{k-1} p^2 F_k \right) n_{i-1} \frac{C_{i}}{2} (p+1)^2 h^2.
    \end{split}
\end{equation}

Substituting Eqs. (\ref{l51}), (\ref{l52}) and (\ref{l53}) into Eq. (\ref{i_layer}), denoting $\mathcal{F}_{i-1} \!=\! \prod \nolimits_{k=1}^{i-1} n_{k-1} p^2 F_k$, then we have

\begin{equation}\label{i_layer_result}
    \begin{split}
    & \lb| \lb( \hat{\Upsilon}\left[ \hat{\Phi}_{N-1} \cdots \hat{\Phi}_{i} \left[ \fdk^{-1} \left[ \hat{\Phi}_{i-1} \cdots \hat{\Phi}_{2} \left[\hat{\Psi} \right]\cdots\right]\right]\right] (\hat{I}) - \hat{\Upsilon}\left[ \hat{\Phi}_{N-1} \cdots \hat{\Phi}_{i+1} \left[ \fdk^{-1}\left[ \hat{\Phi}_{i} \cdots \hat{\Phi}_{2} \left[\hat{\Psi} \right]\cdots\right]\right]\right] (\hat{I}) \rb)_{ij}^{c_N} \rb| \\
    \leq & \left(\prod_{k=i+1}^{N} n_{k-1} p^2 F_k \right) \frac{n_{i-1}(p+1)^2 h^2}{2} \left( \lb\| \nabla^2 \varphi^i (x) \rb\| \lb| e^{i-1}_{c_{i-1}} (x, B) \rb| + \lb| \varphi^i (x)\rb| \lb\| \nabla^2 e^{i-1}_{c_{i-1}} (x, B) \rb\| + 2 \lb\| \nabla \varphi^i (x)\rb\| \lb\| \nabla e^{i-1}_{c_{i-1}} (x, B) \rb\| \right)\\
    \leq & 2 \left(\prod_{k=i}^{N} n_{k-1} p^2 F_k \right) \left( \frac{H_i}{F_i} \lb| e^{i-1}_{c_{i-1}} (x, B) \rb| + \lb\| \nabla^2 e^{i-1}_{c_{i-1}} (x, B) \rb\| + 2 \frac{G_i}{F_i} \lb\| \nabla e^{i-1}_{c_{i-1}} (x, B) \rb\| \right) h^2\\
    \leq & 2 \left(\prod_{k=i}^{N} n_{k-1} p^2 F_k \right) \mathcal{F}_{i-1}\left( \! \frac{H_i F_0}{F_i} \! +  \! \sum_{m=1}^{i-1} \frac{H_m F_0}{F_m} \! + \! 2 \sum_{l=1}^{i-1} \frac{G_l}{F_l} \sum_{m=1}^{l-1} \frac{G_m F_0}{F_m} \! + \! 2 \sum_{m=1}^{i-1} \frac{G_m G_0}{F_m} \! + \! H_0 \! + \! \sum_{m=1}^{i-1} 2 \frac{G_i G_m F_0}{F_i F_m} \! + \! 2\frac{G_i G_0}{F_i} \! \right) \! h^2 \\
    \leq & 2 \mathcal{F} \left(\sum_{m=1}^{i} \frac{H_m F_0}{F_m} \! + \! 2 \sum_{l=1}^{i} \frac{G_l}{F_l} \sum_{m=1}^{l-1} \frac{G_m F_0}{F_m} \! + \! 2 \sum_{m=1}^{i} \frac{G_m G_0}{F_m} + H_0 \right) h^2.
    \end{split}
\end{equation}

3) For the output layer, with Eqs. (\ref{input_d2}), (\ref{inter_d2}) and (\ref{output_d2}) we have

\begin{equation}\nonumber
    \begin{split}
    & \lb| \lb( \hat{\Upsilon}\left[ \fdk^{-1}\left[ \hat{\Phi}_{N-1} \cdots \hat{\Phi}_{i+1} \left[ \hat{\Phi}_{i} \cdots \hat{\Phi}_{2} \left[\hat{\Psi} \right]\cdots\right]\right]\right] (\hat{I}) - \fdk^{-1}\left[\hat{\Upsilon}\left[ \hat{\Phi}_{N-1} \cdots \hat{\Phi}_{i+1} \left[ \hat{\Phi}_{i} \cdots \hat{\Phi}_{2} \left[\hat{\Psi} \right] \cdots \right]\right]\right] (\hat{I}) \rb)_{ij}^{c_N} \rb| \\
    = & \! \lb| \!\sum_{\substack{c_{N-1} \\ B_{N-1} \in S \\ \delta_{N} \in \Lambda}} \!\!\!\!\! \varphi^{N}_{c_{N \!-\!1} c_N}(B^{-1}_{N \!-\!1}\! \delta_{N}\!) e^{N \!- \!1}_{c_{N \!-\!1}} \!\! \lb(\! A_{\theta_k} \!\! \lb(x \!\!-\!\! \delta_{N}+b\! \rb)\!,\! A_{\theta_k} \! B_{N-1}\!\rb) \! - \!\!\!\!\!\!\! \sum_{\substack{c_{N-1} \\ B_{N-1} \in S \\ \delta_{N} \in \Lambda}} \!\!\!\! \varphi^{N}_{c_{N-1} c_N}\! (B^{-1}_{N-1} A_{\theta_{k}}^{-1} \delta_{N}\!) e^{N-1}_{c_{N-1}} \!\! \lb(\! A_{\theta_k} (x+b) \! - \! \delta_{N},\! B_{N-1}\rb) \! \rb| \!. \\
    \end{split}
    \label{final_layer_inter}
\end{equation}
Then, by Eq. (\ref{Thm2}) from Remark 2 for the output, we have:

\begin{equation}\label{n_layer}
    \begin{split}
    & \lb| \lb( \hat{\Upsilon}\left[ \fdk^{-1}\left[ \hat{\Phi}_{N-1} \cdots \hat{\Phi}_{i+1} \left[ \hat{\Phi}_{i} \cdots \hat{\Phi}_{2} \left[\hat{\Psi} \right]\cdots\right]\right]\right] (\hat{I}) - \fdk^{-1}\left[\hat{\Upsilon}\left[ \hat{\Phi}_{N-1} \cdots \hat{\Phi}_{i+1} \left[ \hat{\Phi}_{i} \cdots \hat{\Phi}_{2} \left[\hat{\Psi} \right] \cdots \right]\right]\right] (\hat{I}) \rb)_{ij}^{c_N} \rb| \\
    \leq & \! \sum_{c_{N-1}} \!\! \lb|\! \sum_{\substack{B_{N \!-\!1} \in S \\ \delta_{N} \in \Lambda}} \!\!\! \varphi^{N}_{c_{N \!-\!1} c_N}\! (\!B^{-1}_{N \!-\!1} \delta_{N}\!) e^{N \!-\!1}_{c_{N \!-\!1}} \!\! \lb( \! D \! A_{\theta_k} \! \lb(x \!\! - \!\! \delta_{N}+b\rb)\!, \! A_{\theta_{k}} B_{N-1} \!\rb) \! - \!\!\!\!\! \sum_{\substack{B_{N \!-\!1} \in S \\ \delta_{N} \in \Lambda}} \!\! \varphi^{N}_{c_{N \!-\!1} c_{N}}(\! B^{-1}_{N-1} A_{\theta_{k}}^{-1} \! \delta_{N}\!) e^{N-1}_{c_{N-1}}\lb(\! A_{\theta_k} (x+b) \! - \!\delta_{N}\!,\! B_{N-1}\!\rb)\! \rb| \\
    \leq & n_{N-1} \frac{C_N}{2} (p+1)^2 h^2,
    \end{split}
\end{equation}
where we do not specifically indicate the numbers of input and output channels, i.e. $\varphi^N(x)=\varphi^N_{c_{N-1} c_N}(x)$, and we have $C_N = \operatorname{sup}\left( \lb\| \nabla^2 \varphi^{N} \! (x) \rb\| \! \lb| e^{N-1}_{c_{N-1}} \! (x, B) \rb| \! + \! \lb| \varphi^{N} \! (x)\rb| \! \lb\| \nabla^2 e^{N-1}_{c_{N-1}} \! (x, B) \rb\| \! + \! 2 \lb\| \nabla \varphi^{N} \! (x)\rb\| \! \lb\| \nabla e^{N-1}_{c_{N-1}} \! (x, B) \rb\| \right) $.
\vspace{2mm}

Substituting Eqs. (\ref{l51}), (\ref{l52}) and (\ref{l53}) into Eq. (\ref{n_layer}), denoting $\mathcal{F}_{N-1} \!=\! \prod \nolimits_{k=1}^{N-1} n_{k-1} p^2 F_k$, then we have

\begin{equation}\label{n_layer_result}
    \begin{split}
    & \lb| \lb( \hat{\Upsilon}\left[ \fdk^{-1}\left[ \hat{\Phi}_{N-1} \cdots \hat{\Phi}_{i+1} \left[ \hat{\Phi}_{i} \cdots \hat{\Phi}_{2} \left[\hat{\Psi} \right]\cdots\right]\right]\right] (\hat{I}) - \fdk^{-1}\left[\hat{\Upsilon}\left[ \hat{\Phi}_{N-1} \cdots \hat{\Phi}_{i+1} \left[ \hat{\Phi}_{i} \cdots \hat{\Phi}_{2} \left[\hat{\Psi} \right] \cdots \right]\right]\right] (\hat{I}) \rb)_{ij}^{c_N} \rb| \\
    \leq & \frac{n_{N-1} (p+1)^2 h^2}{2} \left( \lb\| \nabla^2 \varphi^{N} \! (x) \rb\| \! \lb| e^{N-1}_{c_{N-1}} \! (x, B) \rb| \! + \! \lb| \varphi^{N} \! (x)\rb| \! \lb\| \nabla^2 e^{N-1}_{c_{N-1}} \! (x, B) \rb\| \! + \! 2 \lb\| \nabla \varphi^{N} \! (x)\rb\| \! \lb\| \nabla e^{N-1}_{c_{N-1}} \! (x, B) \rb\| \right)\\
    \leq & 2 n_{N-1} p^2 F_N \left(\frac{H_N}{F_N} \lb| e^{N-1}_{c_{N-1}} \! (x, B) \rb| + \lb\| \nabla^2 e^{N-1}_{c_{N-1}} \! (x, B) \rb\| + 2 \frac{G_N}{F_N} \lb\| \nabla e^{N-1}_{c_{N-1}} \! (x, B) \rb\| \right) h^2 \\
    \leq & 2 n_{N-1} p^2 F_N \mathcal{F}_{N-1} \left( \frac{H_N}{F_N} F_0 \! + \! \sum_{m=1}^{N-1} \frac{H_m F_0}{F_m } \! + \! 2 \sum_{l=1}^{N-1} \frac{G_l}{F_l} \sum_{m=1}^{l-1} \frac{G_m F_0}{F_m} \! + \! 2 \sum_{m=1}^{N-1} \frac{G_m G_0}{F_m} + H_0 \! + \! \sum_{m=1}^{N-1} 2 \frac{G_N G_m F_0}{F_N F_m} \! + \! 2\frac{G_N G_0}{F_N} \right) h^2 \\
    \leq & 2 \mathcal{F} \left( \sum_{m=1}^{N} \frac{H_m F_0}{F_m } + 2 \sum_{l=1}^{N} \frac{G_l}{F_l} \sum_{m=1}^{l-1} \frac{G_m F_0}{F_m} + 2 \sum_{m=1}^{N} \frac{G_m G_0}{F_m} + H_0 \right) h^2.
    \end{split}
\end{equation}

4) 
Substituting Eqs. (\ref{1_layer}),   (\ref{i_layer_result}) and (\ref{n_layer_result}) into Eq. (\ref{hole_error_seperate}) we can get:

\begin{equation}\label{hole_error_seperate_02}
    \begin{split}
    & \left| \operatorname g\left[\tilde{f}_{\theta_{k}b}^{-1} \right]\! (\hat{I}) \! - \! \tilde{f}_{\theta_{k}b}^{-1} \left[\operatorname g \right](\hat{I}) \right| \\
    \leq & 2 \mathcal{F} \left(\frac{H_1}{F_1} F_0 + H_0 + 2\frac{G_1}{F_1} G_0\right) h^2 \! + \! \sum_{i=2}^{N-1} \! 2 \mathcal{F} \left(\sum_{m=1}^{i} \frac{H_m F_0}{F_m } + 2 \sum_{l=1}^{i} \frac{G_l}{F_l} \sum_{m=1}^{l-1} \frac{G_m F_0}{F_m} + 2 \sum_{m=1}^{i} \frac{G_m G_0}{F_m} + H_0 \right) h^2 \! \\
    & + 2 \mathcal{F} \left( \sum_{m=1}^{N} \frac{H_m F_0}{F_m } + 2 \sum_{l=1}^{N} \frac{G_l}{F_l} \sum_{m=1}^{l-1} \frac{G_m F_0}{F_m} + 2 \sum_{m=1}^{N} \frac{G_m G_0}{F_m} + H_0 \right) h^2 \\ 
    \leq & 2 \mathcal{F} \sum_{i=1}^{N} \left(\sum_{m=1}^{i} \frac{H_m F_0}{F_m} + 2 \sum_{l=1}^{i} \frac{G_l}{F_l} \sum_{m=1}^{l-1} \frac{G_m F_0}{F_m} + 2 \sum_{m=1}^{i} \frac{G_m G_0}{F_m} + H_0 \right) h^2. \\
    \end{split}
\end{equation}

Therefore, we achieve Eq. (\ref{error_seperate}). 

Finally, we will provide the error analysis for the N-layer rotation equivariant CNN network. Substituting Eqs. (\ref{error_continue_to_seperate_01}), (\ref{error_continue_to_seperate_02}) and (\ref{error_seperate}) into Eq. (\ref{error_factoring}), we can get:
\begin{equation}\label{error_exact}
    \begin{split}
    & \lb|\fdd^{-1}\operatorname g\left[\fdd \right]\! ({I}) \!-\! \left[\operatorname g \right]\!({I})\rb| \\
    \leq & \frac{2 \pi}{t} \mathcal{F} \left(\operatorname{max}\left\{H,W\right\} + N \lb(p+1\rb) \right) h G_0 + \frac{2 \pi}{t} \mathcal{F} \operatorname{max}\left\{H,W\right\} h G_0 \\
    & ~~~~~~~~~~~~~~~~~~~~~~~~~~~~~~~~~~~~~~~~~~~~~~~~~~~~~~~~~~~~~~~~~~~~~ + 2 \mathcal{F} \sum_{i=1}^{N} \left(\sum_{m=1}^{i} \frac{H_m F_0}{F_m} + 2 \sum_{l=1}^{i} \frac{G_l}{F_l} \sum_{m=1}^{l-1} \frac{G_m F_0}{F_m} + 2 \sum_{m=1}^{i} \frac{G_m G_0}{F_m} + H_0 \right) h^2  \\
    \leq & \frac{2 \pi}{t} \mathcal{F} \!\left(2\operatorname{max}\left\{H,W\right\} + N \lb(p+1\rb) \right) h G_0 + 2 \mathcal{F} \sum_{i=1}^{N} \left(\sum_{m=1}^{i} \frac{H_m F_0}{F_m} + 2 \sum_{l=1}^{i} \frac{G_l}{F_l} \sum_{m=1}^{l-1} \frac{G_m F_0}{F_m} + 2 \sum_{m=1}^{i} \frac{G_m G_0}{F_m} + H_0 \right) h^2 \\
    \end{split}
\end{equation}

Next, in order to get a more concise form, we further scale the entire error bound. By Eq. (\ref{error_exact}), we have:

\begin{equation}
    \begin{split}
    & \lb|\fdd^{-1}\operatorname g\left[\fdd \right] ({I}) \!-\! \left[\operatorname g \right]({I})\rb| \\
     = & \frac{2 \pi}{t} \mathcal{F} \left(2\operatorname{max}\left\{H,W\right\} p^{-1} + N \lb(p+1\rb) p^{-1} \right) p h G_0 \! + \! 2 \mathcal{F} \sum_{i=1}^{N} \! (N \!+ \!1 \!- \! i) \! \left( \! \frac{H_i F_0}{F_i} \! + \! 2 \frac{G_i}{F_i} \! \sum_{m=1}^{i-1} \! \frac{G_m F_0}{F_m} \! + \! 2 \frac{G_i G_0}{F_i} \! \right) h^2 \! + \! 2 \mathcal{F} N H_0 h^2 \\
     \leq & \frac{2 \pi}{t} \mathcal{F} \left(2\operatorname{max}\left\{H,W\right\} p^{-1} + 2N \right) p h G_0 + 2 \mathcal{F} N \sum_{i=1}^{N} \! \! \left( \! \frac{H_i F_0}{F_i} \! + \! 2 \frac{G_i}{F_i} \! \sum_{m=1}^{i-1} \! \frac{G_m F_0}{F_m} \! + \! 2 \frac{G_i G_0}{F_i} \! \right) h^2 \! + \! 2 \mathcal{F} N H_0 h^2 \\
     \leq & 2 N \mathcal{F} \sum_{i=1}^{N} \! \! \left( \! \frac{H_i F_0}{F_i} \! + \! 2 \frac{G_i}{F_i} \! \sum_{m=1}^{i-1} \! \frac{G_m F_0}{F_m} \! + \! 2 \frac{G_i G_0}{F_i} + H_0 \! \right) h^2 + 2 \pi G_0 \mathcal{F} \left(2\operatorname{max}\left\{H,W\right\} p^{-1} + 2N \right) p h t^{-1}
    \end{split}
\end{equation}

If we denote
\begin{equation}
    \begin{split}
    & C_1 = 2 N \mathcal{F} \cdot \sum_{i=1}^{N} \! \! \left( \! \frac{H_i F_0}{F_i} \! + \! 2 \frac{G_i}{F_i} \! \sum_{m=1}^{i-1} \! \frac{G_m F_0}{F_m} \! + \! 2 \frac{G_i G_0}{F_i} + H_0 \! \right), \\
    & C_2 = 2 \pi G_0 \mathcal{F} \lb(2 \operatorname{max}\{H,\!W\} p^{-1} + 2 N\rb).
    \end{split}
\end{equation}
Then we have
\begin{equation}
    \begin{split}
    & \lb|\fdd^{-1}\operatorname g\left[\fdd \right] ({I}) \!-\! \left[\operatorname g \right]({I})\rb|
    \leq C_1 h^2 + C_2 p h t^{-1}.
    \end{split}
\end{equation}

From above all, we obtained the error bound of the N-layer rotation equivariant CNN network as Eq. (\ref{main_conclusion}).
By defining $I_{\theta b} = [\fdd](I)$, Eq. (\ref{main_conclusion}) can be reformulated as: 
\begin{equation}\label{eq:lemma6}
\begin{split}
  \lb|\fdd^{-1}\operatorname [\operatorname g]\left( I_{\theta b}\right)\! \!-\! \left[\operatorname g\right]\!({I})\rb|
\!\leq\! C_1 h^2\!+\! C_2 { p h}{t^{-1}},
\end{split}
\end{equation}

\end{proof}

Based on the preceding lemmas, we now present and prove our main theorem.

\begin{Thm}\label{thm:thm1}
For an image ${I}$ with size $H\times W\times n_0$, a $N$-layer rotation-translation equivariant CNN network $\operatorname g(\cdot)$, whose channel number of the $i^{th}$ layer is $n_i$, rotation equivariant subgroup is $S\leqslant O(2)$, $|S|=t$, and activation function is set as ReLU. If the latent continuous function of the $c^{th}$ channel of ${I}$ denoted as $r_c: \mathbb{R}^2 \! \rightarrow \! \mathbb{R}$, and the latent continuous function of any convolution filters in the $i^{th}$ layer denoted as $\varphi^{i}: \mathbb{R}^2 \! \rightarrow \! \mathbb{R}$, where $i \in \{1, \cdots, N\}$, $c \in \{1, \cdots, n_{0}\}$, for any $x\in \R^2$, the following conditions are satisfied:
    \begin{equation}
        \begin{split}
            & |r_c(x)| \leq F_0, \|\nabla r_c(x)\| \leq G_0, \|\nabla ^2 r_c(x)\| \leq H_0, \\
            & |\varphi^{i}(x)| \leq F_i, \|\nabla \varphi^{i}(x)\| \leq G_i, \|\nabla ^2 \varphi^{i}(x)\| \leq H_i,\\
            &\forall \|x\|\geq\nicefrac{(p+1)h}{2},~ \varphi_{i}(x)=0,\\
        \end{split}
    \end{equation}
    where $p$ is the filter size, $h$ is the mesh size, $\nabla$ and ${\nabla}^2$ denote the operators of gradient and Hessian matrix, respectively. 
For an arbitrary $0 \leq \theta \leq 2 \pi$ and a feature map of equivariant convolution ${Z}$ with size $H\times W\times tC$, $A_{\theta}$ denotes the rotation matrix, $b\in \mathbb{R}^2$ denotes the translation, and the following result is satisfied:
\begin{equation}\label{eq:thm1}
\begin{split}
\lb\|\tilde{Z}_{\theta b}-Z\rb\|_F \leq C_3 \lb\|\fdd^{-1}(I_{\theta b})-I\rb\|_F + C_1 h^2\!+\! C_2 { p h}{t^{-1}},
\end{split}
\end{equation}
where
 \begin{equation}
    \begin{split}
    &Z = \operatorname{g}(I), {Z}_{\theta b} = \operatorname{g}(I_{\theta b}), \tilde{Z}_{\theta b} = \fdd^{-1}(Z_{\theta b}), Z, {Z}_{\theta b}, \tilde{Z}_{\theta b}\in \mathbb{R}^{H\times W\times tC}, \\
        & C_1 = 2 N \mathcal{F} \cdot \sum_{i=1}^{N} \! \! \left( \! \frac{H_i F_0}{F_i} \! + \! 2 \frac{G_i}{F_i} \! \sum_{m=1}^{i-1} \! \frac{G_m F_0}{F_m} \! + \! 2 \frac{G_i G_0}{F_i} + H_0 \! \right), \\
        & C_2 = 2 \pi G_0 \mathcal{F} \lb(2 \operatorname{max}\{H,\!W\} p^{-1} + 2 N\rb), \mathcal{F} \!=\prod \nolimits_{i=1}^N n_{i-1} p^2 F_i.\\
        & C_3 = .\\
    \end{split}
\end{equation}
\end{Thm}

\section{Complementary Results}
\begin{figure}
    \centering
    \includegraphics[width=0.95\linewidth]{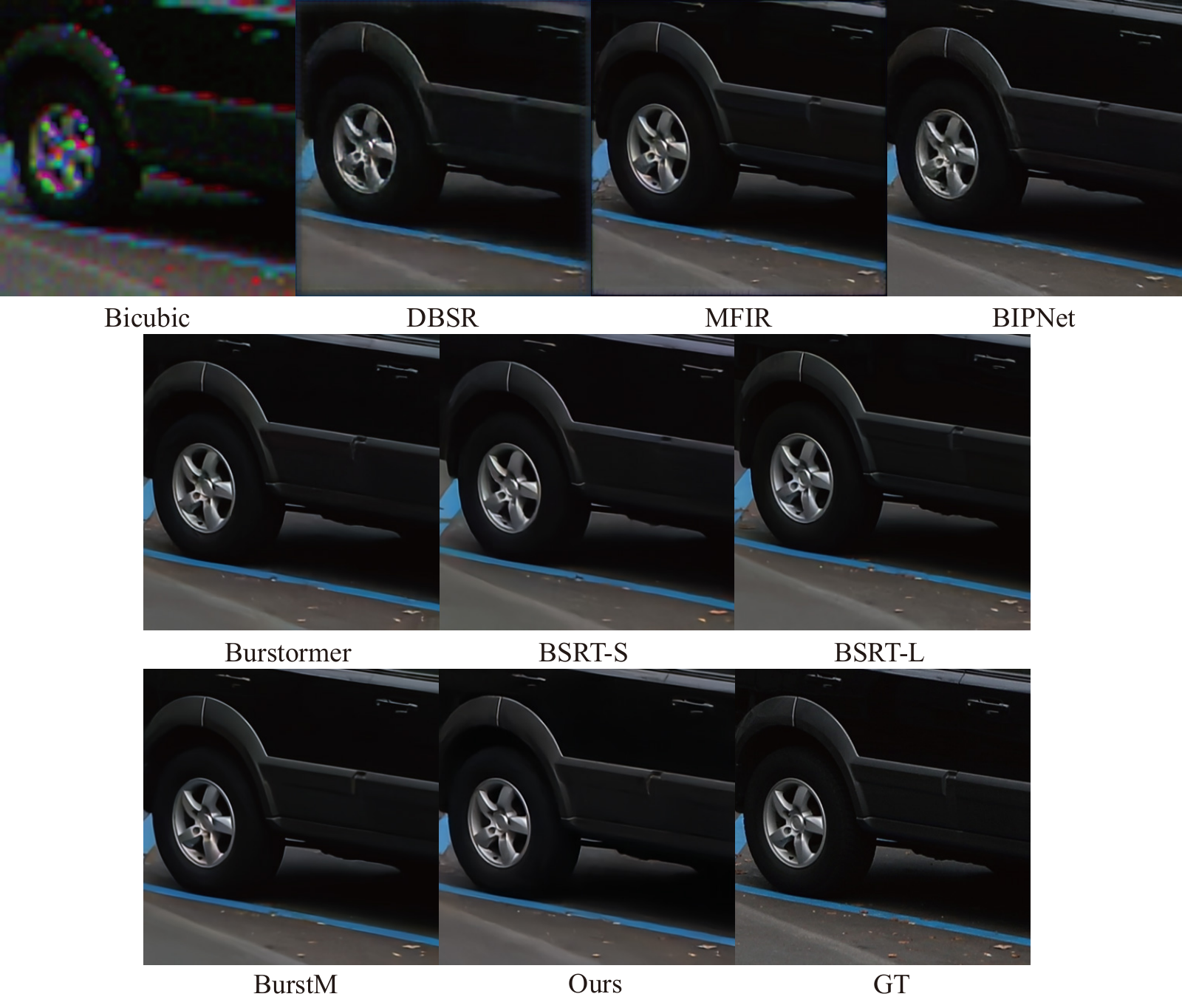}
    \caption{Visual comparison of x4 BISR on the SyntheticBurst
dataset.}
    \label{fig:sup_z_x4_1}
\end{figure}

\begin{figure}
    \centering
    \includegraphics[width=0.95\linewidth]{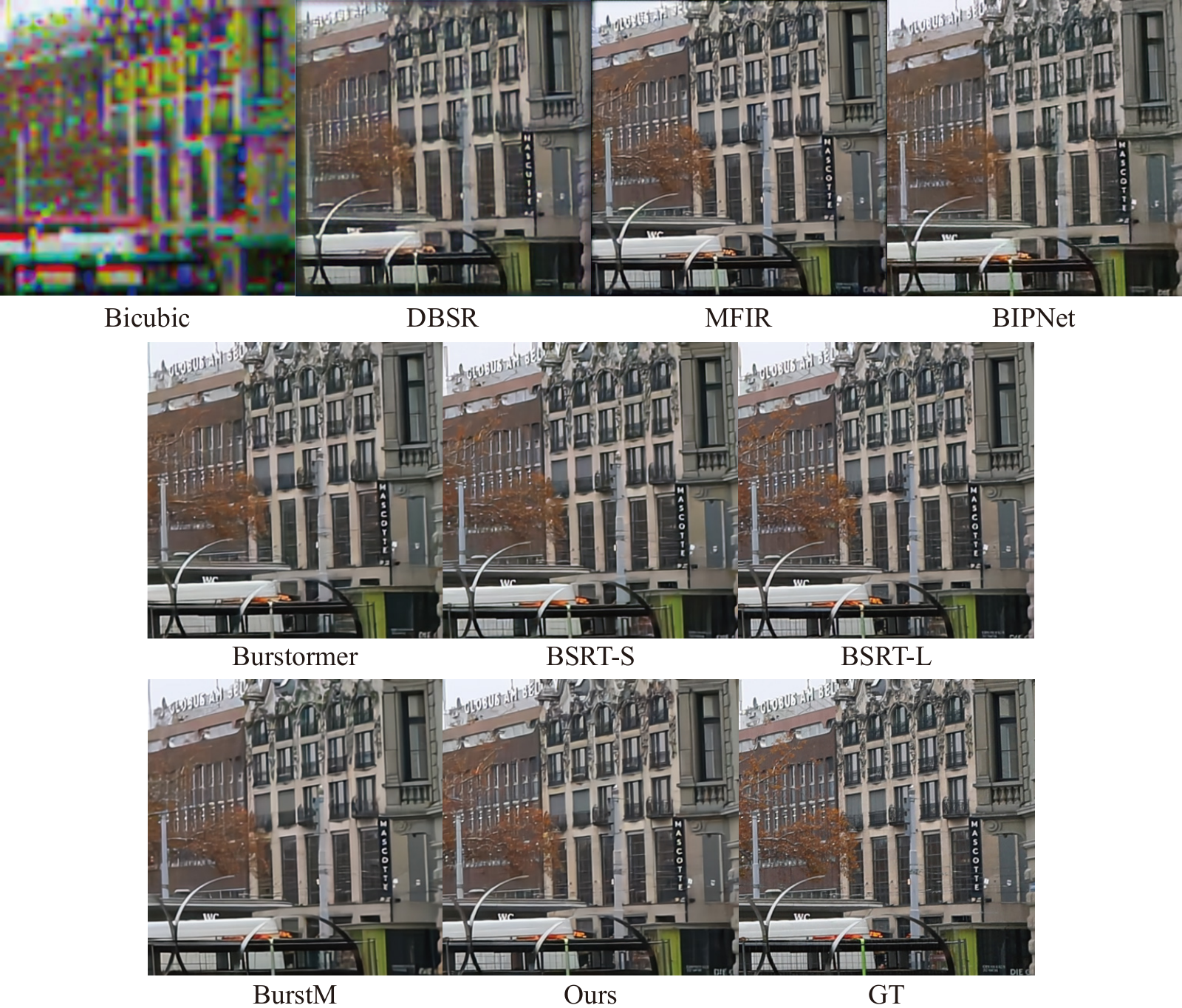}
    \caption{Visual comparison of x4 BISR on the SyntheticBurst
dataset.}
    \label{fig:sup_z_x4_2}
\end{figure}

\begin{figure}
    \centering
\includegraphics[width=0.95\linewidth]{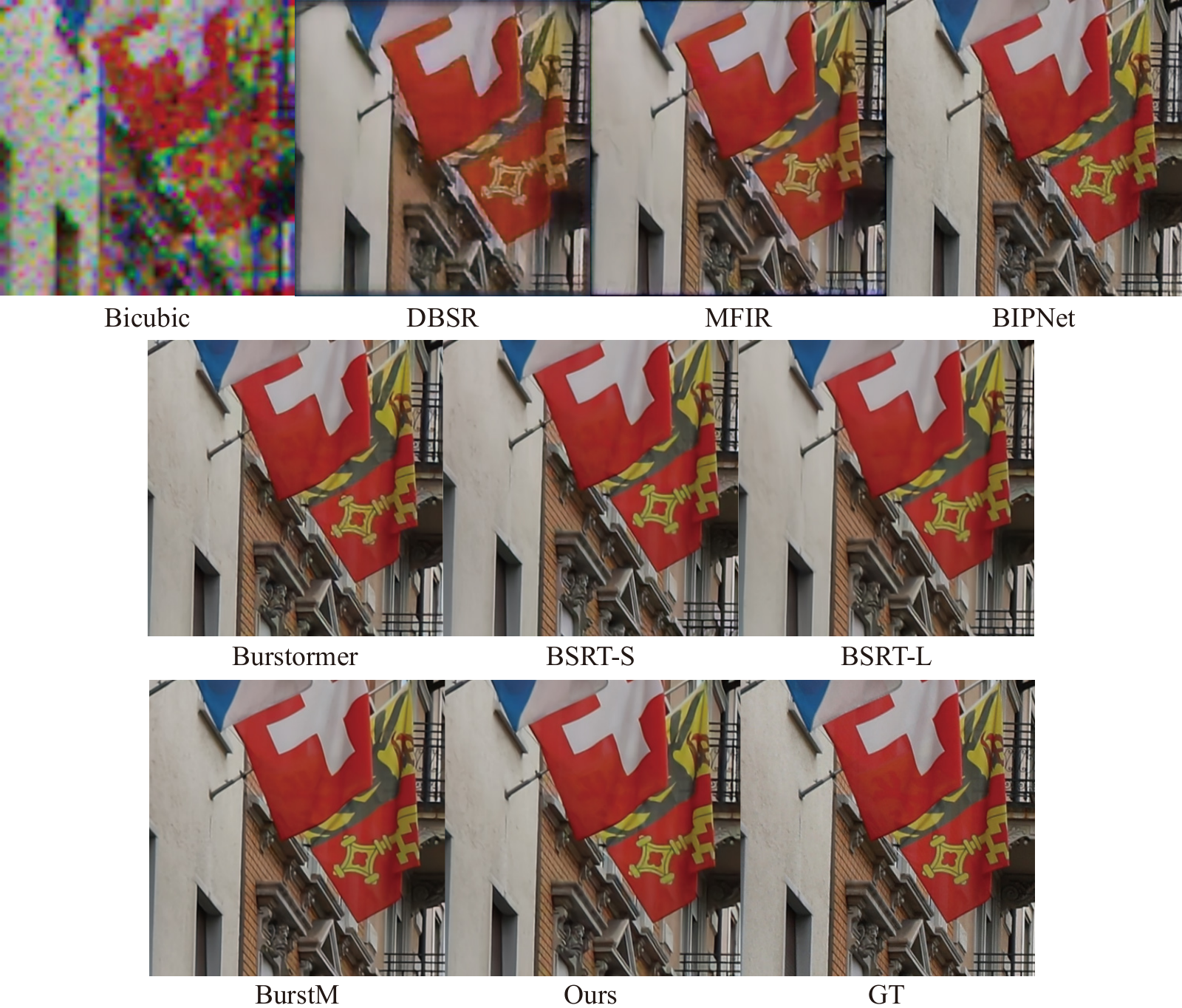           
               }
    \caption{Visual comparison of x4 BISR on the SyntheticBurst
dataset.}
    \label{fig:sup_z_x4_3}
\end{figure}

\begin{figure}
    \centering
    \includegraphics[width=0.95\linewidth]{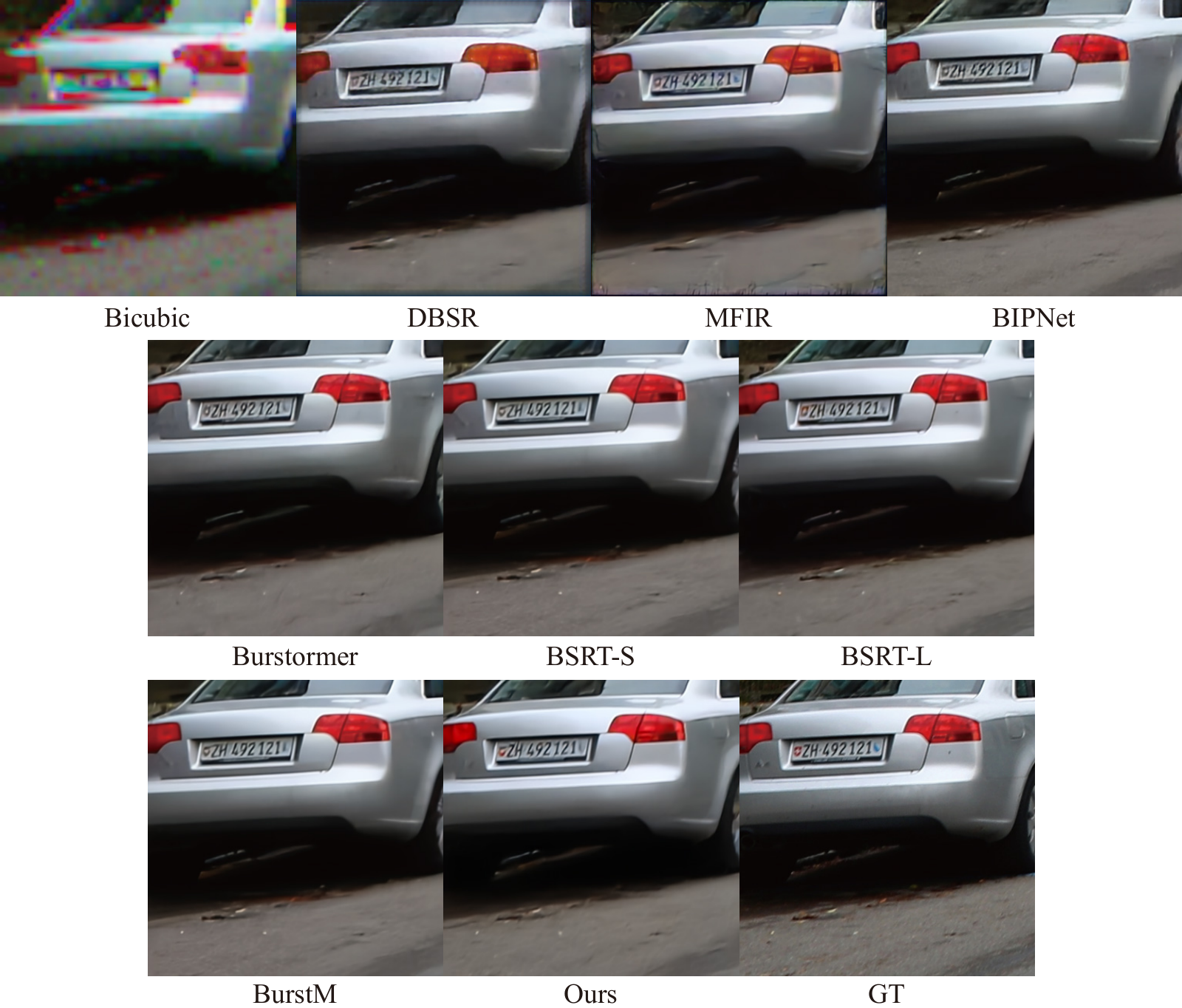}
    \caption{Visual comparison of x4 BISR on the SyntheticBurst
dataset.}
    \label{fig:sup_z_x4_4}
\end{figure}

\begin{figure}
    \centering
\includegraphics[width=0.95\linewidth]{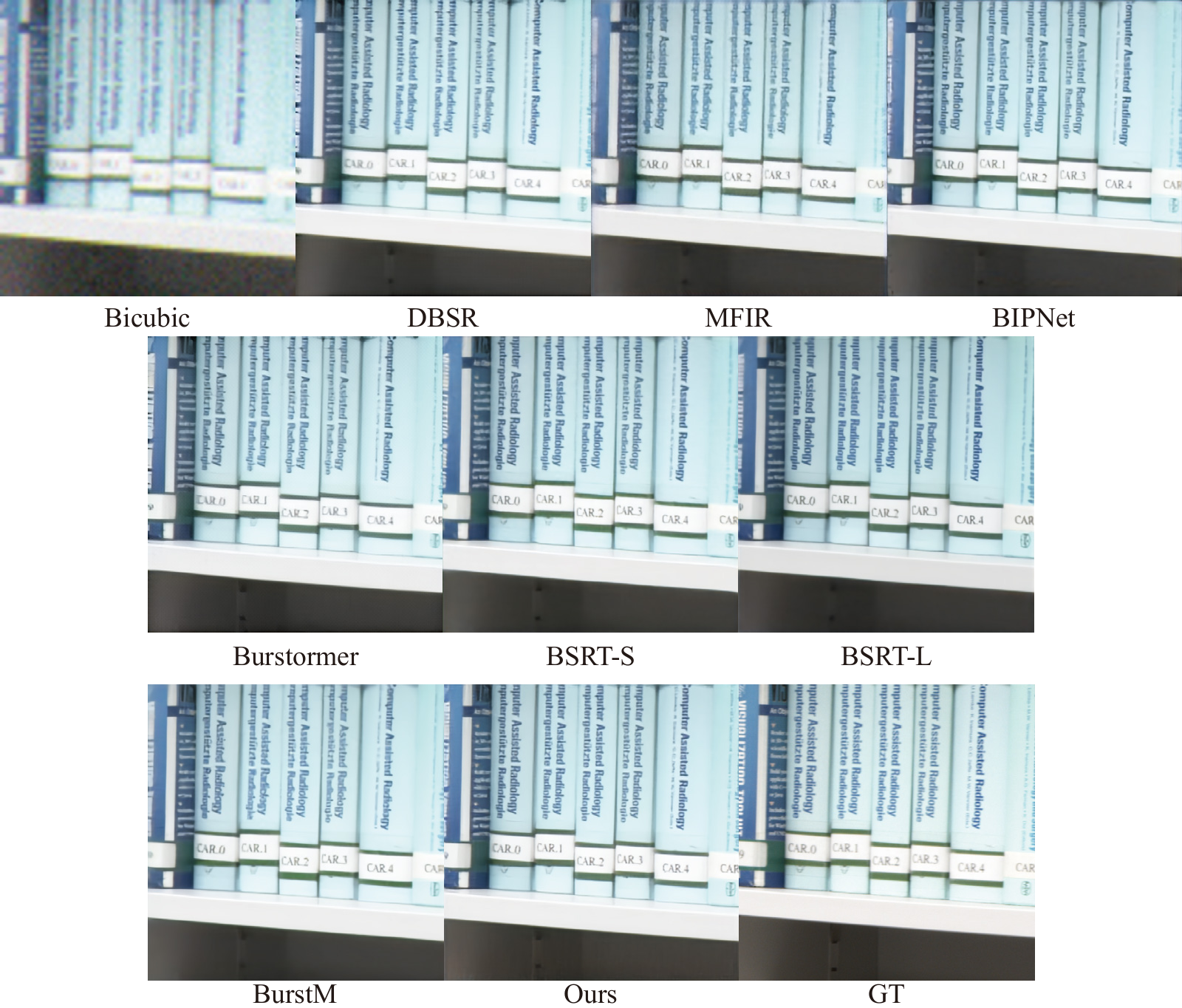}         
    \caption{Visual comparison of x4 BISR on the BurstSR
dataset.}
    \label{fig:sup_r_x4_1}
\end{figure}

\begin{figure}
    \centering
\includegraphics[width=0.95\linewidth]{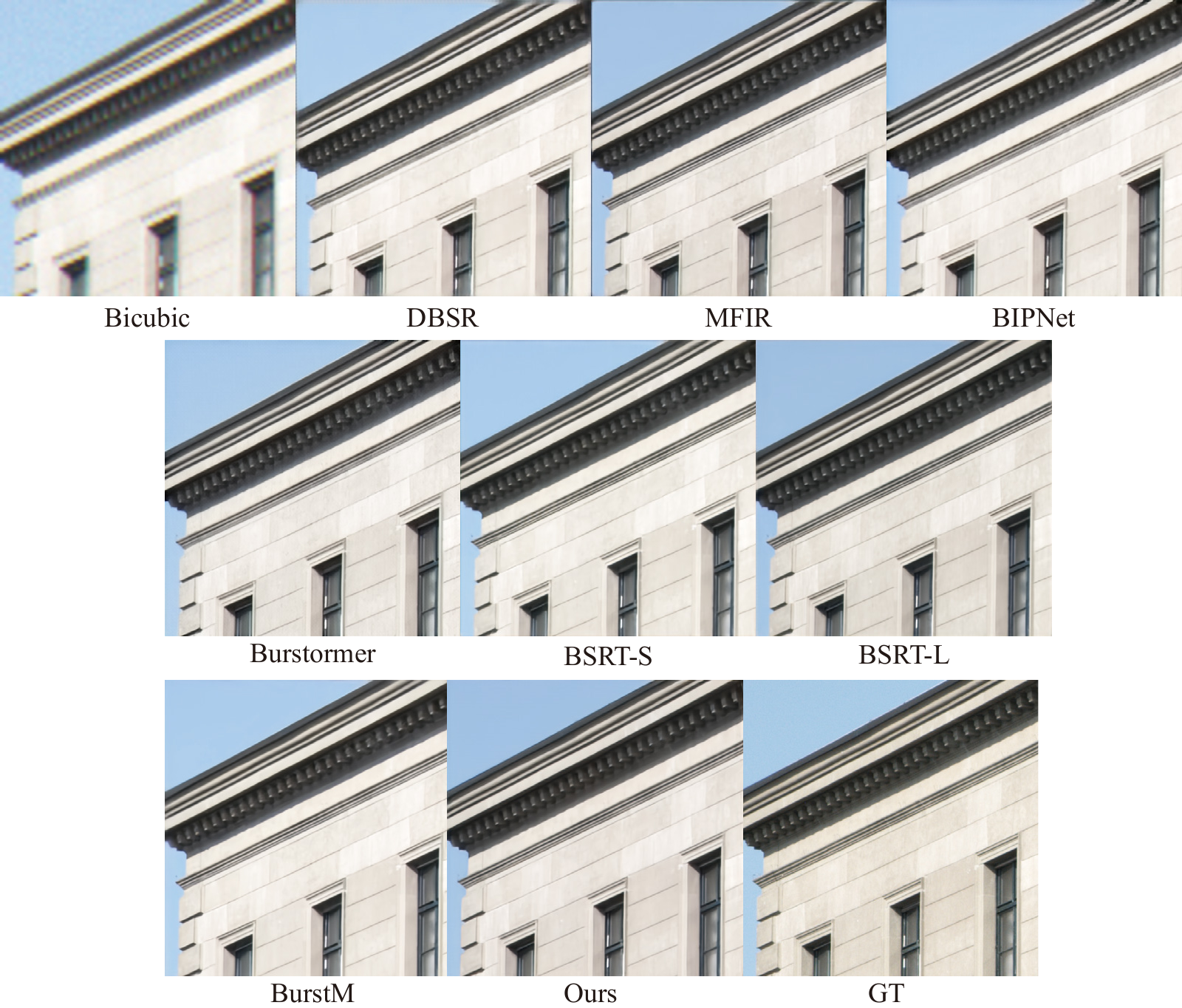}          
    \caption{Visual comparison of x4 BISR on the BurstSR
dataset.}
    \label{fig:sup_r_x4_2}
\end{figure}

\begin{figure}
    \centering
\includegraphics[width=0.95\linewidth]{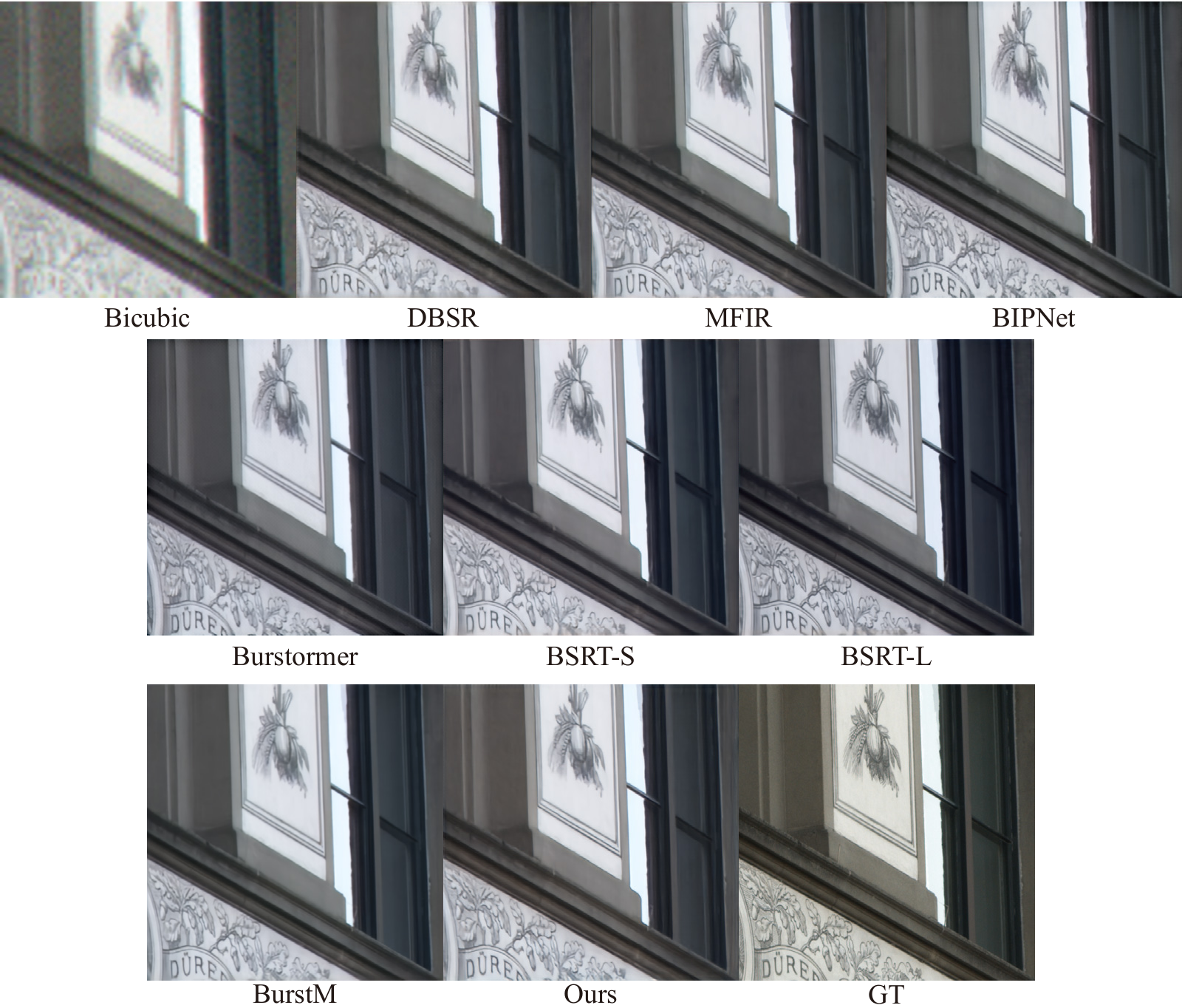}         
    \caption{Visual comparison of x4 BISR on the BurstSR
dataset.}
    \label{fig:sup_r_x4_3}
\end{figure}

\begin{figure}
    \centering
\includegraphics[width=0.95\linewidth]{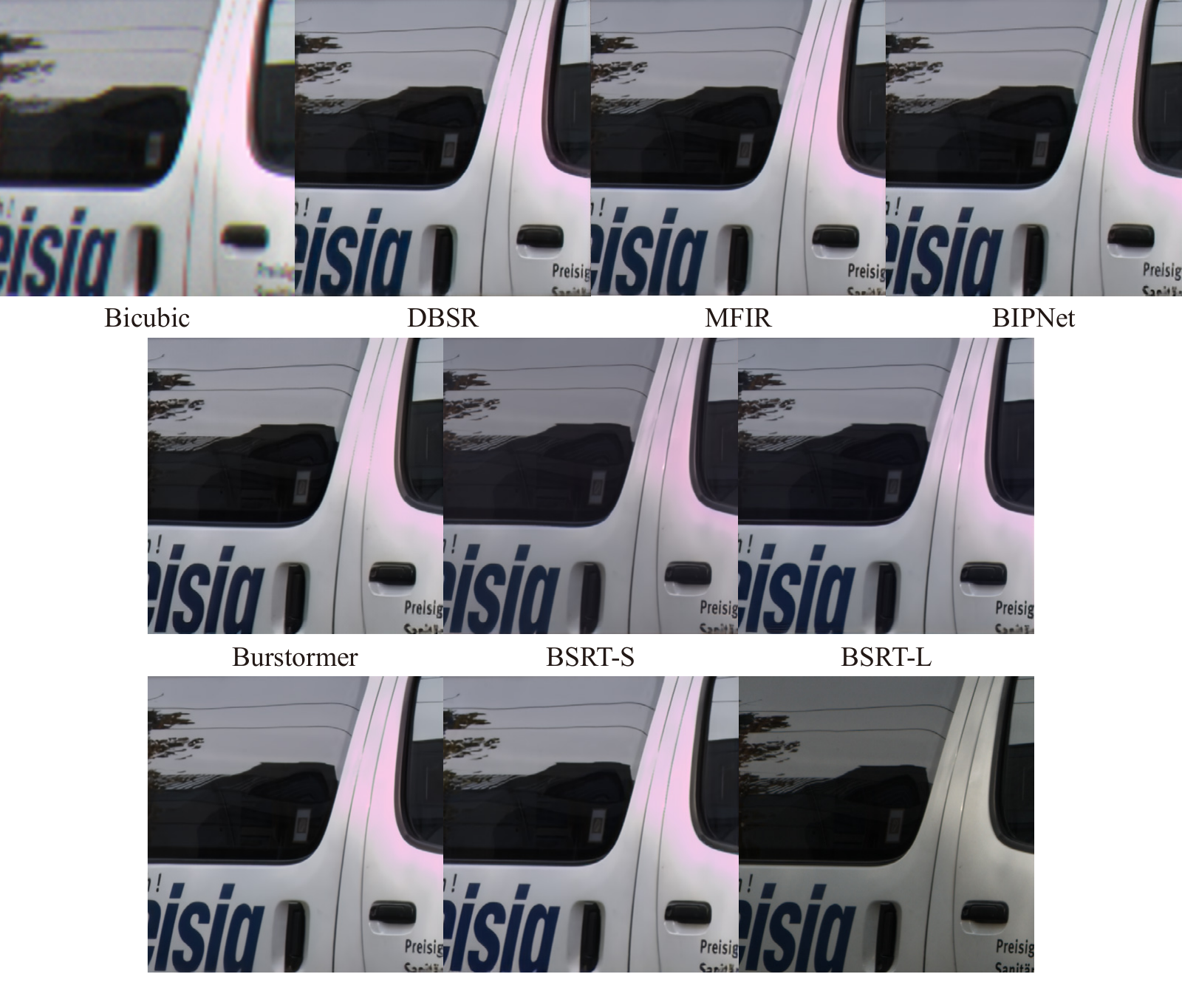}         
    \caption{Visual comparison of x4 BISR on the BurstSR
dataset.}
    \label{fig:sup_r_x4_4}
\end{figure}

\begin{figure}
    \centering
\includegraphics[width=0.95\linewidth]{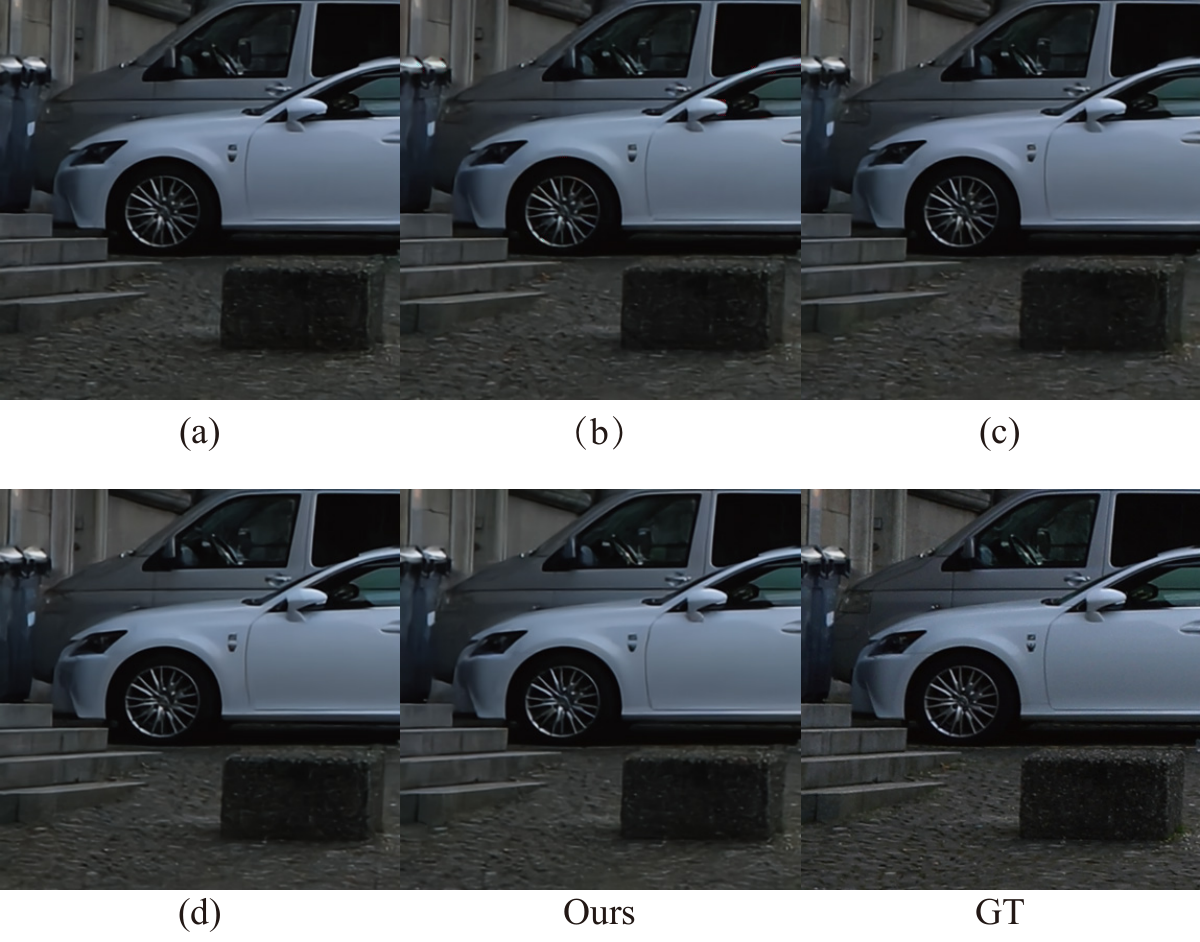}          
    \caption{Visual comparison of ablation study for x4 BISR on the SyntheticBurst dataset.}
    \label{fig:sup_z_ab_1}
\end{figure}

\begin{figure}
    \centering
\includegraphics[width=0.95\linewidth]{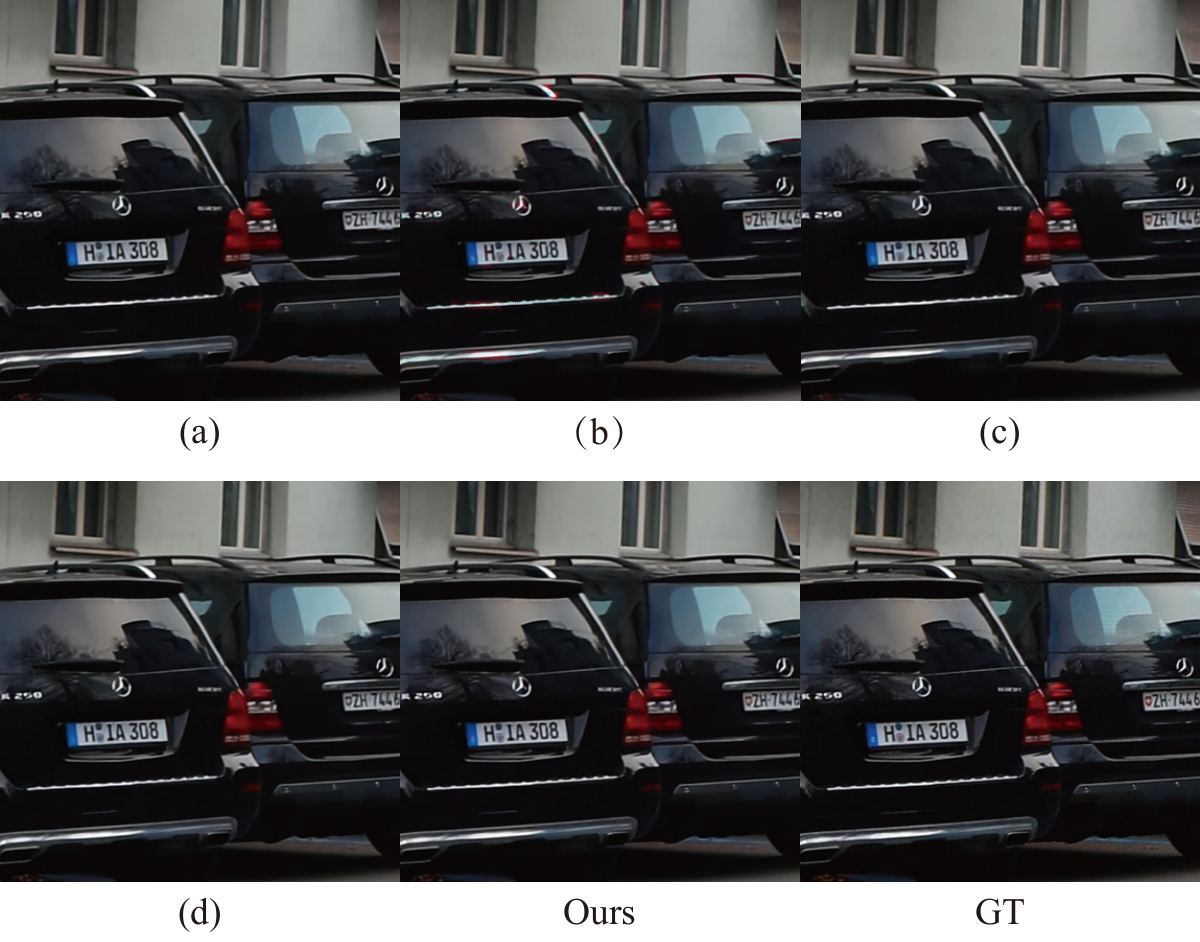}  
    \caption{Visual comparison of ablation study for x4 BISR on the SyntheticBurst dataset.}
    \label{fig:sup_z_ab_2}
\end{figure}

\begin{figure}
    \centering
\includegraphics[width=0.95\linewidth]{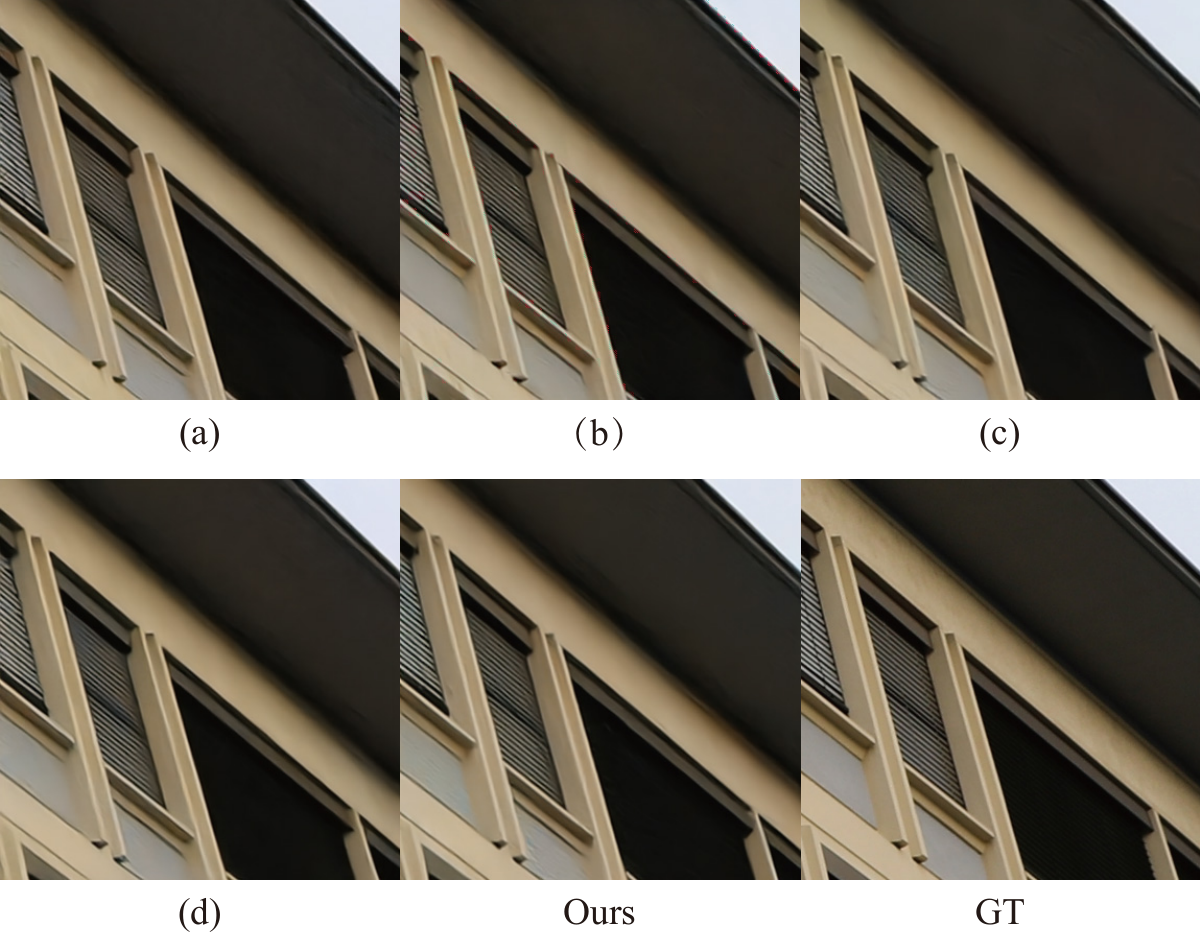}         
    \caption{Visual comparison of ablation study for x4 BISR on the SyntheticBurst dataset.}
    \label{fig:sup_z_ab_3}
\end{figure}

In this section, we first provide full-size visualized results of comparison experiments and ablation study in the main text.
Fig.~\ref{fig:sup_z_x4_1}-\ref{fig:sup_z_x4_4} are the full-size version of comparison results on x4 SyntheticBurst~\cite{bhat2021ntire} dataset.
Fig.~\ref{fig:sup_r_x4_1}-\ref{fig:sup_r_x4_4} are the full-size version of comparison results on x4 BurstSR~\cite{dbsr} dataset.
Fig.~\ref{fig:sup_z_ab_1}-\ref{fig:sup_z_ab_3} are the full-size visualized results of ablation study on x4 SyntheticBurst dataset.

In addition, we provide comprehensive multi-scale super-resolution (SR) comparisons on the SyntheticBurst and BurstSR datasets for both ×2 and ×3 scaling factors to further validate the robustness and generalization of our method. Among the compared methods, only BurstM \cite{kang2025burstm} supports multi-scale SR, while other methods exhibit inferior performance compared to both BurstM and our approach. Therefore, we focus our comparison on BurstM as the primary baseline.

\begin{figure}
    \centering
\includegraphics[width=0.9\linewidth]{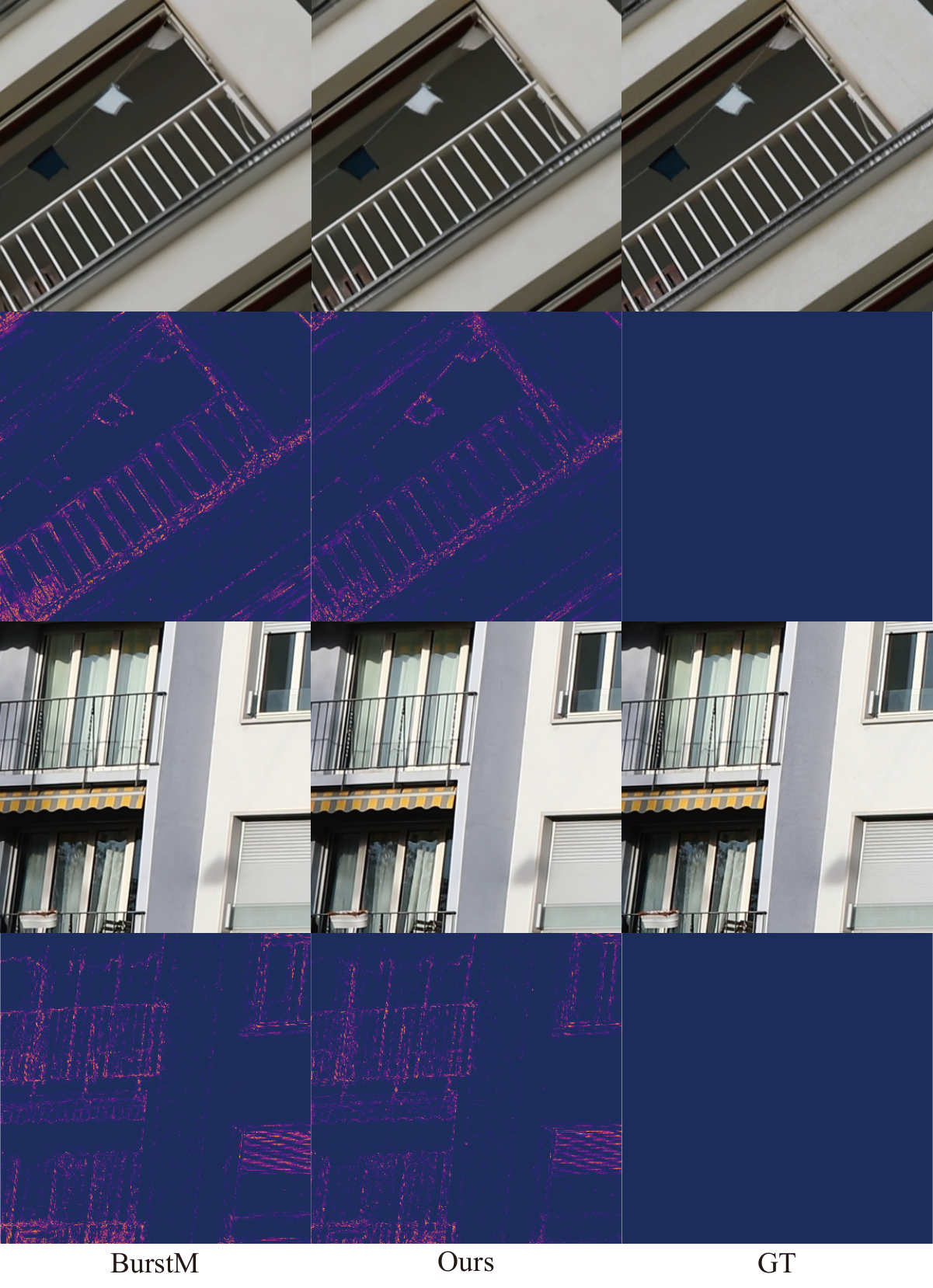}          
    \caption{Visual comparison x2 BISR on the SyntheticBurst dataset.}
    \label{fig:sup_syn_x2}
\end{figure}

\begin{figure}
    \centering
\includegraphics[width=0.9\linewidth]{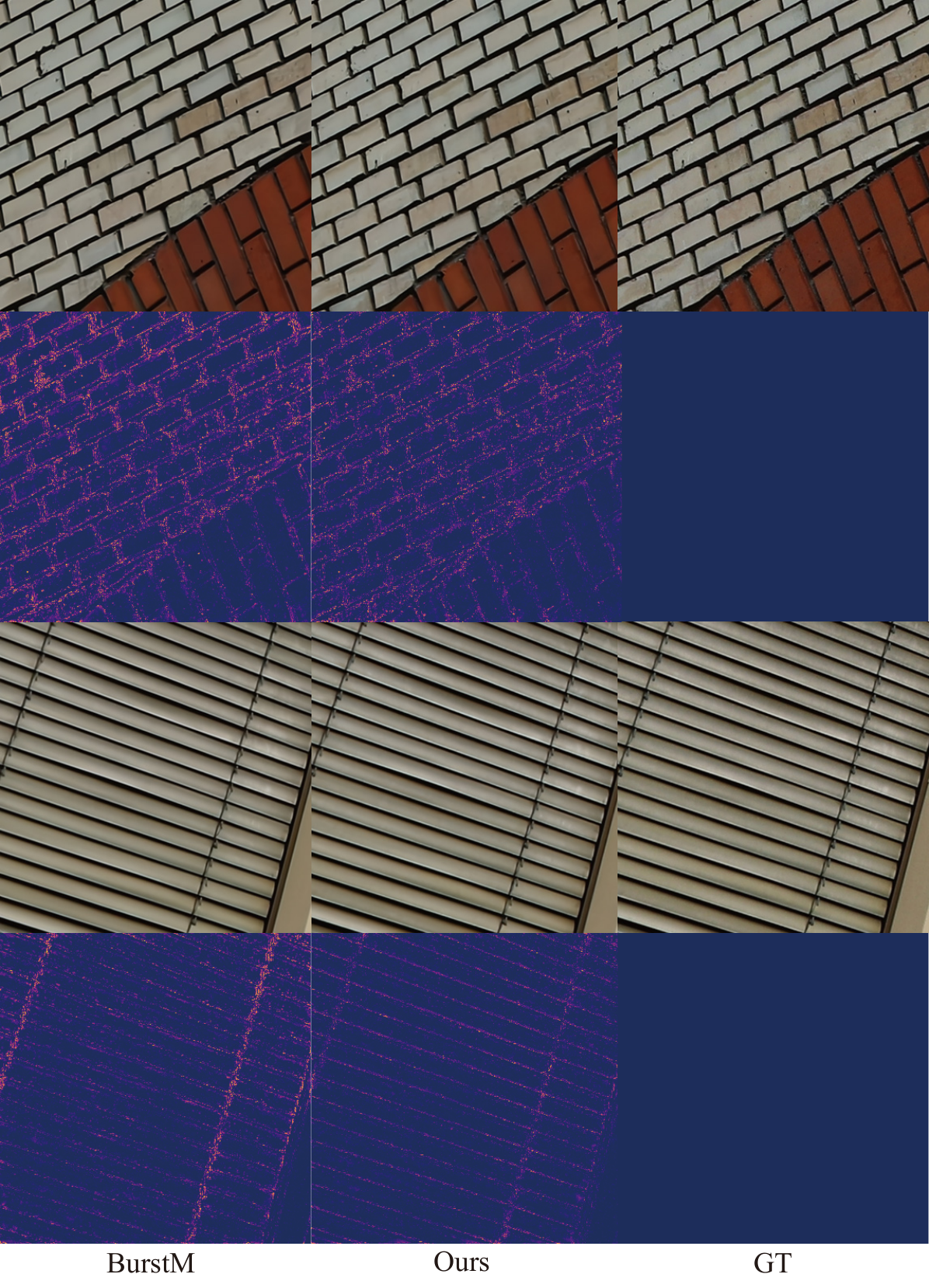}          
    \caption{Visual comparison x3 BISR on the SyntheticBurst dataset.}
    \label{fig:sup_syn_x3}
\end{figure}

\begin{figure}
    \centering
\includegraphics[width=0.9\linewidth]{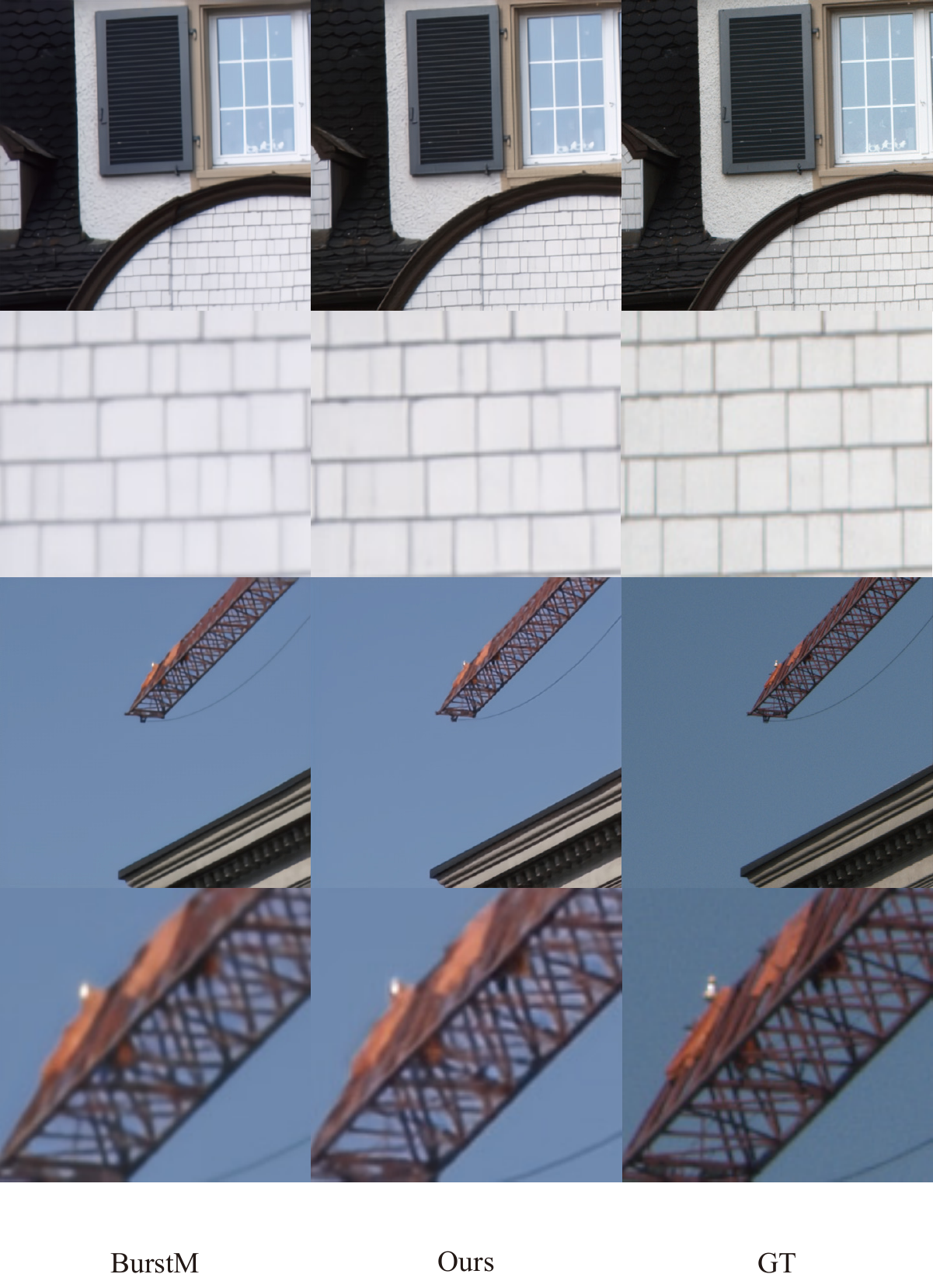}          
    \caption{Visual comparison x2 BISR on the BurstSR dataset.}
    \label{fig:sup_r_x2}
\end{figure}

\begin{figure}
    \centering
\includegraphics[width=0.9\linewidth]{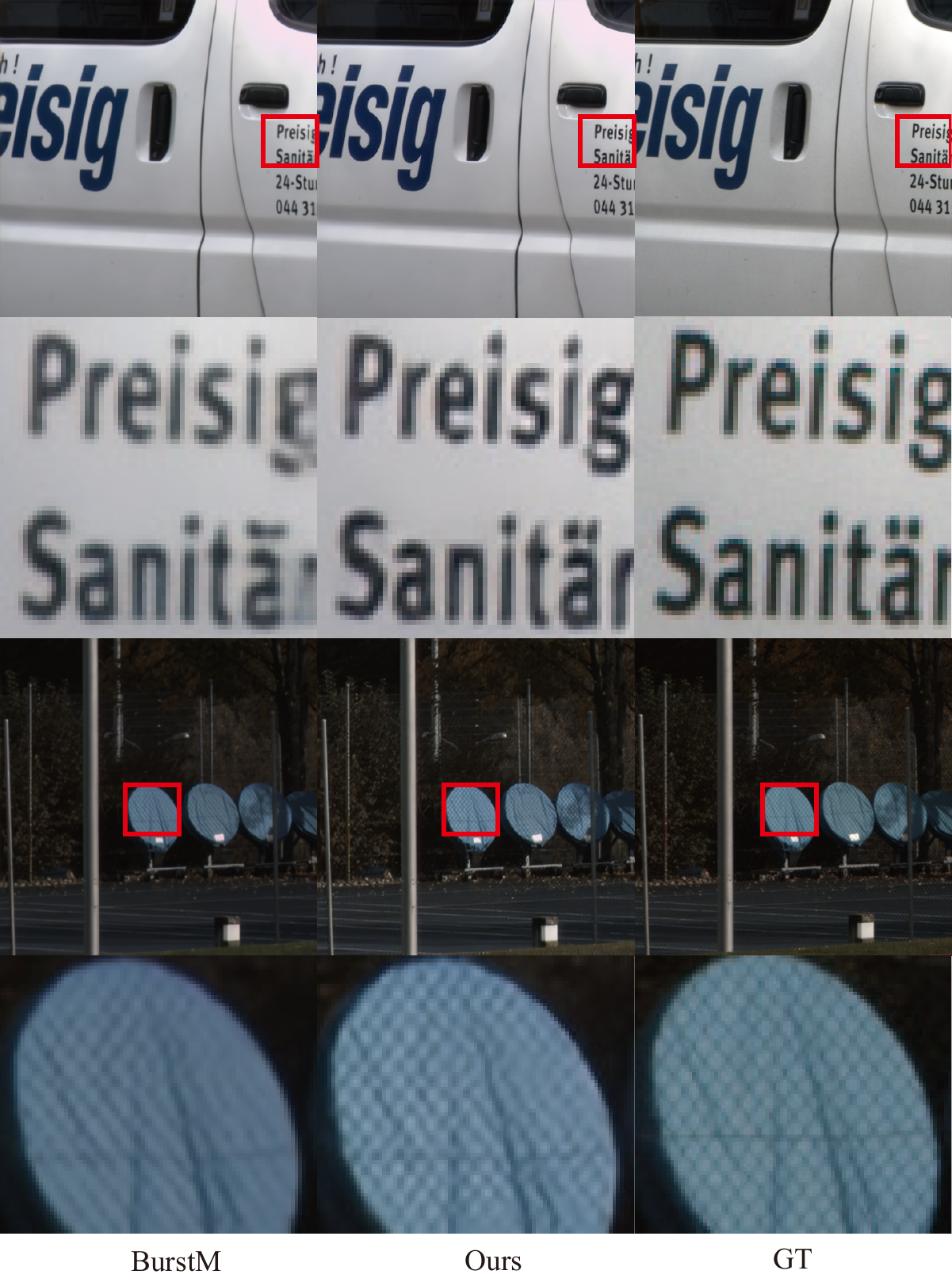}          
    \caption{Visual comparison x3 BISR on the BurstSR dataset.}
    \label{fig:sup_r_x3}
\end{figure}

The quantitative results are in Table 1 in the main text. The qualitative results on the x2 and x3 SyntheticBurst dataset \cite{bhat2021ntire} are presented in Fig. \ref{fig:sup_syn_x2} and Fig. \ref{fig:sup_syn_x3}. The error maps clearly demonstrate that our method achieves superior reconstruction quality, recovering finer details than BurstM.
For the x2 and x3 BurstSR dataset \cite{dbsr}, as shown in Fig. \ref{fig:sup_r_x2} and Fig. \ref{fig:sup_r_x3}, we provide enlarged patches for visual comparison due to the absence of ground truth images. The results highlight that our method preserves more structural information and produces visually sharper results compared to BurstM, further validating its effectiveness in real-world burst SR scenarios.
These consistent improvements across datasets and scaling factors underscore the robustness and generalizability of our approach.




\clearpage
{
    \small
    \bibliographystyle{ieeenat_fullname}
    \bibliography{references}
}